\documentclass[letterpaper,twoside,journal]{IEEEtran}
\usepackage{amsmath,amsfonts}
\usepackage{algorithmic}
\usepackage{algorithm}
\usepackage{array}
\usepackage[caption=false,font=normalsize,labelfont=sf,textfont=sf]{subfig}
\usepackage{textcomp}
\usepackage{stfloats}
\usepackage{url}
\usepackage{verbatim}
\usepackage{graphicx}
\usepackage{cite}

\usepackage{enumitem}
\usepackage{float}              % already present at line 24
\newfloat{paperlist}{tbp}{lop}  % name, default placement, list-file extension
\floatname{paperlist}{List}
\usepackage{multirow}
\usepackage{longtable}
\usepackage{tabularx}
\usepackage{ragged2e}
\usepackage{booktabs}
\usepackage{adjustbox}
\usepackage[dvipsnames]{xcolor}
\usepackage[hidelinks]{hyperref}
\usepackage[printonlyused]{acronym}
\usepackage{orcidlink}
\usepackage{fancyhdr}
\fancypagestyle{firstpage}{%
  \fancyhf{}%
  \fancyhead[C]{\parbox{\textwidth}{\centering\footnotesize This work has been submitted to the IEEE for possible publication. Copyright may be transferred without notice, after which this version may no longer be accessible.}}%
}

\begin{document}
\bstctlcite{IEEEexample:BSTcontrol}

\title{Biomazon: A Multimodal Dataset for 3D Forest Structure and Biomass Modeling in the Amazon Basin}

\author{Sayan~Mandal\orcidlink{0000-0003-3637-6029}, Rocco~Sedona\orcidlink{0000-0003-4089-972X}, Simon~Besnard\orcidlink{0000-0002-1137-103X}, Mikhail~Urbazaev\orcidlink{0000-0002-0327-6278}, Morris~Riedel\orcidlink{0000-0003-1810-9330}, Ehsan~Zandi\orcidlink{0000-0003-0135-257X}, Gabriele~Cavallaro\orcidlink{0000-0002-3239-9904}
\thanks{This work was supported by the 3D-ABC \cite{3dabc} project and computing time through the John von Neumann Institute for Computing (NIC) on the GCS Supercomputer JUWELS \cite{JUWELS} at Jülich Supercomputing Centre (JSC).}
\thanks{Corresponding author: Gabriele Cavallaro (g.cavallaro@fz-juelich.de)}
\thanks{Sayan Mandal, Morris Riedel and Gabriele Cavallaro are affiliated to Jülich Supercomputing Centre (JSC), Forschungszentrum Jülich, 52428 Jülich, Germany and School of Engineering and Natural Sciences (SENS), University of Iceland, 102 Reykjavík, Iceland (email: sa.mandal@fz-juelich.de; m.riedel@fz-juelich.de; g.cavallaro@fz-juelich.de).}
\thanks{Rocco Sedona and Ehsan Zandi are affiliated to Jülich Supercomputing Centre (JSC), Forschungszentrum Jülich, 52428 Jülich, Germany (email: r.sedona@fz-juelich.de; e.zandi@fz-juelich.de).}
\thanks{Simon Besnard and Mikhail Urbazaev are affiliated to Global Land Monitoring Group, GFZ Helmholtz Centre for Geosciences, Potsdam, Germany (email: simon.besnard@gfz.de; urbazaev@gfz.de).}
}

% \markboth{IEEE JOURNAL OF SELECTED TOPICS IN APPLIED EARTH OBSERVATIONS AND REMOTE SENSING}


\maketitle
\thispagestyle{firstpage}

\begin{abstract}
Accurate, spatially explicit characterization of tropical forest structure is essential for carbon accounting and ecosystem monitoring, yet most \acs{ML} pipelines predict canopy-top height proxies (e.g., \acs{RH95}/\acs{RH98}) or \acs{AGBD} as separate scalar targets, rather than learning the forest vertical structure as an ordered profile. The community lacks a \acs{ML}-ready multimodal benchmark for predicting the entire \acs{GEDI} \acs{RH} profile jointly with \acs{AGBD}, or for evaluating methods that enforce physically consistent ordering across \acs{RH} percentiles. We address this with \emph{Biomazon}, a 20\,m multimodal benchmark dataset over the Amazon Basin that pairs \acs{GEDI} \acs{RH} and \acs{AGBD} targets with multi-sensor predictors (Sentinel-1/2, ALOS-2 PALSAR-2, Copernicus \acs{DEM}, Dynamic World \acs{LULC}, and AlphaEarth embeddings) under standardized spatial splits and evaluation protocols. Using a shared encoder-decoder with task-specific heads as a baseline framework, we conduct a comprehensive ablation study of (i) backbone/model scale, (ii) modality contributions, and (iii) the use of auxiliary embeddings under standalone and fusion settings, and we report both single-target and joint-target results to quantify tradeoffs under a unified training protocol. Finally, we contextualize baseline performance through regionally aligned comparisons against existing gridded products, including \acs{GEDI} \acs{L4D} \acs{RH10}-\acs{RH98} and \acs{AGBD}, at matching temporal scale. \emph{Biomazon}, together with the accompanying protocols and baseline results, establishes a reference benchmark for future work on structurally consistent \acs{RH}-profile prediction and structure-biomass modeling in tropical forests. 
\end{abstract}


\section{Introduction}
\label{sec:introduction}

Tropical forests store roughly half of the world's forest carbon stocks \cite{carbon_sink} and host a disproportionate share of its biodiversity \cite{pillay_species}, making the spatially explicit characterization of their three-dimensional structure foundational to carbon accounting and biodiversity assessment. Forest structural complexity is a key element of ecosystem functioning \cite{de_Conto_2024}, and \acs{GEDI}’s footprint \ac{AGBD} estimates explicitly depend on vertical-structure (\ac{RH}) metrics rather than a single canopy-height proxy \cite{DUNCANSON2022112845, Kellner_2022}. NASA’s \ac{GEDI} was designed to directly observe this 3D structure from space using waveform lidar sampled at footprint scale across the Earth’s temperate and tropical forests \cite{Dubayah_2020}. Yet despite \ac{GEDI}’s 3D measurement capability, much of the machine-learning literature operationalizes “structure” as a single scalar, typically canopy height \cite{Lang_2023, Potapov_2021} or a small set of high relative-height percentiles, largely because these targets are convenient to train and evaluate wall-to-wall.

\begin{figure}[!t]
   \centering
    \includegraphics[width=\columnwidth]{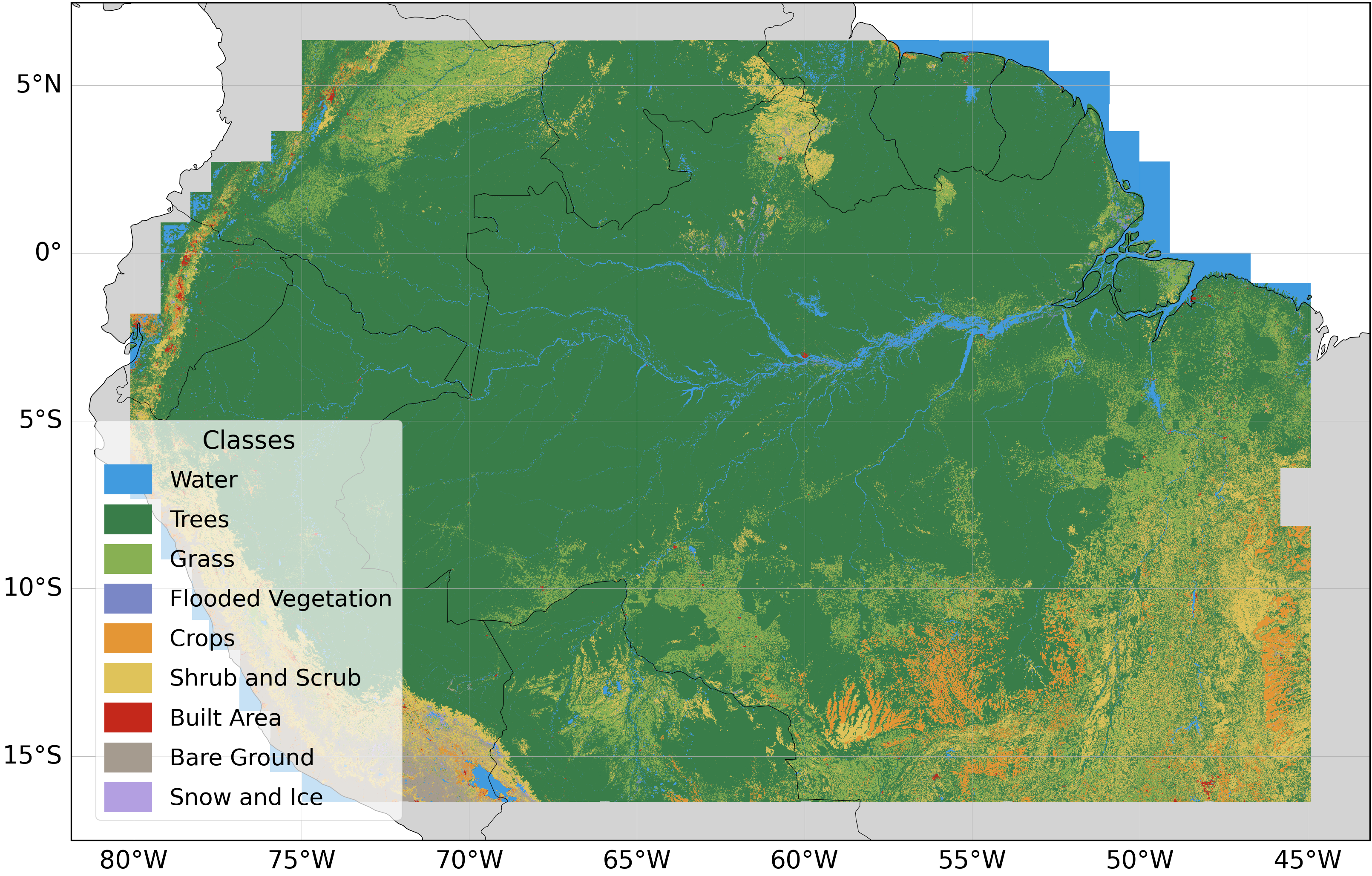}
    \caption{Spatial coverage of \emph{Biomazon} dataset in the Amazon Basin visualized using the Dynamic World V1 \acs{LULC} \cite{brown2022dynamic} modality (01/04/2019–30/03/2023), spanning 80.10°W–44.91°W and 16.37°S–6.33°N; gray indicates areas outside the coverage region.}
    \label{fig:overview}
\end{figure}

\IEEEpubidadjcol

This scalar framing has enabled substantial progress in large-area canopy height mapping by fusing sparse lidar supervision with dense satellite imagery. Global products demonstrate the feasibility of extrapolating \ac{GEDI} heights with optical time series at 10–30\,m resolution \cite{Potapov_2021}, sometimes with probabilistic modeling and uncertainty estimates \cite{Lang_2023}. In parallel, regional studies refine this paradigm by tailoring architectures and training strategies to specific ecosystems and sensors—ranging from \ac{GEDI}–Sentinel fusion for national-scale mapping and complex terrain \cite{Nazir_2026, Ahmad_2026, 10.3389/frsen.2025.1724950}, to models that leverage ICESat-2 in high-latitude ecotones \cite{Travers_Smith_2024}, and approaches that push resolution using \ac{VHR} RGB or Planet NICFI imagery when such data are available \cite{Tolan_2024, Wagner_2025}.

However, collapsing \ac{GEDI}’s vertical information into one number misses a central scientific opportunity: \ac{GEDI} waveforms encode an ordered height distribution that can be summarized as a \acf{RH} profile across percentiles (e.g., \acs{RH10}…\acs{RH98}), providing a structured description of vertical occupancy \cite{Dubayah_2021_GEDI_L2A_V002}. Modeling this \ac{RH} profile as a first-class prediction target changes the task from scalar regression to structured output learning, and introduces a physical consistency requirement: \ac{RH} percentiles must be monotone non-decreasing by construction, which we enforce with a simple anchored cumulative-sum parameterization. Today, the community lacks a widely used, machine-learning-ready benchmark that (i) treats the full \ac{RH} profile as the primary target, (ii) pairs it with \ac{AGBD} to study structure–biomass coupling under unified protocols, and (iii) evaluates whether methods enforce physically consistent ordering across percentiles.

A second gap is methodological and practical. While foundation embeddings such as AlphaEarth Foundations \cite{brown2025alphaearthfoundationsembeddingfield} and TESSERA \cite{feng2025tesseratemporalembeddingssurface} aim to compress multi-sensor, spatiotemporal information into analysis-ready representations for downstream mapping, evidence is mixed on whether such embeddings consistently outperform strong task-specific baselines (e.g., spectral indices) for biomass estimation in tropical forests \cite{lucero2026spectral}. The lack of a standardized benchmark for \ac{RH} profiles and \ac{AGBD} makes it difficult to quantify when embeddings help, how they compare to raw modalities under controlled protocols, and how they behave under single-target and joint-target training setups.

We address these gaps with \emph{Biomazon}, a multimodal benchmark dataset over the Amazon Basin (spatial coverage shown in Fig.~\ref{fig:overview}) that integrates Sentinel-1/2, ALOS-2 PALSAR-2, Copernicus GLO-30 \acs{DEM}, Dynamic World V1 \acs{LULC} \cite{brown2022dynamic} and AlphaEarth embeddings \cite{brown2025alphaearthfoundationsembeddingfield} aligned with \ac{GEDI} \ac{RH} and \ac{AGBD} targets and  having standardized spatial splits and evaluation procedures. Our design is motivated by recent benchmark efforts that accelerated progress in related problems by providing reproducible datasets and protocols \cite{Fogel_2025, debary2025preditree, NEURIPS2023_40daf2a0, Sialelli_2025}. However, existing benchmarks typically target either forest height or biomass estimation, and thus do not support prediction of the full \ac{RH} profile, nor its joint prediction with \ac{AGBD}. Using a shared encoder–decoder with task-specific heads as a baseline framework, we perform controlled ablations over backbone/model scale, modality contributions, and auxiliary embeddings under both standalone and fusion settings. Finally, to contextualize baseline performance for end users, we provide product-aligned comparisons to widely used gridded forest-structure and \ac{AGBD} products, including \ac{GEDI} \acs{L4D} imputed \ac{RH} and \ac{AGBD} layers \cite{Seo_2025_GEDI_L4D_Imputed_Waveforms_V2}, global canopy-height maps \cite{Lang_2023, Potapov_2021, Wagner_2025, Tolan_2024}, and biomass products such as ESA Biomass CCI \cite{esa_cci}.

This paper makes four primary contributions:

\begin{enumerate}
    \item \textbf{Benchmark dataset + protocols}: We introduce \emph{Biomazon}, a 20\,m resolution \acs{ML}-ready multimodal dataset for the Amazon Basin spanning April 2019--March 2023, supporting joint prediction of the full \ac{GEDI} \ac{RH} profile and \ac{AGBD}, with standardized spatial splits and evaluation.
    \item \textbf{\ac{RH}-profile as a structured \acs{ML} task}: We cast full \ac{RH}-profile prediction as a structured learning task, moving beyond scalar canopy-height proxies to model the ordered vertical distribution of forest structure.
    \item \textbf{Task-specific anchored monotone parameterization}: We propose and benchmark an \ac{RH}-specific anchored monotone parameterization for full \ac{RH}-profile prediction, using \acs{RH100} as an anchor and cumulative nonnegative increments to enforce monotonic outputs while accommodating negative values in the lower \ac{RH} percentiles observed in \ac{GEDI} labels.
    \item \textbf{Systematic ablations + product-aligned evaluation}: We provide a comprehensive study of (i) encoder model scale (Prithvi variants: 5M/100M/300M parameters), (ii) raw modalities vs. AlphaEarth embeddings, (iii) full-\ac{RH} vs. individual-\ac{RH} targets, and (iv) joint-target vs. single-target training, and we contextualize results via aligned comparisons to existing canopy-height and \acs{GEDI}-derived gridded products.
\end{enumerate} 

The rest of this article is organized as follows: Sec.~\ref{sec:background} presents related works. Sec.~\ref{sec:biomazon} introduces the \emph{Biomazon} dataset. Sec.~\ref{sec:methodology} describes our methodology and Sec.~\ref{sec:results} details the results and discussions of our ablation study. We highlight the main observations, potential limitations and future work directions in Sec.~\ref{sec:summary_limitations_future} and consequently conclude our article in Sec.~\ref{sec:conclusion}.

\section{Background}
\label{sec:background}

\subsection{GEDI-Supervised Multi-Sensor Modeling of Canopy Height and AGBD}
\label{subsec:background1}

Most canopy-height and biomass pipelines treat lidar-derived targets as sparse supervision and learn a mapping from dense satellite predictors to wall-to-wall predictions. The methodological machinery overlaps heavily between the two, with only the target swapping between a high \ac{RH} percentile (canopy height) and \ac{AGBD}. At global scale, Lang et al.\ \cite{Lang_2023} fuse \ac{GEDI} with Sentinel-2 using a probabilistic deep-learning approach with uncertainty, delivering a 10\,m canopy-height product, and Potapov et al.\ \cite{Potapov_2021} extrapolate \ac{GEDI} heights with Landsat time series at 30\,m for 2019. Regional pipelines adapt this fusion to local complexity \cite{Nazir_2026, Ahmad_2026, Travers_Smith_2024}, while higher-resolution variants push toward \ac{VHR} imagery or super-resolved Sentinel-2 under high-density lidar supervision \cite{Tolan_2024, Wagner_2025, Ekaterina_2025, boudras2026serahnativesentinelspatial}. Target-calibration work highlights that \ac{GEDI} height accuracy can vary and benefits from \ac{ALS}-based correction \cite{CHO2025100221}.

For \ac{AGBD}, \ac{GEDI}'s footprint-level \ac{L4A} product is grounded in the mission algorithm paper and ATBD \cite{DUNCANSON2022112845, Kellner_2022}. Multi-sensor \ac{EO} deep-learning models map biomass by integrating \ac{GEDI} with Sentinel-1/2 and ALOS-2 PALSAR-2 \cite{dong2023forestabovegroundbiomassestimation} and via attentive neural processes with calibrated uncertainty for interpolation from sparse \ac{GEDI} to wall-to-wall maps \cite{young2026interpolationgedibiomassestimates}. Classic L-band \ac{SAR} work clarifies both signal and saturation behavior in tropical and boreal forests \cite{hamdan2011remotely, Peregon_2013}, and fusion of MODIS--Landsat NDVI products has been operationalized for vegetation \ac{AGB} mapping \cite{rs17223754}. Together, these studies motivate \emph{Biomazon}'s sensor selection (Sentinel-1/2, ALOS-2 PALSAR-2 alongside \ac{DEM} and \ac{LULC}) and clarify why we treat \ac{GEDI} vertical structure as more than a single ``canopy height map'' target, pairing \ac{AGBD} with \ac{RH} profiles in a tropical biome under a consistent protocol.

\subsection{Modeling Vertical Structure Beyond Canopy Top-Height: RH Metrics and Profile Learning}
\label{subsec:background2}

GEDI waveforms are summarized as \acf{RH} percentiles (RH0…RH100), which are inherently ordered and provide a compact description of the vertical return distribution. Nevertheless, most \acs{ML} pipelines still pick a small subset—typically high percentiles used as canopy-height proxies—because they are convenient to regress and benchmark. Multiple lines of evidence suggest that “profile thinking” is useful. Xiao et al. \cite{11010835} build a national canopy-height product by explicitly fusing multiple \ac{RH} metrics (RH0–RH100 at 10\% steps) with Landsat-8 and Sentinel-1 rather than relying on a single \acs{RH95}/\acs{RH98} proxy, and show that this improves accuracy. 

Beyond canopy top-height, several \acs{GEDI} applications directly connect different \acs{RH} ranges to different forest strata, reinforcing the value of learning ordered \acs{RH} structure rather than isolated percentiles. Fricker et al. \cite{Fricker_2021} use \acs{RH25} and \acs{RH75} as understory and overstory height proxies respectively, and Mohammadpour et al. \cite{Mohammadpour_2025} incorporate \acs{GEDI} metrics via decision rules for understory fuel characterization. Consistently, the \acs{GEDI} \ac{L4A} uses more than one \ac{RH} percentile to estimate \ac{AGBD} within its biomass modeling framework \cite{DUNCANSON2022112845, Kellner_2022}, underscoring that operational biomass retrieval already exploits information distributed across the vertical profile.

A closely related recent direction is to model the \ac{GEDI} waveform itself rather than only selected \ac{RH} metrics. DUNIA \cite{fayad2025dunia} aligns satellite imagery with \ac{GEDI} full-waveform lidar through cross-modal representation learning and evaluates waveform retrieval/generation alongside canopy height, canopy cover, and plant area index. Since \acs{RH} metrics are waveform-derived energy quantiles, such approaches are adjacent to full \acs{RH}-profile modeling and can, in principle, yield monotone \ac{RH} profiles after waveform-to-RH conversion. Nevertheless, waveform-oriented representation learning and direct \acs{RH}-profile supervision remain distinct experimental formulations, with different primary targets, losses, and evaluation protocols.

Motivated by this, \emph{Biomazon} turns the ordered \ac{RH} vector into a first-class structured target so models can be compared on both accuracy and structural validity rather than only on a canopy-top scalar.

\subsection{Benchmark Datasets and Protocols: From Canopy Height and Biomass Tasks to Structured Forest Targets}
\label{subsec:background4}

Recent benchmark efforts like Pangea \cite{pangea} and GEO-Bench \cite{geobench} show why standardized datasets and protocols matter for reproducible progress in building foundation models. More broadly, the same principle is crucial for developing and comparing task-specific models in remote sensing. Open-Canopy \cite{Fogel_2025} provides an open, country-scale benchmark for very high-resolution canopy height estimation (and change) with aerial lidar ground truth, explicitly designed for repeatable computer-vision comparisons. Preditree \cite{debary2025preditree} contributes a multi-temporal sub-meter dataset of multispectral imagery aligned with \ac{ALS} canopy height maps. BioMassters \cite{NEURIPS2023_40daf2a0} provides a multi-modal Sentinel-1/2 time-series benchmark for biomass estimation using \ac{ALS} references and AGBD \cite{Sialelli_2025} curates a global-scale multimodal \acs{ML}-ready biomass dataset with \ac{GEDI} \ac{AGBD} as target.

Beyond single-task benchmarks, standardized protocols also enable unified modeling and evaluation across multiple forest attributes. Weber et al. \cite{Weber_2025} train a single multi-head model to predict \ac{AGBD}, canopy height, and canopy cover globally from multi-sensor imagery with uncertainty estimation, showing that co-modeling forest attributes can be practical at scale—yet structure is still framed as a height scalar rather than an ordered \ac{RH} profile. \emph{Biomazon} fills the missing niche these benchmarks leave open: a multimodal, \ac{ML}-ready protocol for full \ac{RH} profile prediction together with \ac{AGBD} in the Amazon, so methods can be compared on structured outputs, inter-target interactions, and product-aligned evaluation under standardized splits.

\subsection{Foundation Embeddings as Predictors: AlphaEarth Foundations and TESSERA}
\label{subsec:background5}

EO representation learning increasingly offers “analysis-ready” embeddings meant to reduce multi-sensor preprocessing burdens and support label-efficient mapping. AlphaEarth Foundations \cite{brown2025alphaearthfoundationsembeddingfield} proposes an embedding field model that assimilates spatial, temporal, and measurement contexts across multiple sources and releases annual global 10\,m embedding layers. TESSERA \cite{feng2025tesseratemporalembeddingssurface} similarly provides open, global 10\,m “ready-to-use” embeddings learned from Sentinel-1/2 time series with self-supervised training and reports competitive performance across downstream tasks. At the same time, Lucero et al. \cite{lucero2026spectral} report that spectral indices can outperform AlphaEarth embeddings for \ac{AGB} estimation in tropical Andean forests under their experimental setup, warning against assuming embeddings dominate strong domain baselines everywhere. \emph{Biomazon} therefore treats embeddings as first-class predictors alongside raw modalities and evaluates them under identical splits and training recipes, enabling clear, region-specific conclusions about when embeddings help \ac{RH}-profile prediction and \ac{AGBD}, and whether gains persist under single-target vs joint-target learning.

\subsection{Physically Consistent Learning for Ordered RH Profiles}
\label{subsec:background6}

Predicting a \ac{RH} profile is structurally different from predicting a scalar because the outputs must satisfy an ordering constraint $\mathrm{RH}_{0} \le \mathrm{RH}_{10} \le \cdots \le \mathrm{RH}_{100}$ by definition of percentiles, and violating monotonicity yields physically inconsistent profiles even if individual percentile errors look small. Monotonic neural modeling has long been studied \cite{monotonic}, with non-crossing quantile architectures guaranteeing ordered outputs via cumulative-sum constructions \cite{10.5555/3495724.3497058, Zhou_2021, luo2021distributional}. In \emph{Biomazon} we benchmark a monotone-by-construction \ac{RH} parameterization anchored at \acs{RH100} with nonnegative cumulative increments, so ordering is enforced during learning and the accuracy--consistency tradeoff is measurable across models and modalities. Because forest structure and biomass products are often consumed as maps, we additionally contextualize model performance against widely used gridded references --- global canopy-height \cite{Lang_2023, Tolan_2024, Potapov_2021, Wagner_2025} and biomass \cite{esa_cci} products, plus \ac{GEDI} \ac{L4D} imputed \ac{RH} and \ac{AGBD} products \cite{Seo_2025_GEDI_L4D_Imputed_Waveforms_V2} as a natural comparator for RH-profile modeling pipelines.

\section{Biomazon Dataset}
\label{sec:biomazon}

\begin{figure*}[!t]
   \centering
    \includegraphics[width=\textwidth]{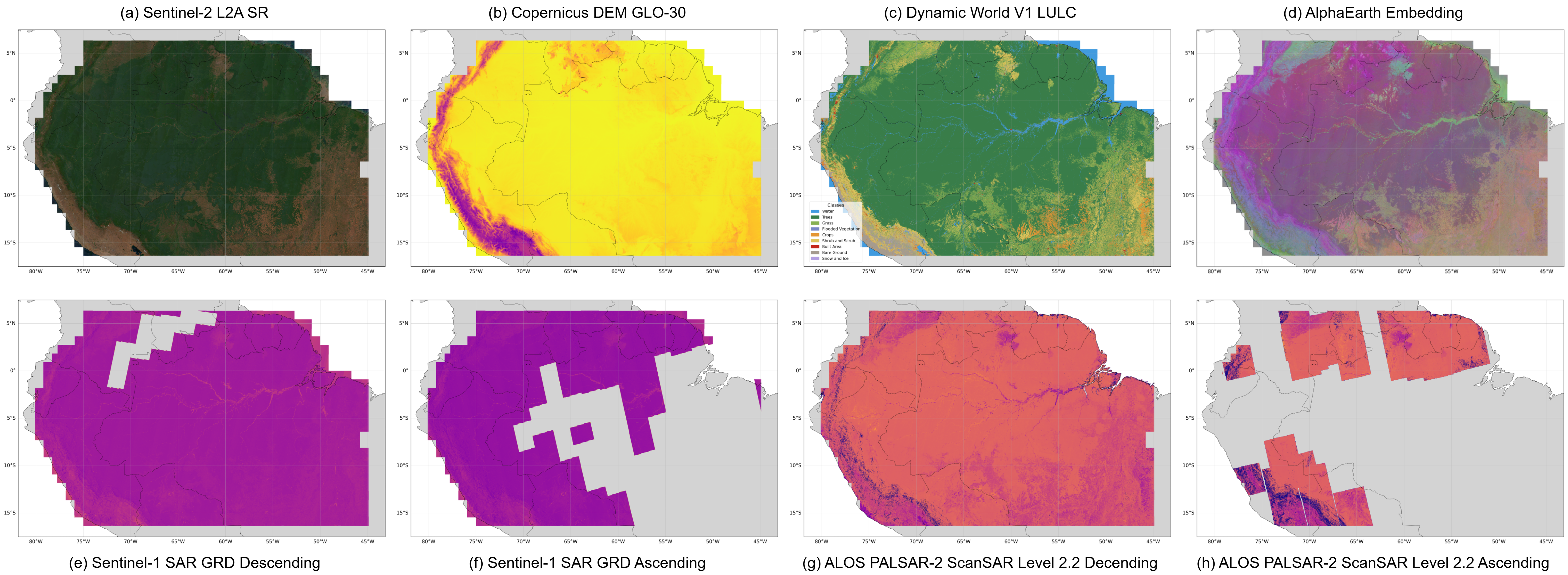}
    \caption{Extent of spatial coverage of the acquisitions of the input modalities across the study area. We can see (a), (b), (c) and (d) have full coverage. (g) ALOS-2 PALSAR-2 ScanSAR Level 2.2 Descending is also complete for the land surface of the study area and masked on sea surfaces. (f) Sentinel-1 and (g) ALOS-2 PALSAR-2 ascending modalities have major missing acquisitions.}
    \label{fig:coverage_area}
\end{figure*}

\subsection{Study Region}
\label{subsec:studyregion}

The Amazon Basin is a compelling study area for \emph{Biomazon} because it combines global ecological significance with strong spatial heterogeneity in forest structure and biomass. Beyond being one of the world’s largest continuous tropical forest regions, its standing biomass represents one of the largest terrestrial carbon pools, estimated in the order of ~150–200 GtC \cite{Wagner_2025}, underscoring why structure and biomass monitoring in this biome is globally consequential. It is also a major component of the Earth system whose forests modulate regional climate via moisture recycling and land–atmosphere coupling, so changes in forest cover and condition can translate into measurable shifts in rainfall patterns and dry-season intensity \cite{Qin_2025, Cui_2026}. At the same time, the region is under widespread degradation from interacting stressors (e.g., fire, edge effects, logging, and drought) \cite{lapola_drivers, Flores2024}, with impacts on carbon emissions that can rival deforestation in magnitude \cite{lapola_drivers}, making it an ideal setting to benchmark models that aim to learn vertical structure and biomass jointly rather than as disconnected scalars.

\subsection{Modalities}
\label{subsec:biomazon1}

\begin{table}[t]
\centering
\caption{Summary of \emph{Biomazon} modalities and band composition.}
\label{tab:modality_summary}
\renewcommand{\arraystretch}{1.15}
\setlength{\tabcolsep}{4pt}
\begin{tabularx}{\columnwidth}{l r >{\RaggedRight\arraybackslash}X}
\hline
Modality & Band Count & Band Names \\
\hline
SEN2 & 10 & B2, B3, B4, B8, B11, B12, B5, B6, B7, B8A \\
SEN1 & 6 & VV\_asc, VH\_asc, VV\_desc, VH\_desc, angle\_asc, angle\_desc \\
ALOS & 6 & HH\_asc, HV\_asc, HH\_desc, HV\_desc, LIN\_asc, LIN\_desc \\
DEM  & 1 & DEM \\
LULC & 1 & label \\
AEX  & 64 & A00, A01, ..., A63 \\
RH   & 101 & RH0, RH1, ..., RH98, RH99, RH100 \\
AGBD & 1 & AGBD \\
\hline
\end{tabularx}
\end{table}

We use the predictor modalities and targets summarized in Table~\ref{tab:modality_summary}, which also lists the bands retained for each modality, and in Fig.~\ref{fig:coverage_area} we show the study-region coverage of each co-registered predictor modality. For consistency across sources, all predictors and targets in \emph{Biomazon} are mapped to a common 20\,m spatial grid over the Amazon Basin, co-registered tile-by-tile within the \ac{HLS} \ac{MGRS} (HLS-MGRS) tiling system across 913 tiles, and aligned to a shared benchmark period spanning 1 April 2019 to 30 March 2023. We use \ac{GEE} \cite{gee} to download all products except the \ac{GEDI} data, which we download from \texttt{gediDB} \cite{Besnard2025}. Spatial resampling to the 20\,m grid uses \ac{GEE}'s default nearest-neighbor method, which subsamples one source pixel per output pixel rather than averaging across them. This is particularly relevant for the \acs{SAR} products (Sentinel-1, ALOS-2) because it preserves the dB-scale values without the log-scale bias that arithmetic averaging in dB would introduce. The following subsections describe the modality-specific preprocessing and temporal compositing steps, followed by the dataset creation methodology and overall statistics.

\subsubsection{Sentinel-2 L2A Surface Reflectance}
\label{subsubsec:modality1}

We use Sentinel-2 \ac{L2A} surface reflectance \cite{S2L2A_Copernicus} as the optical backbone of our dataset, preferring its atmospherically corrected \ac{BOA} reflectance over \ac{L1C} \ac{TOA} reflectance for better comparability of vegetation signals \cite{Ginting_2024}. We favor Sentinel-2 over Landsat-8 because its 10--20\,m bands and dedicated red-edge channels outperform Landsat-8's 30\,m bands in forest-variable, canopy-cover and \ac{LAI} retrieval \cite{ASTOLA2019257, KORHONEN2017259}. From the \ac{MSI} configuration we retain the ten land-surface bands listed in Table~\ref{tab:modality_summary}, which span the visible, \ac{NIR}, red-edge, and \ac{SWIR} regions, that are most informative for vegetation biophysical retrieval. The red-edge bands in particular improve chlorophyll/\ac{LAI} retrieval, forest canopy-closure estimation and \acs{GEDI}-supervised canopy-height and \ac{AGB} mapping \cite{FRAMPTON201383, Delegido2011, Hua2021, Tamiminia2024}.

For data preparation, we queried the harmonized collection from \ac{GEE} \cite{S2SRHarmonized_GEE} for each \ac{HLS}-\ac{MGRS} tile over the benchmark period, retained only acquisitions containing all ten required bands, and linked them to the Cloud Score+ \texttt{cs} \ac{QA} band \cite{cloudscore_gee, cloudscore_paper} by shared \texttt{system:index} for per-pixel quality scores. We composited via \texttt{qualityMosaic(cs)} (best per-pixel score), converted integer-scaled bands to reflectance using the documented $0.0001$ scale factor, and reprojected to the common 20\,m \ac{HLS}-\ac{MGRS} grid.

\subsubsection{Sentinel-1 SAR GRD}
\label{subsubsec:modality2}

We include Sentinel-1 \ac{GRD} C-band \ac{SAR} \cite{sen1} as an all-weather structural modality. \ac{GEE} provides Sentinel-1 \ac{GRD} preprocessed with thermal-noise removal, radiometric calibration, and terrain correction, in decibel (dB) scale with an incidence-angle band \cite{sen1, sen1_algo}. We restrict the collection to \ac{IW} 10\,m dual-polarization products containing \texttt{VV}, \texttt{VH}, and \texttt{angle}, and build separate ascending and descending median composites (Table~\ref{tab:modality_summary}). Dual-polarization \texttt{VV}/\texttt{VH} encodes complementary scattering, with cross-polarized returns more strongly associated with canopy multiple scattering than co-polarized ones \cite{tianreview}. Where one orbit direction is unavailable, the corresponding pass-specific bands are filled with \(-9999.0\) \texttt{NODATA} to preserve a fixed six-band layout.

\begin{figure}[!t]
   \centering
    \includegraphics[width=\columnwidth]{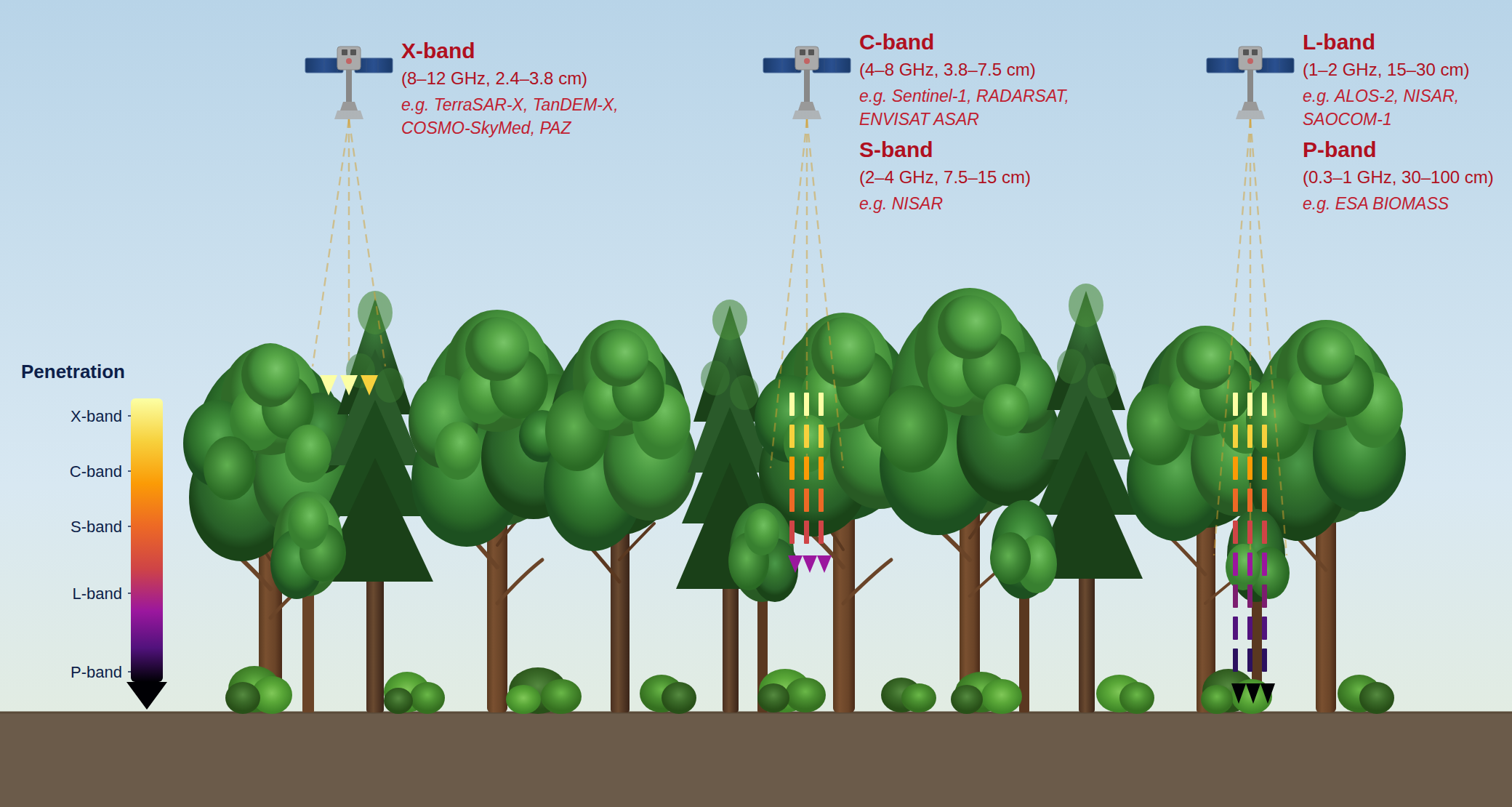}
    \caption{Radar signal interaction with forest canopy across major \ac{SAR} frequency bands (X-, C-, S-, L-, and P-band) in forested environments. Lower-frequency bands such as L- and P-band penetrate the full canopy volume, interacting with trunks, large branches, and the ground surface, providing sensitivity to above-ground biomass density (\ac{AGBD}) and the vertical distribution of forest structure essential for retrieving complete relative height (\ac{RH}) profiles. Higher-frequency bands (X-, C-, and S-band), whose phase centers lie near the canopy top, serve as effective proxies for canopy top height estimation but offer limited sensitivity to deeper structural layers that shape the lower portions of the \ac{RH} distribution and drive total biomass accumulation.}
    \label{fig:sar_penetration}
\end{figure}

C-band waves interact mainly with leaves and small or secondary branches \cite{zhou_leaves}, so they saturate earlier than L-band on tall closed-canopy biomass. We therefore expect Sentinel-1 to contribute primarily to canopy-surface and upper-canopy structural information (Fig.~\ref{fig:sar_penetration}), with residual sensitivity in lower-to-moderate biomass regimes \cite{Musthafa_sar, tianreview, huang_sar}. Prior \ac{GEDI}--Sentinel-1--Sentinel-2 fusion has supported wall-to-wall modeling of vegetation structure, canopy height, and biomass \cite{Kacic_structure, SCHWARTZ2024103711, Tamiminia2024}.

\subsubsection{ALOS-2 PALSAR-2 ScanSAR Level~2.2}
\label{subsubsec:modality3}

To extend \ac{SAR} information beyond C-band's shallow canopy sensitivity, we include JAXA ALOS-2 PALSAR-2 ScanSAR Level~2.2 dual-polarization \texttt{HH}/\texttt{HV} L-band data \cite{alos2}, which \ac{GEE} distributes ortho-rectified, radiometrically terrain-corrected, slope-corrected via AW3D30, and \ac{CARD4L} compliance. The longer L-band wavelength penetrates deeper into the canopy and saturates at higher \ac{AGB} than C-band (Fig.~\ref{fig:sar_penetration}), making it well suited to forest \ac{AGB} mapping and to detecting woody-cover and vegetation-volume change \cite{Musthafa_sar, tianreview, huang_sar, WESSELS2023113369}.

We retained acquisitions containing the \texttt{HH}, \texttt{HV}, \texttt{LIN} (local incidence angle), and \texttt{MSK} (data-quality bitmask) bands. \texttt{HH} and \texttt{HV} are stored as digital numbers (\acsp{DN}) in the \ac{GEE} asset and are converted to terrain-flattened gamma-naught backscatter in dB via
\begin{equation}
\label{eq:alos_convert}
    \gamma^{0}_{\text{dB}} = 10 \log_{10}(\text{DN}^2) - 83.0~\text{dB},
\end{equation}
where \acs{DN} is the digital number stored in the \ac{GEE} asset. Acquisitions are split into ascending and descending passes via \texttt{PassDirection}. For each pass we kept only \texttt{MSK}-valid pixels and computed median composites of \texttt{HH}, \texttt{HV}, and \texttt{LIN} (the latter rescaled to degrees by \texttt{LIN}/100). The resulting pass-specific composites (Table~\ref{tab:modality_summary}) are reprojected to the common 20\,m grid.

\subsubsection{Copernicus DEM GLO-30}
\label{subsubsec:modality4}

We include Copernicus \ac{DEM} GLO-30 \cite{copernicus_dem} as a static elevation-context modality, retaining only its 30\,m \texttt{DEM} band (derived from the edited WorldDEM/TanDEM-X product) and reprojecting it to the 20\,m grid. Although distributed as a \ac{DEM}, the product is not a \ac{DTM} but formally a \ac{DSM} that includes vegetation and other above-ground features, so we use it as an auxiliary elevation-context predictor rather than as a strict bare-earth terrain model in closed-canopy forest.

\subsubsection{Dynamic World V1 LULC}
\label{subsubsec:modality5}

We include Dynamic World V1 \ac{LULC} \cite{brown2022dynamic, lulc} as a categorical context modality. Dynamic World provides near-real-time 10\,m class probabilities from Sentinel-2 \ac{L1C} plus a \texttt{label} band across nine categories (water, trees, grass, flooded\_vegetation, crops, shrub\_and\_scrub, built, bare, snow\_and\_ice; Fig.~\ref{fig:overview},~\ref{fig:coverage_area}). We retain only the temporal mode composite of the \texttt{label} band. Prior work shows that augmenting \ac{EO} inputs with \ac{LULC} improves \ac{AGB} estimation by adding ecological and phenological context \cite{Avitabile_2012, Sialelli_2025}, motivating its inclusion as an auxiliary context layer for structure--biomass modeling.

\subsubsection{AlphaEarth Embeddings}
\label{subsubsec:modality6}

We include AlphaEarth Foundations embeddings (\acs{AEX}) \cite{brown2025alphaearthfoundationsembeddingfield} as a compact latent predictor alongside the raw modalities. AlphaEarth assimilates multi-source observations across space, time, and measurement types into a shared representation, with each 10\,m annual pixel stored as a 64-dimensional embedding summarizing surface conditions over one calendar year \cite{aex}. Unlike reflectance, backscatter, or elevation bands, these dimensions are coordinates in a learned feature space and are not intended to be interpreted independently. Recent work shows \ac{AEX} carries useful forest-structure and biomass information for canopy-height and \ac{AGB} modeling \cite{hamoudzadeh2026inferringheightearthembeddings, ngo_forest, rs18030436, pascual}, though it does not uniformly outperform strong domain-specific baselines \cite{lucero2026spectral}. We therefore treat \ac{AEX} as a first-class benchmark modality while retaining the raw optical, \ac{SAR}, \ac{DEM}, and \ac{LULC} predictors for sensor-specific interpretability. At the time of dataset creation, the \ac{AEX} \ac{GEE} collection spanned 2021--2024, so we retained the 2021 layer (the midpoint of our 2019--2023 benchmark window). We downloaded all 64 bands \texttt{A00}--\texttt{A63} (Table~\ref{tab:modality_summary}) for our study area, reprojected to 20\,m.

\subsubsection{GEDI RH and AGBD}
\label{subsubsec:modality7}

Target variables are derived from \ac{GEDI} spaceborne lidar \cite{Dubayah_2020}: \ac{RH} profiles from \ac{L2A} \cite{Dubayah_2021_GEDI_L2A_V002} and \ac{AGBD} from \ac{L4A} \cite{DUNCANSON2022112845, Kellner_2022}. The \ac{RH} profile is a 101-dimensional vector (RH0--RH100), where each element records the height below which a given fraction of returned waveform energy is accumulated, encoding the full vertical distribution of canopy material within a footprint. \ac{AGBD} (Mg\,ha$^{-1}$) is derived from \ac{RH} metrics through regionally calibrated allometric models.

We retrieved \ac{GEDI} L2A-B and L4A-C Version~2 data with the \texttt{gediDB} toolbox \cite{Besnard2025} over the benchmark period (1~April 2019 to 30~March 2023), retaining only power beams with sensitivity in $[0.9, 1.0]$ for their higher signal-to-noise ratio relative to coverage beams and to exclude low-quality returns \cite{CHO2025100221}. Per footprint we extracted the 101-element \ac{RH} profile and the scalar \ac{AGBD}. To produce spatially continuous targets, footprints were gridded onto the common 20\,m \ac{HLS}-\ac{MGRS} grid (5490$\times$5490 pixels per tile): per cell, qualifying footprints were averaged to a mean \ac{RH} profile and mean \ac{AGBD}, yielding a 102-band GeoTIFF per tile with \texttt{NODATA}=$-9999.0$ where no footprints fell. The rasters preserve \ac{GEDI}'s sparse sampling pattern and serve as pixel-level supervision.

\subsection{Dataset Creation and Overall Statistics}
\label{subsec:biomazon2}

\begin{table*}[!t]
\centering
\caption{Modality availability by split (number of samples with each modality present).}
\label{tab:modality_availability}
\renewcommand{\arraystretch}{1.15}
\setlength{\tabcolsep}{6pt}
\begin{tabular*}{\textwidth}{@{\extracolsep{\fill}} lrrrrrrrrrr @{}}
\hline
Split & Samples & All req. present & SEN2 & AEX & GLODEM & LULC & SEN1 Desc & SEN1 Asc & ALOS Desc & ALOS Asc \\
\hline
Train & 953{,}076 & 734{,}955 & 953{,}076 & 953{,}076 & 953{,}076 & 953{,}076 & 887{,}225 & 561{,}023 & 792{,}901 & 214{,}299 \\
Val   & 127{,}045 & 127{,}045 & 127{,}045 & 127{,}045 & 127{,}045 & 127{,}045 & 127{,}045 & 61{,}294  & 127{,}045 & 16{,}987  \\
Test  & 249{,}276 & 249{,}276 & 249{,}276 & 249{,}276 & 249{,}276 & 249{,}276 & 249{,}276 & 134{,}224 & 249{,}276 & 36{,}017  \\
\hline
Total & 1{,}329{,}397 & 1{,}111{,}276 & \multicolumn{8}{c}{--} \\
\hline
\end{tabular*}
\end{table*}

\begin{figure}[!t]
   \centering
    \includegraphics[width=\columnwidth]{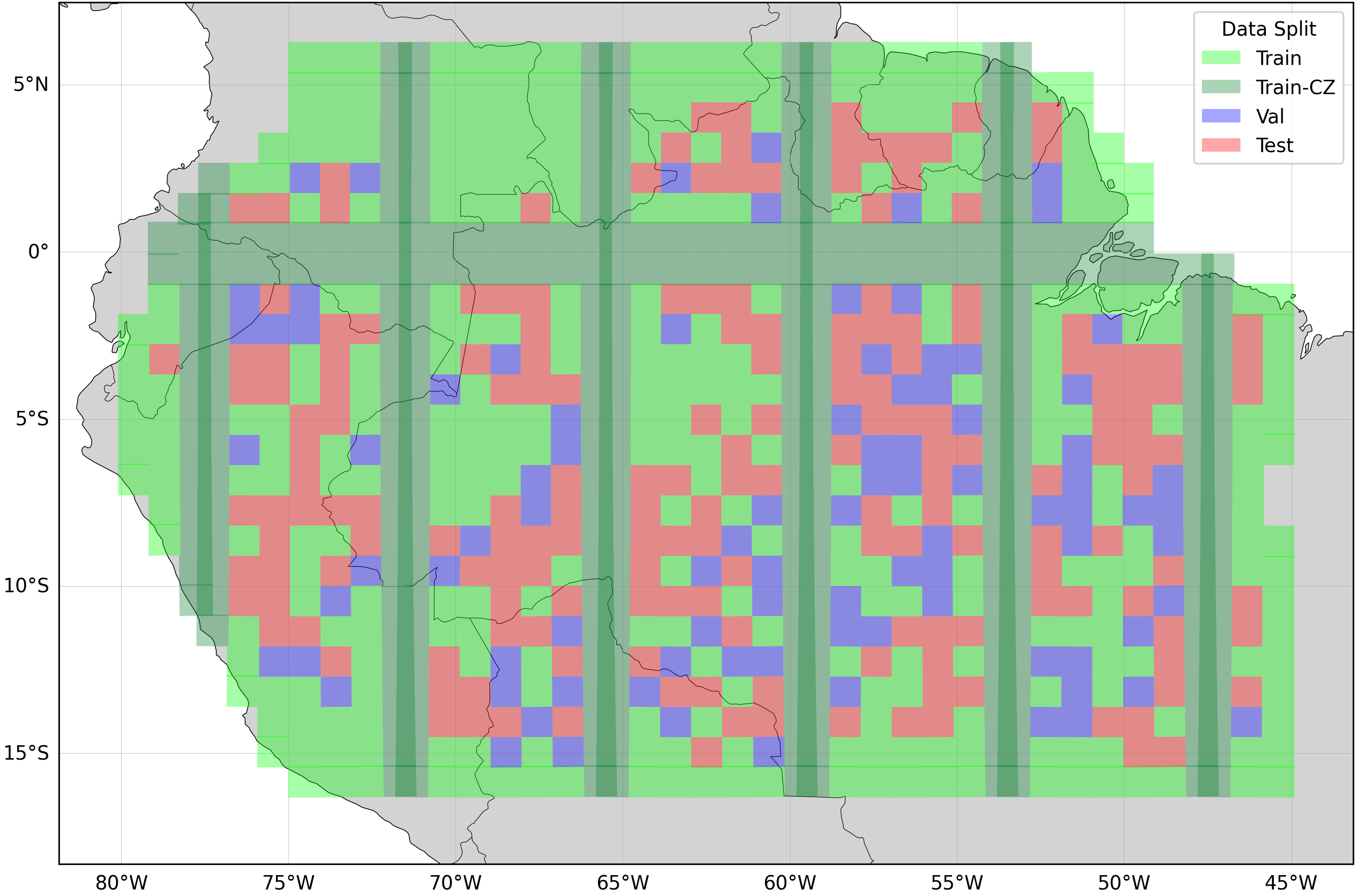}
    \caption{Train-Val-Test split of \emph{Biomazon}. Splitting is done at tile-level after removal of tile-overlap. Light green: train, dark-green: train tiles lying in cross-zone showing uneven overlapping, dark-blue: val tiles, red: test tiles.}
    \label{fig:train-val-test-splits}
\end{figure}

\begin{figure*}[!t]
   \centering
    \includegraphics[width=\textwidth]{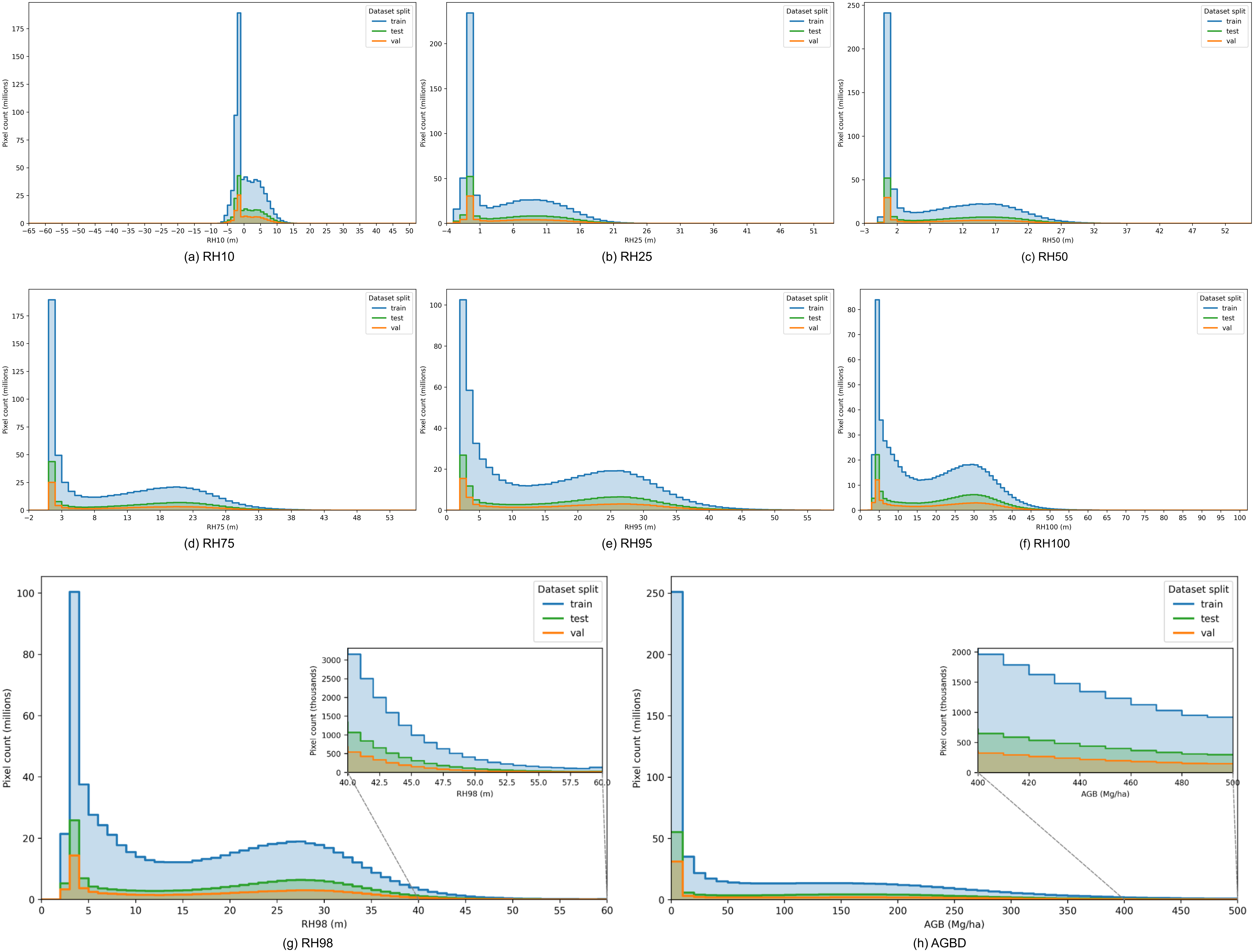}
    \caption{GEDI target distributions in the \emph{Biomazon} dataset. Expanded tail views are shown for \acs{RH98} and \acs{AGBD}, the predominant scalar targets in existing forest-structure and biomass datasets.}
    \label{fig:distributions}
\end{figure*}

To construct the final benchmark dataset, we tile and patch all co-registered modalities and targets across the Amazon Basin. The study area is covered by 913 \ac{HLS}-\ac{MGRS} grid tiles, each spanning $5490\times5490$ pixels at 20\,m resolution ($\approx$109.8$\times$109.8\,km). For each tile, we extract overlapping patches of $256\times256$ pixels ($=$5.12$\times$5.12\,km) with a stride of 128\,pixels (50\% overlap). A patch is retained only if at least 10 valid pixels are present in either the \ac{RH} or \ac{AGBD} band of the \ac{GEDI} raster, ensuring a minimum amount of lidar supervision per patch. Because the target distributions are skewed, extreme values lie in sparsely sampled tail regions relative to the central range (Fig.~\ref{fig:distributions}), hence we filter them during patch extraction to improve training robustness: \ac{RH} shots are first filtered by \acs{RH98} $\in [0, 60]$\,m and then by \acs{RH25} $\geq -3.5$\,m, and \ac{AGBD} is filtered separately to $[0, 500]$\,Mg\,ha$^{-1}$. Out-of-range target pixels are set to NaN and excluded from loss computation. The \acs{RH98} cap of 60\,m is well above the canopy of most Amazonian stands --- the tallest confirmed Amazonian tree is an 88.5\,m angelim (\emph{Dinizia excelsa}), with tallest individuals of Brazil nut (\emph{Bertholletia excelsa}), sumaúma (\emph{Ceiba pentandra}), piquiá (\emph{Caryocar villosum}), and angelim (\emph{Dinizia excelsa}) reaching 50--60\,m \cite{tallest_tree}. The cap therefore excludes a small number of genuine extreme-height returns in favor of removing a larger population of waveform-processing artifacts. The \acs{RH25} floor of $-3.5$\,m corresponds to the 2nd percentile of the negative \acs{RH25} distribution across all 296.7 million valid \ac{GEDI} shots (44.6\% of which report negative \acs{RH25}, with median $-1.26$\,m and mean $-1.36$\,m), retaining plausible negative returns while discarding extreme artifacts. \ac{AGBD} is capped at 500\,Mg\,ha$^{-1}$ following \cite{Sialelli_2025, young2026interpolationgedibiomassestimates}. Each patch stores a fixed 88-band tensor of the predictor modalities (Table~\ref{tab:modality_summary}) together with per-band validity masks recording which bands are complete.

Splitting is performed at tile level to prevent spatial leakage (Fig.~\ref{fig:train-val-test-splits}) via the procedure given in List~\ref{list:tile_split}.

% \refstepcounter{paperlist}\label{list:tile_split}List~\thepaperlist: Tile-level splitting procedure.
\begin{paperlist}[!t]
\caption{Tile-level splitting procedure.}\label{list:tile_split}
\begin{enumerate}
    \item \label{step:tile_trim} \textit{Tile trimming:} Because adjacent \ac{HLS}-\ac{MGRS} tiles share an overlap strip of $\approx$4.9\,km per side, we first trim each tile to its midline-partitioned footprint so that no ground location appears in more than one tile.
    \item \label{step:cross_zone} \textit{Cross-zone exclusion:} Tiles that straddle two UTM zones are kept exclusively in the training set, since their uneven, zone-dependent overlaps make clean spatial separation difficult.
    \item \label{step:required_bands} \textit{Required modalities:} We designate Sentinel-2, \ac{AEX}, \ac{DEM}, \ac{LULC}, Sentinel-1 descending, and ALOS-2 descending as \textit{required} modalities as they achieve near-complete coverage (Fig.~\ref{fig:coverage_area}).
    \item \label{step:val_test_split} \textit{Val/test eligibility:} Among the remaining single-zone tiles, those where every required band has $\geq$80\% valid pixels are randomly assigned to either the validation or test set. All other tiles form the training set, which consequently contains a mixture of fully and partially observed patches (Table~\ref{tab:modality_availability}).
\end{enumerate}
\end{paperlist}

To ensure that all models, whether fixed-channel-input (e.g., Prithvi \cite{szwarcman2025prithvi}) or variable-channel-input (e.g., ChannelViT \cite{bao2024channelvisiontransformersimage}), are evaluated on the same set of \ac{GEDI} targets, validation and test patches are retained only when all required modalities (List~\ref{list:tile_split} Step~\ref{step:required_bands}) are jointly valid. The \ac{GEDI} statistics are therefore identical for these splits in Appendix~\ref{sec:app_stats} (Table~\ref{tab:gedi_stats} and Table~\ref{tab:gedi_stats_extras}), while training-set statistics differ as it includes partially observed patches. The dataset comprises 1,329,397 patches (953,076 train, 127,045 val, 249,276 test), of which 1,111,276 have all required modalities present (Table.~\ref{tab:modality_availability}). Sentinel-2, \ac{AEX}, GLO-30 \ac{DEM}, and Dynamic World \ac{LULC} are available for every patch, while ascending-orbit \ac{SAR} shows the lowest coverage due to limited orbit availability. A detailed per-split breakdown of \ac{GEDI} target distributions and the associated descriptive statistics is provided in Appendix~\ref{sec:app_stats}.

\section{Methodology}
\label{sec:methodology}

\begin{figure*}[!t]
   \centering
    \includegraphics[width=\textwidth]{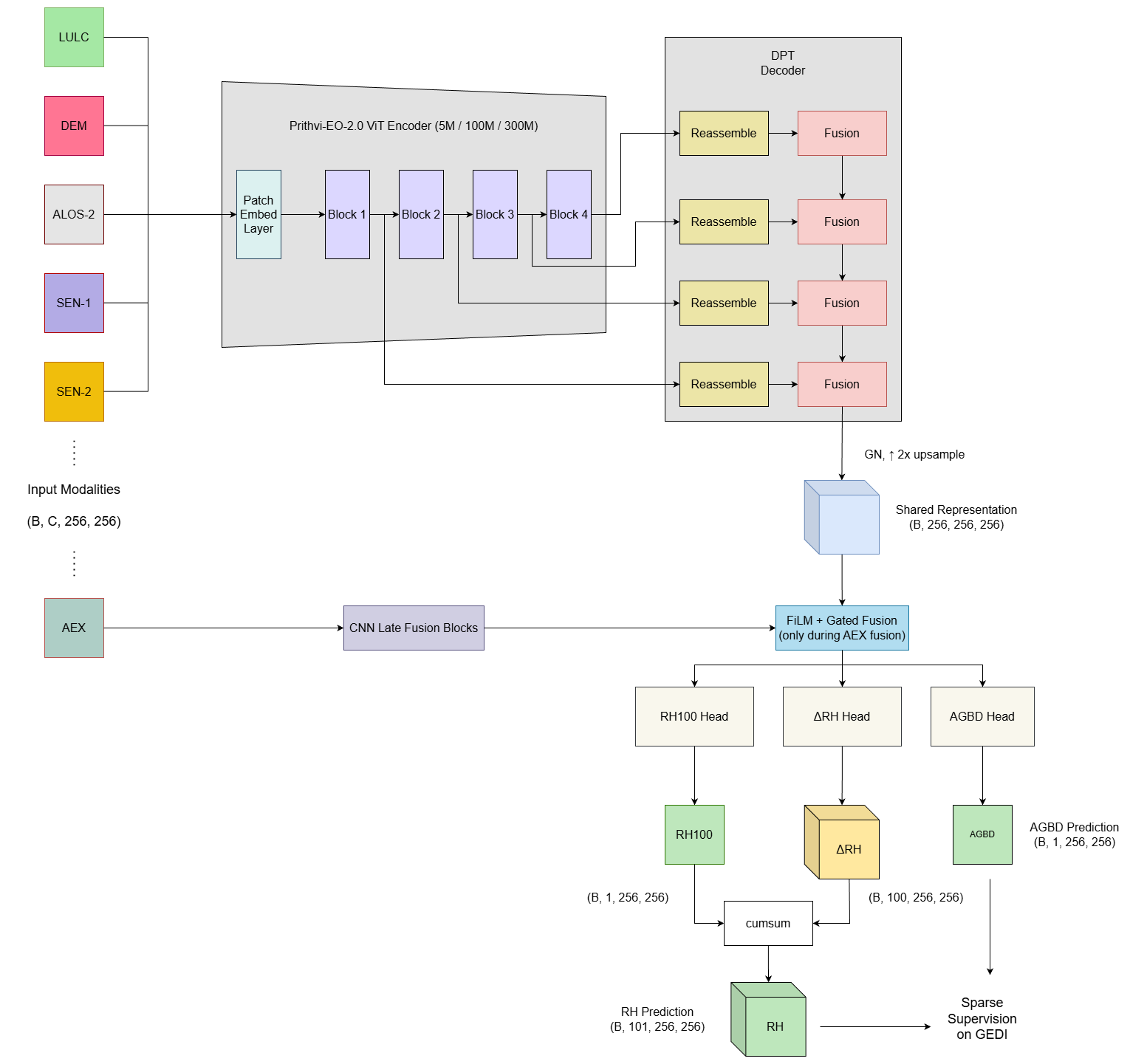}
    \caption{Proposed architecture with \ac{AEX} fusion (Config~8). The decoder heads are detailed in Fig.\ref{fig:aex_base_architecture}.}
    \label{fig:model_architecture}
\end{figure*}

This section describes the baseline framework used to benchmark \emph{Biomazon}. The overall pipeline follows a shared encoder-decoder architecture with task-specific prediction heads for \ac{RH}-profile and \ac{AGBD} estimation (Fig.~\ref{fig:model_architecture}), fine-tuned end-to-end on the multimodal patches described in Sec.~\ref{sec:biomazon}. We additionally include a lightweight \ac{CNN} baseline that operates solely on AlphaEarth embeddings.

\subsection{Baseline Architecture}
\label{subsec:methodology1}

\begin{figure}[!t]
   \centering
    \includegraphics[width=\columnwidth]{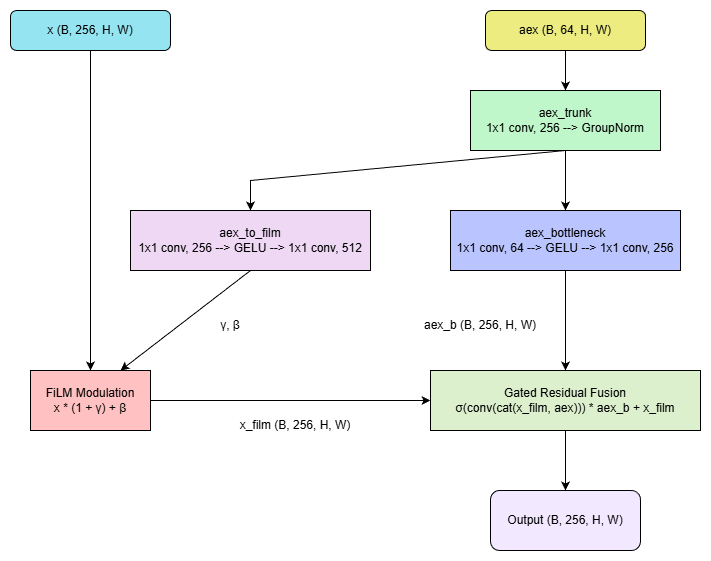}
    \caption{Proposed AlphaEarth embeddings fusion architecture in Config~8.}
    \label{fig:aex_fusion_architecture}
\end{figure}

\begin{figure}[!t]
   \centering
    \includegraphics[width=\columnwidth]{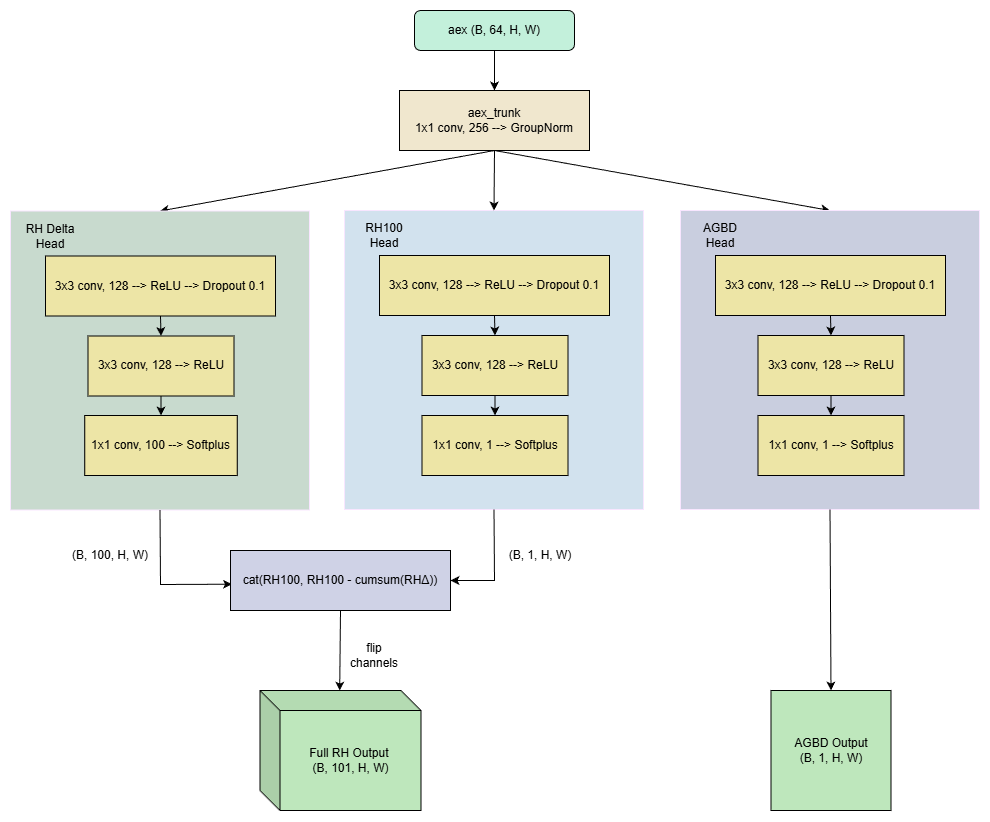}
    \caption{Proposed AlphaEarth embeddings Base architecture (Config~0). Referred to as \acs{AEXB} or \acs{AEX-Base} in this work.}
    \label{fig:aex_base_architecture}
\end{figure}

\subsubsection{Encoder}
\label{subsubsec:encoder}

We adopt the Prithvi-EO-2.0 \cite{szwarcman2025prithvi} family of \acp{ViT} \cite{dosovitskiy2020vit} as the encoder backbone. Prithvi-EO-2.0 is a geospatial foundation model pretrained on multi-temporal \ac{HLS} imagery with \ac{TL} coordinate encodings via a masked autoencoding objective. We evaluate three model scales --- Prithvi-5M (192 embedding dimensions, 12 blocks, 3 heads), Prithvi-100M (768, 12, 12), and Prithvi-300M (1024, 24, 16) --- all initialized from the Prithvi-EO-2.0 \ac{TL}-pretrained checkpoints. We do not supply per-sample temporal--location coordinates at training or inference (our inputs are multi-year composites), so the \ac{TL} embedding layers are frozen as non-trainable buffers. We omit Prithvi-600M because its 14$\times$14 patch size does not evenly divide our 256$\times$256 patch (256/14 is non-integer), yielding an incompatible token grid.

The encoder processes each input patch as a single-frame 5D tensor of shape (B,C,1,256,256), with temporal dimension $T=1$ since our inputs are temporal composites. Spatial tokenization uses a patch size of 16$\times$16, yielding 16$\times$16=256 spatial tokens plus one \texttt{[CLS]} token. The input-channel count \texttt{C} varies by modality configuration: for the full configuration (Table~\ref{tab:test_rmse_ablation} Configs 7,8) we concatenate Sentinel-2 (10 bands), Sentinel-1 descending (\texttt{VV}, \texttt{VH}, \texttt{VV}/\texttt{VH} ratio; 3 bands), ALOS-2 descending (\texttt{HH}, \texttt{HV}, \texttt{HH}/\texttt{HV} ratio; 3 bands), \ac{DEM} (1 band), and one-hot \ac{LULC} (9 classes), giving $C=26$. Because Prithvi's pretrained patch-embedding layer expects a fixed six-channel input (the \ac{HLS} bands), it cannot directly accept variable-channel inputs. For each configuration we therefore instantiate a new patch-embedding layer with the required input channels \texttt{C}. The pretrained Prithvi patch-embedding weights are then copied into the channels corresponding to the six \ac{HLS}-aligned bands (Blue, Green, Red, \acs{NIR}, \acs{SWIR}1, \acs{SWIR}2), and the remaining channels are initialized from scratch. Throughout this paper, ``S2-HLS'' denotes the six \ac{HLS}-aligned Sentinel-2 bands (Config~1) and ``S2'' denotes the full 10-band Sentinel-2 stack (Config~2 onward).

Intermediate feature maps are extracted at four evenly spaced encoder layers. For Prithvi-5M and Prithvi-100M, these are layers \{2, 5, 8, 11\}. For Prithvi-300M, they are layers \{5, 11, 17, 23\}. At each extraction point, the \texttt{[CLS]} token is separated from the spatial tokens, and the spatial tokens are reshaped ("unpatchified") from the flat sequence back to a 2D feature map of shape (B,D,16,16), where \texttt{D} is the embedding dimension.

\subsubsection{Decoder}
\label{subsubsec:decoder}

The four multi-scale feature maps from the encoder are decoded using a \ac{DPT} decoder \cite{DPTHead}, which turns the \ac{ViT}'s patch tokens into a dense feature map via two stages: a \texttt{Reassemble} stage that reshapes tokens into 2D maps at four scales, and a fusion stage that merges them coarse-to-fine.

\texttt{Reassemble} first injects the encoder's global \texttt{[CLS]} token into each spatial token so that every patch carries both its local features and the global image context: \texttt{[CLS]} is broadcast to all spatial positions, concatenated channel-wise with the spatial tokens, and projected back to the original feature dimension via a linear layer with GELU activation \cite{hendrycks2016gaussian}. Each level is then mapped to post-process channels [128, 256, 512, 1024] via 1$\times$1 convolutions and resized to a distinct scale (4$\times$ and 2$\times$ transposed-convolution upsamples for levels~1--2, identity for level~3, stride-2 convolution for level~4). A 3$\times$3 convolution per level finally projects all four to a common 256 channels.

The fusion stage merges this pyramid coarse-to-fine. Starting from the coarsest level, the first \texttt{Feature Fusion Block} applies a pre-activation residual convolution unit (two 3$\times$3 convolutions, ReLU pre-activation) to the level-4 feature, bilinearly upsamples 2$\times$, and applies a 1$\times$1 projection at 256 channels (there is no previous-block output to fuse with). Each subsequent block first refines the current-level feature with a pre-activation residual convolution unit, adds it to the upsampled previous-block output, refines the sum with a second residual unit, bilinearly upsamples 2$\times$, and applies the same 1$\times$1 projection. We use the \texttt{expand=False} configuration of \ac{DPT}, in which the channel width is held constant at 256 across fusion blocks. In the alternative \texttt{expand=True} configuration the projection would instead halve the channel width at each fusion step. After four blocks, a final 3$\times$3 convolution with ReLU activation yields a 256-channel feature map at 128$\times$128. GroupNorm (32 groups) and a bilinear 2$\times$ upsample recover the full 256$\times$256 resolution, giving the shared representation that is fed to the task-specific \ac{RH} and \ac{AGBD} heads.

\subsubsection{AlphaEarth Late Fusion}
\label{subsubsec:late_fusion}

When AlphaEarth embeddings are used in late-fusion mode \texttt{(aex\_l)}, the 64-dimensional \ac{AEX} feature map is injected into the shared decoder representation after upsampling, via a two-stage fusion module. First, a \ac{FiLM} conditioning stage \cite{Perez_2018} projects the \ac{AEX} features through a 1$\times$1 convolution trunk with GroupNorm into 256 channels, then predicts per-channel scale ($\gamma$) and shift ($\beta$) parameters through a two-layer 1$\times$1 convolution with GELU activation, applying the affine modulation $\mathbf{x}' = \mathbf{x} \odot (1 + \gamma) + \beta$. The \ac{FiLM} parameters are zero-initialized so that the initial modulation is an identity transformation. Second, a \textit{gated residual fusion} computes a bottleneck residual from the \ac{AEX} trunk features (two 1$\times$1 convolutions with GELU, bottleneck dimension 64, near-zero initialization), and blends it with the \ac{FiLM}-modulated features via a learned sigmoid gate initialized at 0.5 (zero-initialized weights): $\mathbf{x}_{\text{out}} = \mathbf{x}' + g \odot \mathbf{aex\_b}$, where $g = \sigma(\text{Conv}([\mathbf{x}', \mathbf{aex\_b}]))$ and \texttt{aex\_b} is the output of the bottleneck layer, as shown in Fig.~\ref{fig:aex_fusion_architecture}. Together, the bottleneck and near-zero initialization limit the capacity and initial contribution of the \ac{AEX} late-fusion branch, aiming to reduce the risk of an overly direct shortcut-like pathway to the output and to encourage the gradual learning of task-relevant complementary modulations of the shared decoder representation.

\subsection{RH Prediction Head}
\label{subsubsec:rhhead}

\subsubsection{Problem formulation}
The \ac{RH} prediction head is designed to predict, for each pixel, the full \ac{RH} profile as a discrete quantile function over percentiles $\{0,1,\dots,100\}$. Formally, given a multimodal input patch $x$, the \ac{RH} head outputs a vector-valued function $\hat{\mathbf{r}}(x) \in \mathbb{R}^{K}$ with $K=101$, interpreted as an estimate of the conditional quantiles
\begin{equation}
\hat{\mathbf{r}}(x) \approx \big[ Q_{0}(H \mid x), \, Q_{1}(H \mid x), \, \dots, \, Q_{100}(H \mid x) \big],
\end{equation}
where $Q_{p}(H \mid x)$ denotes the $p$-th percentile of the vertical height distribution at that pixel. By definition of a quantile function, $\hat{\mathbf{r}}(x)$ must satisfy the shape constraint
\begin{equation}
\hat r_0(x) \le \hat r_1(x) \le \cdots \le \hat r_{100}(x),
\end{equation}
so the \ac{RH} head must enforce monotonicity across the percentile dimension. In the following, we review two monotone-by-construction parameterizations for dense \acs{RH} profiles and motivate our choice based on the empirical support of the training labels.

\subsubsection{Anchored Monotone Parameterization for Dense Relative Height Profiles}
\label{subsubsec:anchored_monotone}

Let $x$ denote the multimodal input patch and, for each pixel, let the target relative height (RH) profile be the vector $\mathbf{r}(x) \in \mathbb{R}^{K}$ with $K=101$, corresponding to percentiles $\{0,1,\dots,100\}$ and ordered as $\mathbf{r} = [r_0, r_1, \dots, r_{100}]$. The prediction task is to estimate a dense monotone quantile function $\hat{\mathbf{r}}(x)$ satisfying
\begin{equation}
\hat r_0(x) \le \hat r_1(x) \le \cdots \le \hat r_{100}(x),
\label{eq:monotone_constraint}
\end{equation}
while matching the empirical support of the \acs{RH} labels.

\paragraph{One-shot cumulative parameterization.}
A naive monotone-by-construction strategy for ordered outputs predicts nonnegative increments $\boldsymbol{\delta}(x) \in \mathbb{R}^{K}$ and defines
\begin{equation}
\boldsymbol{\delta}(x) = \operatorname{softplus}(\mathbf{u}(x)) \succeq \mathbf{0},
\qquad 
\hat{\mathbf{r}}(x) = \operatorname{cumsum}\!\left(\boldsymbol{\delta}(x)\right),
\label{eq:oneshot_cumsum}
\end{equation}
which enforces \eqref{eq:monotone_constraint} since $\hat r_{i+1}(x) - \hat r_i(x) = \delta_{i+1}(x) \ge 0$ for all $i$. However, \eqref{eq:oneshot_cumsum} also induces the support constraint
\begin{equation}
\hat r_i(x) = \sum_{j \le i} \delta_j(x) \ge 0 \qquad \forall i,
\label{eq:nonneg_support}
\end{equation}
because all increments are nonnegative. In \ac{GEDI}-derived \ac{RH} profiles, the empirical distribution exhibits negative values at low percentiles (i.e., $\exists\, p \le 30$ such that $r_p < 0$) due to ground reference uncertainty, geolocation error, and waveform noise. Under such signed support, any estimator constrained by \eqref{eq:nonneg_support} is mis-specified: for any index $i$ with $r_i^\ast < 0$, the pointwise error satisfies
\begin{equation}
|\hat r_i - r_i^\ast| \ge |r_i^\ast|,
\label{eq:irreducible_bias}
\end{equation}
so the nonnegativity constraint introduces an irreducible lower bound on the attainable loss and systematically biases the lower tail of the profile. Prior non-crossing and monotone quantile formulations avoid this pathology by adding a free offset to the cumulative increments rather than summing from zero: \cite{10.5555/3495724.3497058} apply an affine map $q_i = \alpha\,\psi_i + \beta$ to a cumulative sum $\psi_i$, with non-negative scale $\alpha$ and a free additive offset $\beta$, and \cite{Zhou_2021} add cumulative nonnegative increments to an unconstrained baseline $\Delta_0$. Both therefore admit signed support. Our contribution is not the free offset itself, but a top-anchored variant specialized to \ac{RH} profiles, which predicts a free top anchor and cumulative nonnegative drops toward the lower percentiles, and which we apply densely per pixel to match the signed empirical support of \ac{GEDI} \ac{RH} labels.

\paragraph{Anchored monotone parameterization (our approach)}
Our formulation is a task-specific adaptation of monotone-by-construction ordered-output parameterizations, specialized here to \ac{RH} profiles whose lower percentiles can be negative in the training data. To preserve monotonicity while allowing unrestricted lower support, we adopt an anchored formulation that predicts a free top anchor and cumulative nonnegative drops from the top of the profile. Specifically, we predict
\begin{equation}
\hat r_{100}(x) = \operatorname{softplus}(a(x)), 
\qquad 
\mathbf{d}(x) = \operatorname{softplus}(\mathbf{v}(x)) \in \mathbb{R}_{+}^{K-1},
\end{equation}
and define:
\begin{equation}
\hat r_{100-j}(x) = \hat r_{100}(x) - \sum_{t=1}^{j} d_t(x), 
\qquad j = 1,\dots,100,
\label{eq:anchored_cumsum}
\end{equation}
with $\hat{\mathbf{r}}(x) = [\hat r_0(x), \dots, \hat r_{100}(x)]$. Monotonicity follows directly since for all $j = 0,1,\dots,99$,
\begin{equation}
\hat r_{100-(j+1)}(x) - \hat r_{100-j}(x) = - d_{j+1}(x) \le 0,
\end{equation}
which is equivalent to Eq.~\ref{eq:monotone_constraint} after reindexing to ascending percentiles. Unlike Eq.~\ref{eq:oneshot_cumsum}, the anchored construction does not impose a lower support constraint: the lowest percentile satisfies
\begin{equation}
\hat r_0(x) = \hat r_{100}(x) - \sum_{t=1}^{100} d_t(x) \in \mathbb{R},
\end{equation}
so the model admits negative \ac{RH} values whenever the cumulative drop exceeds the anchor. This parameterization therefore enforces the shape constraint (Eq.~\ref{eq:monotone_constraint}) while matching the signed empirical support of the \ac{RH} labels and avoiding the bias implied by Eq.~\ref{eq:nonneg_support}.

\paragraph{Dense per-pixel prediction.}
The anchored parameterization Eq.~\ref{eq:anchored_cumsum} is applied independently at each spatial location, yielding a dense monotone vector $\hat{\mathbf{r}}(x) \in \mathbb{R}^{101}$ per pixel. This formulation defines a valid monotone quantile function with free vertical offset, and is statistically consistent with the observed support of \ac{GEDI} \ac{RH} profiles.

\subsubsection{Network realization}
\label{subsubsec:network_realization}

The \ac{RH} head is split into two parallel sub-heads operating on the shared 256-channel decoder feature map. The \textit{anchor head} predicts the scalar \acs{RH100} value per pixel through two stacked $3{\times}3$ convolution layers (256$\to$128$\to$128 channels, each followed by ReLU, with 10\% Dropout2d after the first), a $1{\times}1$ projection to 1 channel, and a Softplus activation ensuring $\hat{r}_{100} > 0$. The strict positivity of the anchor is consistent with the filtered \emph{Biomazon} \acs{RH100} distribution (dataset minimum 0.03\,m; Appendix~\ref{sec:app_stats} Table~\ref{tab:gedi_stats}), but assumes \acs{RH100} is strictly positive and would need revisiting for datasets retaining near-zero negative \acs{RH100} values (e.g., bare ground). The \texttt{delta head} has an identical architecture but projects to 100 output channels with Softplus, producing the nonnegative drops $\mathbf{d}(x)$. The full 101-element \ac{RH} profile is then assembled via the anchored cumulative sum (Eq.~\ref{eq:anchored_cumsum}) and flipped to ascending percentile order.

\subsection{AGBD Prediction Head}
\label{subsubsec:agbdhead}

The \acs{AGBD} head follows the same convolutional architecture as the \acs{RH} anchor head: two $3{\times}3$ convolution layers (256$\to$128$\to$128 channels with ReLU) with 20\% Dropout2d after the first layer, a $1{\times}1$ projection to a single output channel, and a Softplus activation to enforce non-negativity ($\hat{y}_{\text{AGBD}} > 0$\,Mg\,ha$^{-1}$). The higher dropout rate relative to the \acs{RH} head (20\% vs.\ 10\%) reflects the greater label noise and distributional skew in the \acs{AGBD} targets (Sec.~\ref{subsec:biomazon2}).

\subsection{AlphaEarth Shallow Network Baseline}
\label{subsubsec:aexbase}

To isolate the predictive content of AlphaEarth embeddings from the encoder--decoder architecture, we include a lightweight \ac{CNN} baseline as depicted in Fig.~\ref{fig:aex_base_architecture}, that operates solely on the 64-band \ac{AEX} input at full $256{\times}256$ resolution. The model consists of a shared trunk: a single $1{\times}1$ convolution projecting from 64 to 256 channels followed by GroupNorm (32 groups), feeding the same task-specific heads used in the main architecture (Sec.~\ref{subsubsec:rhhead}--\ref{subsubsec:agbdhead}). This per-pixel design has no spatial downsampling or multi-scale processing, so it tests how much structure and biomass information is recoverable from \ac{AEX} alone without the inductive biases of a \ac{ViT} encoder or \ac{DPT} decoder. This architecture is referred to as \acs{AEXB} or \acs{AEX-Base} throughout this paper.

\subsection{Loss Function}
\label{subsubsec:loss}

We train all configurations with a sparse Huber loss that handles the irregular sampling pattern of \ac{GEDI} targets. Because \ac{GEDI} footprints cover only a small fraction of each $256{\times}256$ patch, most target pixels are marked as NaN and excluded from loss computation. Only pixels with valid lidar observations contribute. For each valid pixel, predictions and targets are first z-score normalized using precomputed per-channel training-set statistics (mean and standard deviation for each of the 101 \ac{RH} percentiles and for \ac{AGBD} independently). The normalized residuals are then passed through a Huber loss \cite{huber1964robust} with transition point $\delta = 1.345$:
\begin{equation}
L_{\delta}(z) = \begin{cases} \tfrac{1}{2}\,z^{2} & \text{if } |z| \leq \delta, \\ \delta\,(|z| - \tfrac{1}{2}\,\delta) & \text{otherwise,} \end{cases}
\label{eq:huber}
\end{equation}
where $z = \hat{z} - z^{*}$ is the normalized prediction error. The choice of $\delta = 1.345$ corresponds to the 95\% asymptotic efficiency point of the Huber estimator under Gaussian errors \cite{Holland01011977}, providing robustness to outliers while retaining near-optimal efficiency for well-behaved residuals.

Because both \ac{RH} and \ac{AGBD} targets are heavily right-skewed (Sec.~\ref{subsec:biomazon2}, Fig.~\ref{fig:distributions}), we apply \ac{LDS} \cite{lds} to reweight each pixel's loss contribution by the inverse of its smoothed label-bin density, upweighting rare tail values and downweighting common ones. Weight tables are precomputed from training-set histograms: for \ac{RH}, per-percentile tables (percentiles 25--98) with 1\,m bins and for \ac{AGBD}, 50 bins of width 10\,Mg\,ha$^{-1}$ spanning 0--500\,Mg\,ha$^{-1}$. Bin counts are reweighted by the inverse square root and smoothed with a Gaussian kernel ($k{=}5$, $\sigma{=}2$), then normalized to unit mean. At training time, each valid pixel's Huber loss is multiplied by its \ac{LDS} weight. \ac{RH} channels outside the percentile range covered by the precomputed tables (RH0--RH24 and RH99--RH100) instead receive a fixed weight of 0.5, preventing these noisy extreme percentiles from disproportionately driving the gradient without requiring an explicit density model for them. At validation and test time all weights are set to unity.

For the joint \acs{RH}--\acs{AGBD} task, the total loss is the unweighted sum of the per-task losses:
\begin{equation}
\mathcal{L} = \mathcal{L}_{\text{RH}} + \mathcal{L}_{\text{AGBD}},
\end{equation}
where $\mathcal{L}_{\text{RH}}$ is the mean Huber loss over all valid pixels and all 101 \acs{RH} channels, and $\mathcal{L}_{\text{AGBD}}$ is the mean Huber loss over valid \acs{AGBD} pixels. Because both terms are first averaged over their respective valid pixels and channels, each contributes a single scalar to the total and the two tasks contribute comparably to the joint gradient, so the 101-channel \ac{RH} loss does not dominate the single-channel \ac{AGBD} loss.

\subsection{Implementation Details}
\label{subsec:implementation}

\subsubsection{Input preprocessing}
\label{subsubsec:input_processing}
Each modality undergoes band-specific normalization before being stacked into the model input tensor. Sentinel-2 reflectance bands are z-score normalized using precomputed training-set means and standard deviations. For Sentinel-1 and ALOS-2, the VV and VH (respectively HH and HV) backscatter bands are z-score normalized, and a co-pol/cross-pol ratio band is appended (VV$-$VH or HH$-$HV in dB scale) and separately normalized, yielding three channels per \acs{SAR} source. The GLO-30 \acs{DEM} band is first transformed via the inverse hyperbolic sine function, $\operatorname{arcsinh}(\text{DEM}\,/\,s)$ with IQR-derived scale $s = 278.18$, to compress the heavy-tailed elevation distribution, and then z-score normalized. The Dynamic World \acs{LULC} label is one-hot encoded into 9 binary channels. AlphaEarth embeddings (64 bands) are used without additional normalization and are passed as a separate input to the late-fusion module (Sec.~\ref{subsubsec:late_fusion}), they are not concatenated with the other bands in the encoder input.

\subsubsection{Data augmentation}
\label{subsubsec:data_augmentation}
During training, random geometric augmentation is applied with 50\% probability. When triggered, one of three transformations is selected uniformly at random: horizontal flip, vertical flip, or both. The same transformation is applied consistently to all input bands, \acs{AEX} embeddings, and target maps within a patch.

\subsubsection{Optimizer and schedule}
\label{subsubsec:optimizer}
We use AdamW \cite{adamw} with $(\beta_1,\beta_2) = (0.9,\,0.999)$ and weight decay $0.01$, employing differentiated learning rates across three parameter groups: (i)~encoder parameters at $\text{lr}_{\text{enc}} = 5{\times}10^{-5}$, (ii)~decoder and \acs{RH} head parameters at $\text{lr}_{\text{dec}} = 5{\times}10^{-5}$, and (iii)~\acs{AGBD} head parameters at $\text{lr}_{\text{agb}} = 2{\times}10^{-5}$. The lower \acs{AGBD} learning rate reflects the noisier and more heavily skewed nature of the biomass targets. The learning rate schedule consists of a linear warmup from $0.1{\times}$ the base learning rate over the first 5 epochs, followed by cosine annealing \cite{sgdr} to $\eta_{\min} = 0$ over the remaining 25 epochs, for a total of 30 training epochs. The cosine annealing schedule is:
\begin{equation}
\eta_t = \eta_{\min} + \tfrac{1}{2}(\eta_{\max} - \eta_{\min})\!\left(1 + \cos\!\left(\frac{T_{\text{cur}}}{T_{\max}}\pi\right)\right),
\end{equation}
where $\eta_t$ is the learning rate at epoch $t$, $\eta_{\max}$ is the initial (post-warmup) learning rate, $\eta_{\min}$ is the minimum learning rate, $T_{\text{cur}}$ is the number of epochs elapsed since warmup ended, and $T_{\max} = 25$ is the total number of cosine annealing epochs.

\subsubsection{Distributed training}
\label{subsubsec:dist_train}
All models are trained with PyTorch Distributed Data Parallel (DDP) across 16 nodes with 4 GPUs each (64 GPUs total) on a SLURM-managed HPC cluster, using NCCL as the communication backend. We use a per-GPU batch size of 16 and 16 data-loading workers per GPU, yielding an effective batch size of $16 \times 64 = 1{,}024$. Forward passes are computed in bfloat16 mixed precision via \texttt{torch.amp.autocast} for memory efficiency, while losses and gradient updates remain in full precision. The best model checkpoint is selected based on the composite validation score $0.5 \cdot \overline{\text{RMSE\%}}_{\text{RH}} + 0.5 \cdot \text{RMSE\%}_{\text{AGBD}}$, where $\overline{\text{RMSE\%}}_{\text{RH}}$ is the mean \acs{RMSE}\% across the five evaluated \acs{RH} percentiles (\acs{RH25}, \acs{RH50}, \acs{RH75}, \acs{RH95}, \acs{RH98}). Final test-set metrics are computed using the best checkpoint.

\subsubsection{Evaluation metrics.}
\label{subsubsec:eval_metric}
For each target variable $y$ (\acs{RH25}, \acs{RH50}, \acs{RH75}, \acs{RH95}, \acs{RH98}, and \acs{AGBD}), we report seven metrics computed over the $N$ valid (non-NaN) pixels aggregated across the full test set. Let $y_i$ and $\hat{y}_i$ denote the target and prediction for pixel $i$, and let $\bar{y} = \frac{1}{N}\sum_i y_i$.

\noindent\textit{Root Mean Squared Error} (overall prediction accuracy):
\begin{equation}
\text{RMSE} = \sqrt{\tfrac{1}{N}\textstyle\sum_{i=1}^{N}(y_i - \hat{y}_i)^{2}}\,.
\end{equation}

\noindent\textit{Relative RMSE} (error magnitude relative to target mean):
\begin{equation}
\text{RMSE\%} = \tfrac{\text{RMSE}}{|\bar{y}|} \times 100\,.
\end{equation}

\noindent\textit{Mean Absolute Error} (average error magnitude, less sensitive to outliers than RMSE):
\begin{equation}
\text{MAE} = \tfrac{1}{N}\textstyle\sum_{i=1}^{N}|y_i - \hat{y}_i|\,.
\end{equation}

\noindent\textit{Relative MAE} (MAE normalized by target mean):
\begin{equation}
\text{MAE\%} = \tfrac{\text{MAE}}{|\bar{y}|} \times 100\,.
\end{equation}

\noindent\textit{Bias} (systematic over- or under-prediction):
\begin{equation}
\text{Bias} = \tfrac{1}{N}\textstyle\sum_{i=1}^{N}(\hat{y}_i - y_i)\,.
\end{equation}

\noindent\textit{Relative Bias} (Bias normalized by target mean):
\begin{equation}
\text{Bias\%} = \tfrac{\text{Bias}}{|\bar{y}|} \times 100\,.
\end{equation}

\noindent\textit{Coefficient of Determination} (fraction of target variance explained):
\begin{equation}
R^{2} = 1 - \frac{\sum_{i=1}^{N}(y_i - \hat{y}_i)^{2}}{\sum_{i=1}^{N}(y_i - \bar{y})^{2}}\,.
\end{equation}

To assess stability, every configuration is trained with five different random seeds and we report mean $\pm$ standard deviation across runs.

\section{Results and Discussions}
\label{sec:results}

\begin{table*}[!t]
\centering
\caption{Test \acs{RMSE} (mean $\pm$ std over 5 runs) for \acs{RH} percentiles and \acs{AGBD} across modality configurations. S2-HLS denotes the 6 \acs{HLS} specific bands of Sentinel-2 on which Prithvi models were pretrained. S2 denotes all 10 bands of Sentinel-2 that we've considered. \acs{AEX} refers to AlphaEarth embeddings. Config~0 \acs{AEX-Base} denotes the shallow network baseline for \acs{AEX}.}
\label{tab:test_rmse_ablation}
\renewcommand{\arraystretch}{1.15}
\setlength{\tabcolsep}{3pt}

\begin{adjustbox}{width=\textwidth,center}
\begin{tabular}{|c|l|c|c|c|c|c|c|c|c|c|c|c|c|c|}
\hline
\textbf{Config} & \textbf{Encoder} & \textbf{S2-HLS} & \textbf{S2} & \textbf{S1} & \textbf{ALOS} & \textbf{DEM} & \textbf{LULC} & \textbf{AEX} & \textbf{RH25} $\downarrow$ & \textbf{RH50} $\downarrow$ & \textbf{RH75} $\downarrow$ & \textbf{RH95} $\downarrow$ & \textbf{RH98} $\downarrow$ & \textbf{AGBD} $\downarrow$ \\
\hline
\multirow{2}{*}{1} & Prithvi-5M   & \checkmark &  &  &  &  &  &  & 4.08 $\pm$ 0.01 & 5.23 $\pm$ 0.14 & 5.69 $\pm$ 0.03 & 6.15 $\pm$ 0.01 & 6.40 $\pm$ 0.02 & 79.48 $\pm$ 0.22 \\
                  & Prithvi-100M & \checkmark &  &  &  &  &  &  & 4.08 $\pm$ 0.05 & 5.09 $\pm$ 0.14 & 5.58 $\pm$ 0.08 & 6.04 $\pm$ 0.03 & 6.36 $\pm$ 0.05 & 77.87 $\pm$ 0.38 \\
\hline
\multirow{2}{*}{2} & Prithvi-5M   &  & \checkmark &  &  &  &  &  & 4.11 $\pm$ 0.02 & 5.18 $\pm$ 0.13 & 5.65 $\pm$ 0.04 & 6.17 $\pm$ 0.02 & 6.44 $\pm$ 0.06 & 79.68 $\pm$ 0.41 \\
                  & Prithvi-100M &  & \checkmark &  &  &  &  &  & 4.09 $\pm$ 0.02 & 5.13 $\pm$ 0.09 & 5.61 $\pm$ 0.13 & 6.08 $\pm$ 0.04 & 6.40 $\pm$ 0.09 & 78.49 $\pm$ 0.29 \\
\hline
\multirow{2}{*}{3} & Prithvi-5M   &  & \checkmark & \checkmark &  &  &  &  & 4.15 $\pm$ 0.20 & 5.20 $\pm$ 0.15 & 5.64 $\pm$ 0.06 & 6.10 $\pm$ 0.01 & 6.37 $\pm$ 0.03 & 78.83 $\pm$ 0.28 \\
                  & Prithvi-100M &  & \checkmark & \checkmark &  &  &  &  & 4.07 $\pm$ 0.09 & 4.96 $\pm$ 0.04 & 5.55 $\pm$ 0.10 & 6.01 $\pm$ 0.04 & 6.36 $\pm$ 0.10 & 77.48 $\pm$ 0.40 \\
\hline
\multirow{2}{*}{4} & Prithvi-5M   &  & \checkmark &  & \checkmark &  &  &  & 4.09 $\pm$ 0.03 & 5.16 $\pm$ 0.18 & 5.66 $\pm$ 0.08 & 6.09 $\pm$ 0.01 & 6.35 $\pm$ 0.02 & 79.39 $\pm$ 0.27 \\
                  & Prithvi-100M &  & \checkmark &  & \checkmark &  &  &  & 4.11 $\pm$ 0.14 & 5.00 $\pm$ 0.03 & 5.58 $\pm$ 0.08 & 6.01 $\pm$ 0.01 & 6.32 $\pm$ 0.07 & 78.12 $\pm$ 0.48 \\
\hline
\multirow{2}{*}{5} & Prithvi-5M   &  & \checkmark & \checkmark & \checkmark &  &  &  & 4.00 $\pm$ 0.01 & 5.14 $\pm$ 0.15 & 5.58 $\pm$ 0.10 & 6.02 $\pm$ 0.01 & 6.26 $\pm$ 0.03 & 78.47 $\pm$ 0.27 \\
                  & Prithvi-100M &  & \checkmark & \checkmark & \checkmark &  &  &  & 4.09 $\pm$ 0.20 & 5.03 $\pm$ 0.16 & 5.49 $\pm$ 0.05 & 5.95 $\pm$ 0.02 & 6.26 $\pm$ 0.09 & 77.35 $\pm$ 0.53 \\
\hline
\multirow{2}{*}{6} & Prithvi-5M   &  & \checkmark & \checkmark & \checkmark & \checkmark &  &  & 4.09 $\pm$ 0.09 & 5.07 $\pm$ 0.05 & 5.61 $\pm$ 0.08 & 6.02 $\pm$ 0.03 & 6.35 $\pm$ 0.05 & 77.95 $\pm$ 0.47 \\
                  & Prithvi-100M &  & \checkmark & \checkmark & \checkmark & \checkmark &  &  & 4.02 $\pm$ 0.05 & 5.06 $\pm$ 0.14 & 5.47 $\pm$ 0.08 & 5.94 $\pm$ 0.02 & 6.25 $\pm$ 0.06 & 76.99 $\pm$ 0.44 \\
\hline
\multirow{2}{*}{7} & Prithvi-5M   &  & \checkmark & \checkmark & \checkmark & \checkmark & \checkmark &  & 4.00 $\pm$ 0.00 & 5.03 $\pm$ 0.14 & 5.56 $\pm$ 0.04 & 6.00 $\pm$ 0.02 & 6.26 $\pm$ 0.02 & 77.85 $\pm$ 0.14 \\
                  & Prithvi-100M &  & \checkmark & \checkmark & \checkmark & \checkmark & \checkmark &  & 4.00 $\pm$ 0.03 & 4.99 $\pm$ 0.09 & 5.43 $\pm$ 0.06 & 5.92 $\pm$ 0.01 & 6.22 $\pm$ 0.06 & 77.25 $\pm$ 0.56 \\
                  & Prithvi-300M &  & \checkmark & \checkmark & \checkmark & \checkmark & \checkmark &  & 4.03 $\pm$ 0.02 & 5.05 $\pm$ 0.14 & 5.46 $\pm$ 0.06 & 5.96 $\pm$ 0.02 & 6.36 $\pm$ 0.08 & 76.95 $\pm$ 0.6 \\
\hline
\multirow{3}{*}{8} & Prithvi-5M   &  & \checkmark & \checkmark & \checkmark & \checkmark & \checkmark & \checkmark & 3.99 $\pm$ 0.12 & 4.78 $\pm$ 0.18 & 5.14 $\pm$ 0.10 & \textbf{5.64 $\pm$ 0.03} & 6.01 $\pm$ 0.05 & 73.98 $\pm$ 0.58 \\
                  & Prithvi-100M &  & \checkmark & \checkmark & \checkmark & \checkmark & \checkmark & \checkmark & 3.88 $\pm$ 0.02 & \textbf{4.74 $\pm$ 0.10} & \textbf{5.14 $\pm$ 0.06} & 5.66 $\pm$ 0.02 & 6.05 $\pm$ 0.02 & \textbf{73.95 $\pm$ 0.37} \\
                  & Prithvi-300M &  & \checkmark & \checkmark & \checkmark & \checkmark & \checkmark & \checkmark & 3.90 $\pm$ 0.04 & 4.77 $\pm$ 0.15 & 5.15 $\pm$ 0.08 & 5.68 $\pm$ 0.03 & 6.08 $\pm$ 0.05 & 74.04 $\pm$ 0.32 \\
\hline
0 & CNN (AEX-Base) &  &  &  &  &  &  & \checkmark & \textbf{3.84 $\pm$ 0.01} & 4.82 $\pm$ 0.17 & 5.20 $\pm$ 0.09 & 5.68 $\pm$ 0.01 & \textbf{5.94 $\pm$ 0.02} & 74.46 $\pm$ 0.12 \\
\hline
\end{tabular}
\end{adjustbox}
\end{table*}

We organize the empirical study around three questions that map directly onto the three subsections below. All reported metrics, tables, and figures in this section are computed on the held-out \emph{Biomazon} test set. First, which predictor modalities together carry information about vertical structure and biomass at 20\,m over the Amazon, and does a larger Prithvi-EO-2.0 backbone translate the added capacity into better \acs{RH} or \acs{AGBD} estimates? Second, does training the full \ac{RH} profile jointly with \acs{AGBD} similar to the unified deep learning model used in \cite{Weber_2025} help either target, and how does it compare with training individual \acs{RH98} percentile or \acs{AGBD} alone? Third, how do \emph{Biomazon} baselines compare, over spatially and temporally aligned windows, with \acs{GEDI} \acs{L4D} imputed \acs{RH}/\acs{AGBD} layers and other widely used canopy-height and biomass products? We discuss the results with two complementary concerns. On the machine-learning side, whether scale, modality richness, and foundation embeddings yield monotone improvements or whether simpler baselines already saturate the task. On the remote-sensing side, what the achieved \acs{RMSE} implies for retrievability of distinct canopy strata (understory, mid-canopy, upper canopy) and for wall-to-wall \acs{AGBD} products at 20\,m, given the physical sensitivity of each sensor and the footprint-level noise inherited from \acs{GEDI} supervision. Throughout, we interpret each result as a benchmark signal: what a future foundation model, fusion architecture, or structure--biomass model would need to improve under the same spatial splits and full-\acs{RH}-plus-\acs{AGBD} protocol.

\subsection{Effect of Modalities and Model Scale}
\label{subsec:results1}

\begin{figure*}[!t]
   \centering
    \includegraphics[width=0.82\textwidth]{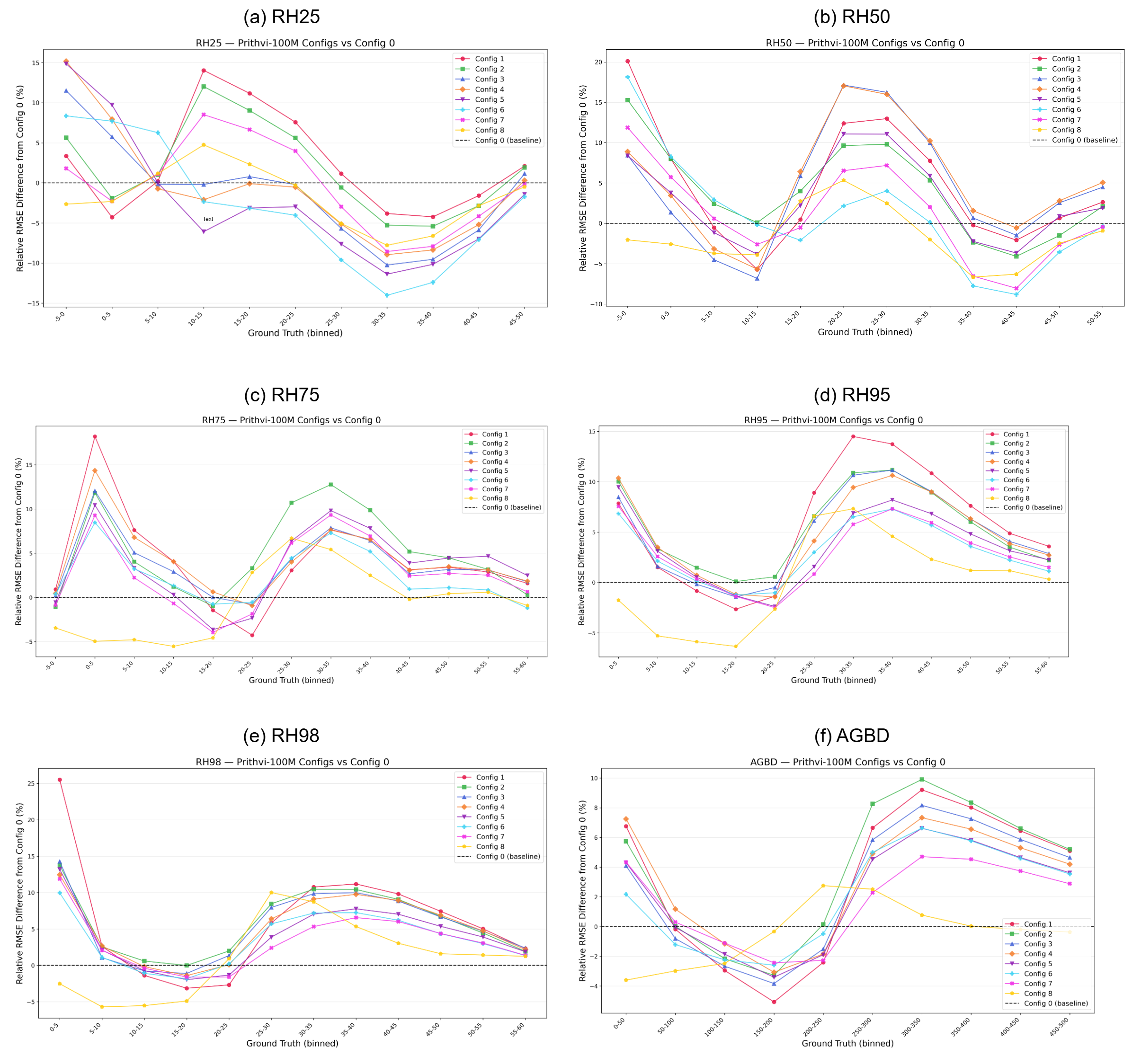}
    \caption{Bin-wise relative \ac{RMSE} difference \% of Config~1-8 vs Config~0, using Prithvi-100M encoder. Config~0 \acs{AEX-Base} is used as reference point. For \acs{RH95}, min value is -0.33 (Appendix~\ref{sec:app_stats}: Table~\ref{tab:gedi_stats},~\ref{tab:gedi_stats_extras}) we clip the negative values to 0 so as to avoid a new [-5, 0] bin.}
    \label{fig:modality_relative_diff}
\end{figure*}

Table~\ref{tab:test_rmse_ablation} reports test \acs{RMSE} for the five \acs{RH} percentiles and \acs{AGBD} across nine modality configurations. We start from the six \acs{HLS}-aligned Sentinel-2 bands (Config~1) that Prithvi was pretrained on, and add one modality at a time, swapping to the full ten-band Sentinel-2 stack and then appending Sentinel-1, ALOS-2 PALSAR-2, the Copernicus DEM, and the Dynamic World V1 \acs{LULC} layer. We treat each cumulative stack as a separate configuration (Configs~1--7). Config~8 adds \acs{AEX} as a late-fusion input on top of Config~7's raw stack via the \acs{FiLM}-gated residual fusion of Sec.~\ref{subsubsec:late_fusion}. We run each configuration at Prithvi-5M and Prithvi-100M to track how backbone capacity interacts with modality richness. Early runs showed only incremental \acs{RMSE} updates when moving from Prithvi-5M to Prithvi-100M across Configs~1--6, so we restrict the 300M backbone to Config~7 (all raw modalities) and Config~8 (\acs{AEX} fused with the \acs{DPT} output of all raw modalities), where any remaining capacity gain is most likely to surface. Config~0 acts as the reference baseline for these comparisons: a shallow \acs{CNN} over the 64-band \acs{AEX} field alone, with no Prithvi--\acs{DPT} stack. Because \acs{AEX} was itself pretrained against \acs{GEDI} together with the same raw modalities \cite{brown2025alphaearthfoundationsembeddingfield}, Config~0 is the tightest reference point that any Prithvi--\acs{DPT} configuration should beat for the encoder and decoder cost to be justified. All runs in this subsection use the unified \texttt{\acs{U-RH-AGBD}} training setup, in which the model jointly predicts the full monotone \acs{RH} profile and \acs{AGBD}. The reported \acs{RH25}, \acs{RH50}, \acs{RH75}, \acs{RH95}, and \acs{RH98} values are evaluated slices of that full-profile prediction rather than separately trained scalar heads. These five \ac{RH} slices were selected to sample the vertical profile while keeping the ablation table compact: \acs{RH25} serves as a proxy for lower-canopy structure, \acs{RH50} the mid-canopy, \acs{RH75} the upper canopy, and \acs{RH95}/\acs{RH98} the canopy-top proxies most commonly used in \acs{GEDI}-supervised height products.

Config~1 vs Config~2 tests whether Prithvi's \acs{HLS} pretraining transfers to \emph{Biomazon}. Config~1 uses the six \acs{HLS}-aligned Sentinel-2 bands the backbone was pretrained on, but at 20\,m instead of \acs{HLS}'s 30\,m and with \emph{Biomazon's} basin-specific band statistics rather than the pretraining ones \cite{szwarcman2025prithvi}. Config~2 widens the stack to all ten Sentinel-2 bands, with the four extra red-edge channels initialized from scratch in the patch-embedding projection. At both Prithvi-5M and Prithvi-100M the two configurations sit within seed variability of each other on every target: the pretrained \acs{HLS} weights produce no measurable transfer advantage once resolution and band statistics change, and the four added bands neither help nor hurt under end-to-end fine-tuning. Stacking the remaining raw modalities up to Config~7 at Prithvi-100M improves every target but only modestly: roughly 1.2~Mg\,ha$^{-1}$ on \acs{AGBD} and 0.15--0.20~m on \acs{RH75}--\acs{RH98}, with \acs{RH25} barely moving. Among the \acs{SAR} streams (Config~3 vs Config~4), C-band Sentinel-1 produces the larger \acs{AGBD} reduction, which is mildly counterintuitive since L-band ALOS-2 should penetrate deeper into woody biomass and saturate later than C-band \cite{huang_sar}. We attribute this gap to ALOS-2's coarser native ScanSAR resolution. The \acs{DEM} and \acs{LULC} streams sit within seed variability over a basin where terrain and land cover are largely homogeneous. The largest single movement in the table is the \acs{AEX} late-fusion step (Config~7~$\to$~8): on its own it delivers a bigger \acs{AGBD} reduction than the entire cumulative raw-sensor stack, and tightens the mid-canopy \acs{RH50}--\acs{RH75} by roughly 5\%, more than canopy-top \acs{RH98} or understory \acs{RH25}. The scaling axis behaves differently. Prithvi-5M to Prithvi-100M tightens \acs{AGBD} by 0.6--1.6~Mg\,ha$^{-1}$ and \acs{RH75}--\acs{RH98} by 0.05--0.15~m in the raw-modality configurations, but the gain shrinks as the input stack grows and almost disappears at Config~8. Prithvi-300M adds no \acs{AGBD} headroom and consistently worsens \acs{RH} metrics relative to Prithvi-100M, pointing to \acs{GEDI} supervision density rather than encoder capacity as the binding constraint. Config~0 ties the two axes together: a lightweight \acs{CNN} on the 64-band \acs{AEX} field alone, with no Prithvi--\acs{DPT} stack, outperforms every raw-modality Prithvi configuration at every backbone size. Only Config~8, with \acs{AEX} fused back into the Prithvi--\acs{DPT} pipeline, catches up, and the two trade wins within roughly 0.1~m on individual percentiles.

\begin{table*}[b]
\centering
\footnotesize
\renewcommand{\arraystretch}{1.15}
\caption{RH98 performance across encoders and training setups. Results are reported as mean $\pm$ std over 5 runs.}
\label{tab:rh98_encoder_paradigm}
\begin{tabular}{llccccccc}
\hline
\textbf{Encoder} & \textbf{Paradigm} & \textbf{RMSE} $\downarrow$ & \textbf{RMSE\%} $\downarrow$ & \textbf{MAE} $\downarrow$ & \textbf{MAE\%} $\downarrow$ & \textbf{$\mathbf{R^2}$} $\uparrow$ & \textbf{BIAS} & \textbf{BIAS\%} \\
\hline

AEXB         & \multirow{4}{*}{U-RH-AGBD}     & 5.94 $\pm$ 0.02 & 31.3 $\pm$ 0.08  & 4.12 $\pm$ 0.02 & 21.67 $\pm$ 0.1  & 0.75 $\pm$ 0.0 & -0.77 $\pm$ 0.08 & -4.09 $\pm$ 0.43 \\
Prithvi-5M   &                                 & 6.01 $\pm$ 0.05 & 31.67 $\pm$ 0.25 & 4.27 $\pm$ 0.08 & 22.49 $\pm$ 0.45 & 0.74 $\pm$ 0.01 & -1.2 $\pm$ 0.25  & -6.33 $\pm$ 1.29 \\
Prithvi-100M &                                 & 6.05 $\pm$ 0.02 & 31.88 $\pm$ 0.08 & 4.3 $\pm$ 0.04  & 22.64 $\pm$ 0.23 & 0.74 $\pm$ 0.0  & -1.26 $\pm$ 0.13 & -6.64 $\pm$ 0.73 \\
Prithvi-300M &                                 & 6.08 $\pm$ 0.05 & 32.04 $\pm$ 0.27 & 4.34 $\pm$ 0.08 & 22.82 $\pm$ 0.43 & 0.74 $\pm$ 0.0  & -1.3 $\pm$ 0.2   & -6.83 $\pm$ 1.09 \\
\hline

AEXB         & \multirow{4}{*}{U-RH}          & 5.92 $\pm$ 0.02 & 31.18 $\pm$ 0.11 & 4.09 $\pm$ 0.01 & 21.53 $\pm$ 0.08 & 0.75 $\pm$ 0.0 & -0.61 $\pm$ 0.19 & -3.21 $\pm$ 1.0 \\
Prithvi-5M   &                                 & \textbf{5.6 $\pm$ 0.01}  & 31.83 $\pm$ 0.09 & \textbf{3.82 $\pm$ 0.02} & 21.69 $\pm$ 0.14 & 0.76 $\pm$ 0.0 & 0.17 $\pm$ 0.31  & 0.99 $\pm$ 1.74 \\
Prithvi-100M &                                 & 5.64 $\pm$ 0.02 & 32.03 $\pm$ 0.13 & 3.92 $\pm$ 0.08 & 22.25 $\pm$ 0.46 & 0.76 $\pm$ 0.0 & 0.44 $\pm$ 0.39  & 2.51 $\pm$ 2.21 \\
Prithvi-300M &                                 & 5.69 $\pm$ 0.08 & 32.32 $\pm$ 0.44 & 3.92 $\pm$ 0.09 & 22.28 $\pm$ 0.54 & 0.76 $\pm$ 0.01 & \textbf{0.16 $\pm$ 0.29}  & \textbf{0.91 $\pm$ 1.62} \\
\hline

AEXB         & \multirow{4}{*}{U-RH98-AGBD}   & 5.84 $\pm$ 0.02 & 30.75 $\pm$ 0.08 & 3.97 $\pm$ 0.01 & 20.88 $\pm$ 0.05 & 0.76 $\pm$ 0.0 & 0.79 $\pm$ 0.1   & 4.14 $\pm$ 0.54 \\
Prithvi-5M   &                                 & 5.81 $\pm$ 0.01 & \textbf{30.59 $\pm$ 0.08} & 4.03 $\pm$ 0.03 & 21.25 $\pm$ 0.16 & 0.76 $\pm$ 0.0 & -0.26 $\pm$ 0.2  & -1.36 $\pm$ 1.05 \\
Prithvi-100M &                                 & 5.84 $\pm$ 0.03 & 30.75 $\pm$ 0.18 & 4.04 $\pm$ 0.04 & 21.29 $\pm$ 0.22 & 0.76 $\pm$ 0.0 & -0.24 $\pm$ 0.44 & -1.26 $\pm$ 2.32 \\
Prithvi-300M &                                 & 5.84 $\pm$ 0.05 & 30.76 $\pm$ 0.26 & 4.03 $\pm$ 0.06 & 21.23 $\pm$ 0.32 & 0.76 $\pm$ 0.0 & -0.32 $\pm$ 0.43 & -1.7 $\pm$ 2.26 \\
\hline

AEXB         & \multirow{4}{*}{RH98}          & 5.81 $\pm$ 0.02 & 30.61 $\pm$ 0.1  & 3.96 $\pm$ 0.02 & \textbf{20.84 $\pm$ 0.08} & 0.76 $\pm$ 0.0 & 0.54 $\pm$ 0.2   & 2.88 $\pm$ 1.06 \\
Prithvi-5M   &                                 & 5.82 $\pm$ 0.01 & 30.66 $\pm$ 0.07 & 3.96 $\pm$ 0.01 & 20.88 $\pm$ 0.03 & 0.76 $\pm$ 0.0 & 0.38 $\pm$ 0.1   & 1.98 $\pm$ 0.51 \\
Prithvi-100M &                                 & 5.84 $\pm$ 0.02 & 30.75 $\pm$ 0.1  & 3.98 $\pm$ 0.01 & 20.95 $\pm$ 0.07 & 0.76 $\pm$ 0.0 & 0.21 $\pm$ 0.28  & 1.09 $\pm$ 1.46 \\
Prithvi-300M &                                 & 5.85 $\pm$ 0.01 & 30.79 $\pm$ 0.07 & 3.98 $\pm$ 0.02 & 20.95 $\pm$ 0.09 & 0.76 $\pm$ 0.0 & 0.26 $\pm$ 0.14  & 1.37 $\pm$ 0.75 \\
\hline

\end{tabular}
\end{table*}

In Fig.~\ref{fig:modality_relative_diff} we investigate the \acs{RMSE} percentage difference of Prithvi-100M as example of Prithvi models, in relative to \acs{AEXB}. It shows that the relative errors have a strong bin-wise structure that is largely shared across the Prithvi-100M configurations. Rather than each modality producing an entirely different curve shape, Configs~1--7 tend to rise and fall together across target bins, suggesting that much of the pattern reflects target-range difficulty under the shared training objective. Within this common structure, the added modalities mainly change the magnitude of the departures from the \acs{AEX-Base} reference. Config~8, which adds \acs{AEX} through late fusion, most consistently reduces these departures but does not shift the full curve below the baseline. For \acs{RH25} and \acs{RH50}, Config~8 drops below \acs{AEX-Base} across parts of the start and upper-middle target range, then rises again toward the largest bins. For \acs{RH95} and \acs{RH98}, the Config~8 curve is non-monotonic: it improves over \acs{AEX-Base} in the low-to-moderate height bins, rises above the baseline in the moderate-to-tall range, and then drops back toward the baseline in the tallest bins. The \acs{AGBD} panel follows the same interpretation: \acs{AEX} late fusion suppresses much of the positive relative error seen in the raw-modality configurations, especially over the higher-biomass range, but it does not dominate \acs{AEX-Base} in every bin. This shared bin-wise structure may partly reflect the \acs{LDS}-weighted objective and the uneven target distribution, while the reduced amplitude of Config~8 reflects the added information carried by \acs{AEX}. Overall, Fig.~\ref{fig:modality_relative_diff} supports the table-level conclusion that \acs{AEX} is the most influential added modality, while clarifying that its advantage is range-dependent rather than uniform across canopy-height and biomass bins.

\subsection{Effect of Training Setups}
\label{subsec:results2}

\begin{table*}[t]
\centering
\footnotesize
\renewcommand{\arraystretch}{1.15}
\caption{AGBD performance across encoders and training setups. Results are reported as mean $\pm$ std over 5 runs.}
\label{tab:agbd_encoder_paradigm}
\begin{tabular}{llccccccc}
\hline
\textbf{Encoder} & \textbf{Paradigm} & \textbf{RMSE} $\downarrow$ & \textbf{RMSE\%} $\downarrow$ & \textbf{MAE} $\downarrow$ & \textbf{MAE\%} $\downarrow$ & $\mathbf{R^2}$ $\uparrow$ & \textbf{BIAS} & \textbf{BIAS\%} \\
\hline

AEXB         & \multirow{4}{*}{U-RH-AGBD}    & 74.46 $\pm$ 0.12 & 61.32 $\pm$ 0.1  & 50.04 $\pm$ 0.28 & 41.21 $\pm$ 0.23 & 0.61 $\pm$ 0.0  & 15.0 $\pm$ 1.29  & 12.35 $\pm$ 1.06 \\
Prithvi-5M   &                                & 73.98 $\pm$ 0.58 & 60.93 $\pm$ 0.48 & 49.34 $\pm$ 0.55 & 40.63 $\pm$ 0.45 & 0.62 $\pm$ 0.01 & 13.81 $\pm$ 2.56 & 11.37 $\pm$ 2.11 \\
Prithvi-100M &                                & 73.95 $\pm$ 0.37 & 60.9 $\pm$ 0.31  & 48.96 $\pm$ 0.4  & 40.32 $\pm$ 0.33 & 0.62 $\pm$ 0.0  & 12.82 $\pm$ 1.85 & 10.56 $\pm$ 1.53 \\
Prithvi-300M &                                & 74.04 $\pm$ 0.32 & 60.98 $\pm$ 0.27 & 49.06 $\pm$ 0.35 & 40.4 $\pm$ 0.29  & 0.62 $\pm$ 0.0  & 11.61 $\pm$ 1.65 & 9.56 $\pm$ 1.35 \\
\hline

AEXB         & \multirow{4}{*}{U-RH98-AGBD}  & 74.64 $\pm$ 0.11 & 61.47 $\pm$ 0.09 & 50.34 $\pm$ 0.14 & 41.46 $\pm$ 0.11 & 0.61 $\pm$ 0.0  & 16.42 $\pm$ 0.68 & 13.53 $\pm$ 0.56 \\
Prithvi-5M   &                                & 73.49 $\pm$ 0.43 & 60.53 $\pm$ 0.35 & 48.75 $\pm$ 0.44 & 40.15 $\pm$ 0.37 & 0.62 $\pm$ 0.01 & 12.07 $\pm$ 2.1  & 9.94 $\pm$ 1.73 \\
Prithvi-100M &                                & 73.48 $\pm$ 0.37 & 60.52 $\pm$ 0.31 & 48.53 $\pm$ 0.47 & 39.97 $\pm$ 0.38 & 0.62 $\pm$ 0.01 & 11.14 $\pm$ 2.12 & 9.18 $\pm$ 1.74 \\
Prithvi-300M &                                & \textbf{73.42 $\pm$ 0.52} & \textbf{60.46 $\pm$ 0.43} & \textbf{48.34 $\pm$ 0.4}  & \textbf{39.82 $\pm$ 0.33} & 0.62 $\pm$ 0.01 & \textbf{9.72 $\pm$ 2.31}  & \textbf{8.0 $\pm$ 1.9} \\
\hline

AEXB         & \multirow{4}{*}{AGBD}         & 73.97 $\pm$ 0.27 & 60.92 $\pm$ 0.22 & 49.69 $\pm$ 0.25 & 40.92 $\pm$ 0.21 & 0.62 $\pm$ 0.0  & 14.53 $\pm$ 1.43 & 11.96 $\pm$ 1.17 \\
Prithvi-5M   &                                & 73.55 $\pm$ 0.25 & 60.57 $\pm$ 0.2  & 48.92 $\pm$ 0.35 & 40.29 $\pm$ 0.29 & 0.62 $\pm$ 0.0  & 12.42 $\pm$ 1.43 & 10.23 $\pm$ 1.18 \\
Prithvi-100M &                                & 73.54 $\pm$ 0.56 & 60.57 $\pm$ 0.46 & 48.82 $\pm$ 0.39 & 40.21 $\pm$ 0.32 & 0.62 $\pm$ 0.01 & 11.37 $\pm$ 1.48 & 9.37 $\pm$ 1.22 \\
Prithvi-300M &                                & 73.81 $\pm$ 0.58 & 60.79 $\pm$ 0.48 & 48.88 $\pm$ 0.58 & 40.25 $\pm$ 0.48 & 0.62 $\pm$ 0.01 & 12.04 $\pm$ 2.25 & 9.92 $\pm$ 1.85 \\
\hline

\end{tabular}
\end{table*}

\begin{figure*}[!t]
   \centering
    \includegraphics[width=0.81\textwidth]{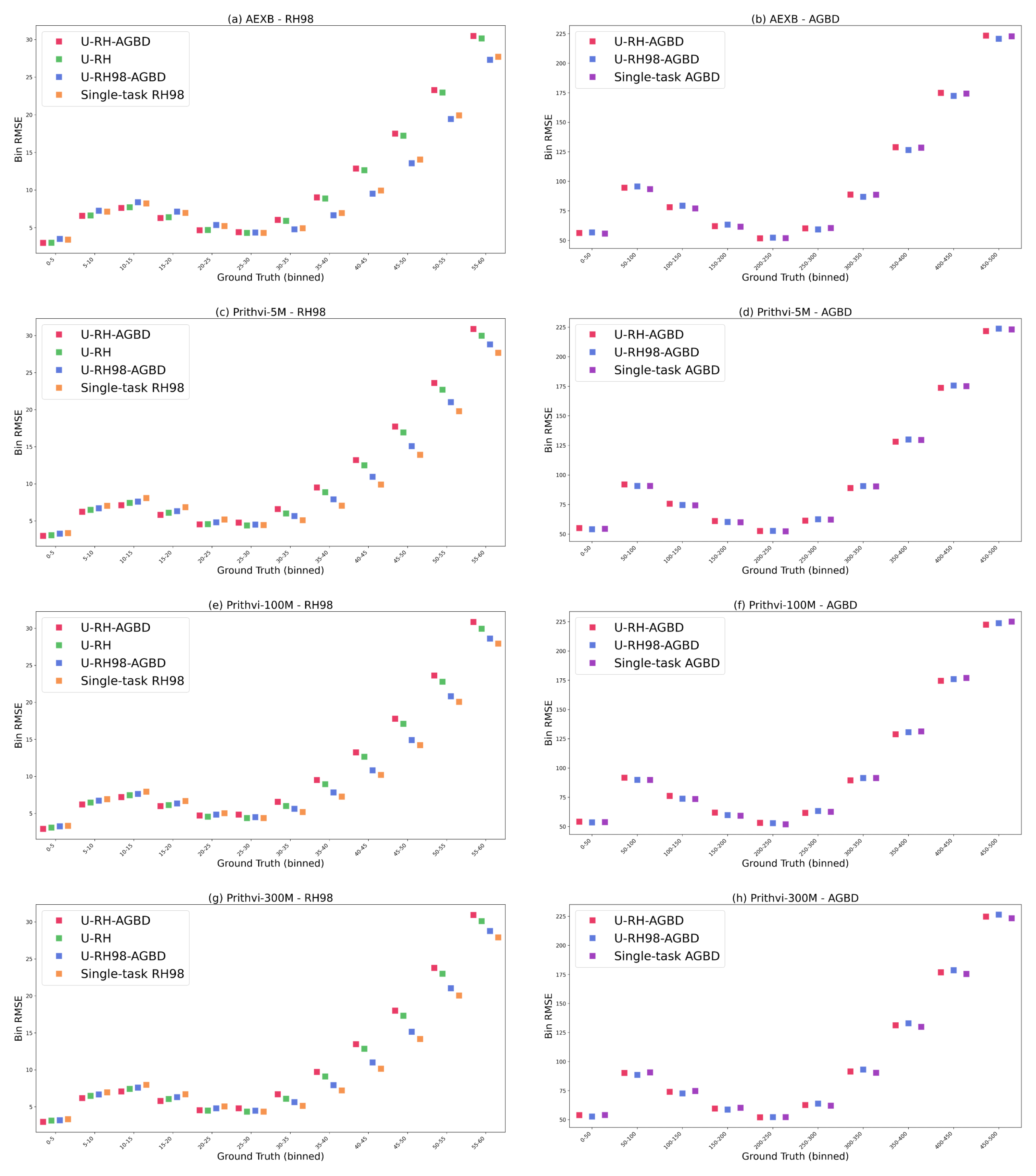}
    \caption{Bin-wise \acs{RMSE} of models under different training setups. Prithvi setups used Config~8 while \acs{AEXB} setups used Config~0. Left column shows \acs{RH98} results of all models while right column shows \acs{AGBD} results.}
    \label{fig:setups}
\end{figure*}

Table~\ref{tab:rh98_encoder_paradigm} and Table~\ref{tab:agbd_encoder_paradigm} isolate the effect of the training setup after fixing the input representation to the strongest modality setting from Sec.~\ref{subsec:results1}. For the Prithvi models, all comparisons use Config~8, while the \acs{AEXB} rows provide an embedding-only reference under the same target definitions. We focus on \acs{RH98} and \acs{AGBD} because canopy-height proxies and biomass density are the predominant scalar targets in the forest-structure mapping literature. This also makes them a natural test case for asking whether the field should move from isolated scalar supervision toward full vertical-structure supervision. We compare four paradigms: \texttt{\acs{U-RH-AGBD}}, which jointly predicts the full monotone \ac{RH} profile and \acs{AGBD}; \texttt{\acs{U-RH}}, which predicts only the full \ac{RH} profile; \texttt{\acs{U-RH98-AGBD}}, which jointly predicts the scalar canopy-height proxy \acs{RH98} and \acs{AGBD}; and the corresponding single-target \texttt{RH98} or \texttt{AGBD} setup. This design tests whether learning the complete vertical profile provides useful supervision for canopy-top height, whether \acs{AGBD} benefits from being trained together with structure, and whether any gain depends on encoder scale. Fig.~\ref{fig:setups} complements these aggregate metrics by showing the residual distributions across \acs{RH98} and \acs{AGBD} target bins.

Table~\ref{tab:rh98_encoder_paradigm} shows that, for the Prithvi encoders, full-profile \texttt{\acs{U-RH}} training gives the best aggregate \acs{RH98} \acs{RMSE} and \acs{MAE}. Relative to single-target \texttt{\acs{RH98}} training, it reduces \acs{RMSE} by about 0.16--0.22\,m, indicating that the ordered \acs{RH} profile provides useful vertical-structure supervision for canopy-top estimation. \texttt{\acs{U-RH98-AGBD}} and single-target \texttt{\acs{RH98}} form the next-best group and remain very close across encoders, whereas \texttt{\acs{U-RH-AGBD}} is consistently weakest. For \acs{AGBD}, Table~\ref{tab:agbd_encoder_paradigm} shows that Prithvi benefits from scalar structure--biomass co-training rather than full-profile supervision: \texttt{\acs{U-RH98-AGBD}} gives the best Prithvi results, with small but consistent improvements in \acs{RMSE}, \acs{MAE}, and reduced positive bias over single-target \texttt{\acs{AGBD}}, while \texttt{\acs{U-RH-AGBD}} does not improve biomass prediction. The \acs{AEXB} baseline differs from this pattern, with the best \acs{RH98} and \acs{AGBD} results both coming from the corresponding single-target scalar setups, suggesting that the shallow embedding-only model gains little from auxiliary-target supervision. Overall, unified training is useful mainly for the shared Prithvi--\acs{DPT} models when the auxiliary target is well matched: the full \ac{RH} profile helps \acs{RH98}, whereas \acs{RH98} is a more effective structural companion for \acs{AGBD} than the entire profile.

Fig.~\ref{fig:setups} shows how bin-wise error is distributed across the target range under different training setups. For \acs{RH98}, the scalar-focused setups preserve the tall-canopy tail better: \texttt{\acs{RH98}} and \texttt{\acs{U-RH98-AGBD}} generally give lower bin \acs{RMSE} in the upper \acs{RH98} bins, while \texttt{\acs{U-RH-AGBD}} is consistently the weakest and \texttt{\acs{U-RH}} lies between them. In the lower and middle \acs{RH98} bins, the differences are smaller and \texttt{\acs{U-RH}} is often competitive, which explains why the \texttt{\acs{U-RH}} training setup improves the aggregate \acs{RH98} metrics despite poorer tail behavior. Since all runs use \acs{LDS}, this suggests that the effect depends on how directly the reweighted signal reaches the target: in scalar \acs{RH98} training, the \acs{LDS}-weighted loss acts directly on the \acs{RH98} head, whereas in full-profile training the \acs{RH98} signal is coupled to the monotone \acs{RH} profile and competes with losses from the remaining percentiles. For \acs{AGBD}, the separation among setups is much weaker, and the bin-wise differences between training setups are not clearly distinguishable.

\subsection{Comparison with GEDI L4D and other products}
\label{subsec:results3}

% =========================
% Complete Table: RH10--RH98 & AGBD
% =========================
\begin{table*}[!t]
\centering
\footnotesize
\renewcommand{\arraystretch}{1.08}
\caption{Product-aligned comparison of our baselines across \acs{RH10}--\acs{RH98} and \acs{AGBD}. The three Prithvi-\acs{DPT} models use Config~8 \texttt{\acs{U-RH-AGBD}}. \acs{AEXB} and Prithvi entries report metrics from five-run ensemble-mean predictions.}
\label{tab:product_aligned_rh_agbd}
\begin{tabular*}{\textwidth}{@{\extracolsep{\fill}}llccccccc@{}}
\hline
\textbf{Name} & \textbf{Product} & \textbf{RMSE} $\downarrow$ & \textbf{RMSE\%} $\downarrow$ & \textbf{MAE} $\downarrow$ & \textbf{MAE\%} $\downarrow$ & \textbf{$\mathbf{R^2}$} $\uparrow$ & \textbf{Bias} & \textbf{Bias\%} \\
\hline

GEDI L4D \cite{Seo_2025_GEDI_L4D_Imputed_Waveforms_V2}     & \multirow{5}{*}{RH10} & 4.02 & 497.98 & 2.27 & 281.31 & -0.12 & 0.84  & 103.56 \\
AEXB         &                       & \textbf{2.7}  & \textbf{334.31} & \textbf{1.88} & \textbf{232.29} & \textbf{0.5}   & \textbf{0.06}  & \textbf{7.45}   \\
Prithvi-5M   &                       & 2.8  & 346.97 & 1.97 & 243.68 & 0.46  & -0.72 & -88.94 \\
Prithvi-100M &                       & 2.71 & 335.29 & 1.88 & 233.08 & 0.49  & -0.07 & -8.69  \\
Prithvi-300M &                       & 2.7  & 334.69 & 1.88 & 233.34 & 0.5   & -0.09 & -11.6  \\
\hline

GEDI L4D \cite{Seo_2025_GEDI_L4D_Imputed_Waveforms_V2}     & \multirow{5}{*}{RH20} & 4.84 & 128.16 & 2.77 & 73.33 & 0.24 & 0.9  & 23.72  \\
AEXB         &                       & \textbf{3.55} & \textbf{93.96}  & \textbf{2.46} & \textbf{65.1}  & \textbf{0.59} & \textbf{0}    & \textbf{-0.07}  \\
Prithvi-5M   &                       & 3.59 & 94.85  & 2.5  & 66.27 & 0.58 & -0.51 & -13.39 \\
Prithvi-100M &                       & 3.56 & 94.17  & 2.47 & 65.25 & 0.59 & 0.08 & 2.12   \\
Prithvi-300M &                       & 3.55 & 94.03  & 2.47 & 65.33 & 0.59 & -0.15 & -3.92  \\
\hline

GEDI L4D \cite{Seo_2025_GEDI_L4D_Imputed_Waveforms_V2}     & \multirow{5}{*}{RH30} & 5.32 & 89.82 & 3.04 & 51.33 & 0.37 & 0.91  & 15.43  \\
AEXB         &                       & 4.24 & 71.58 & 2.9  & 49.02 & 0.6  & 1.01  & 17.06  \\
Prithvi-5M   &                       & 4.3  & 72.62 & 3.06 & 51.72 & 0.59 & -1.44 & -24.34 \\
Prithvi-100M &                       & 4.07 & 68.65 & \textbf{2.83} & \textbf{47.73} & 0.63 & -0.49 & -8.29  \\
Prithvi-300M &                       & \textbf{4.05} & \textbf{68.43} & 2.83 & 47.75 & \textbf{0.64} & \textbf{-0.49} & \textbf{-8.29}  \\
\hline

GEDI L4D \cite{Seo_2025_GEDI_L4D_Imputed_Waveforms_V2}    & \multirow{5}{*}{RH40} & 5.66 & 73.63 & 3.23 & 41.99 & 0.44 & 0.92 & 11.92 \\
AEXB         &                       & 4.57 & 59.48 & 3.13 & 40.76 & 0.64 & 1.09 & 14.22 \\
Prithvi-5M   &                       & 4.38 & 56.97 & 3.04 & 39.61 & 0.67 & 0.13 & 1.66  \\
Prithvi-100M &                       & 4.56 & 59.31 & 3.16 & 41.08 & 0.64 & 1.04 & 13.52 \\
Prithvi-300M &                       & \textbf{4.37} & \textbf{56.91} & \textbf{3.03} & \textbf{39.44} & \textbf{0.67} & \textbf{0.08} & \textbf{1.03}  \\
\hline

GEDI L4D \cite{Seo_2025_GEDI_L4D_Imputed_Waveforms_V2}    & \multirow{5}{*}{RH50} & 5.93 & 64.11 & 3.38 & 36.54 & 0.49 & 0.91 & 9.83  \\
AEXB         &                       & 4.74 & 51.25 & 3.23 & 34.92 & 0.68 & 0.8  & 8.63  \\
Prithvi-5M   &                       & 4.68 & 50.63 & 3.2  & 34.63 & 0.68 & 0.68 & 7.33  \\
Prithvi-100M &                       & 4.67 & 50.53 & 3.2  & 34.64 & 0.69 & 0.57 & 6.14  \\
Prithvi-300M &                       & \textbf{4.59} & \textbf{49.65} & \textbf{3.2}  & \textbf{34.55} & \textbf{0.7}  & \textbf{-0.18} & \textbf{-1.9} \\
\hline

GEDI L4D \cite{Seo_2025_GEDI_L4D_Imputed_Waveforms_V2}    & \multirow{5}{*}{RH60} & 6.18 & 57.48 & 3.52 & 32.8  & 0.53 & 0.9  & 8.36  \\
AEXB         &                       & 5.04 & 46.93 & 3.4  & 31.68 & 0.69 & 1.12 & 10.44 \\
Prithvi-5M   &                       & \textbf{4.79} & \textbf{44.55} & \textbf{3.34} & \textbf{31.12} & \textbf{0.72} & \textbf{-0.44} & \textbf{-4.05} \\
Prithvi-100M &                       & 4.96 & 46.17 & 3.37 & 31.35 & 0.7  & 0.99 & 9.19  \\
Prithvi-300M &                       & 4.94 & 46.02 & 3.38 & 31.42 & 0.7  & 1.03 & 9.61  \\
\hline

GEDI L4D \cite{Seo_2025_GEDI_L4D_Imputed_Waveforms_V2}    & \multirow{5}{*}{RH70} & 6.41 & 52.27 & 3.67 & 29.91 & 0.56 & 0.88 & 7.18  \\
AEXB         &                       & 5.09 & 41.52 & 3.45 & 28.17 & 0.72 & 0.68 & 5.57  \\
Prithvi-5M   &                       & 4.95 & 40.34 & 3.43 & 27.98 & 0.74 & -0.23 & -1.86 \\
Prithvi-100M &                       & 4.98 & 40.57 & \textbf{3.41} & \textbf{27.8}  & 0.73 & \textbf{0.05} & \textbf{0.42}  \\
Prithvi-300M &                       & \textbf{4.95} & \textbf{40.32} & 3.42 & 27.86 & \textbf{0.74} & 0.09 & 0.72  \\
\hline

GEDI L4D \cite{Seo_2025_GEDI_L4D_Imputed_Waveforms_V2}    & \multirow{5}{*}{RH80} & 6.66 & 47.78 & 3.83 & 27.47 & 0.58 & 0.85 & 6.12  \\
AEXB         &                       & 5.29 & 37.96 & 3.58 & 25.71 & 0.74 & 0.8  & 5.73  \\
Prithvi-5M   &                       & \textbf{5.12} & \textbf{36.7}  & \textbf{3.51} & \textbf{25.2}  & \textbf{0.75} & \textbf{0.03} & \textbf{0.24}  \\
Prithvi-100M &                       & 5.49 & 39.38 & 3.78 & 27.12 & 0.72 & 1.78 & 12.74 \\
Prithvi-300M &                       & 5.17 & 37.09 & 3.55 & 25.46 & 0.75 & 0.68 & 4.88  \\
\hline

GEDI L4D \cite{Seo_2025_GEDI_L4D_Imputed_Waveforms_V2}    & \multirow{5}{*}{RH90} & 6.98 & 43.46 & 4.05 & 25.19 & 0.6  & 0.8  & 4.97 \\
AEXB         &                       & 5.62 & 35    & 3.92 & 24.38 & 0.74 & 1.36 & 8.49 \\
Prithvi-5M   &                       & \textbf{5.36} & \textbf{33.39} & 3.73 & 23.21 & \textbf{0.76} & \textbf{0.44} & \textbf{2.75} \\
Prithvi-100M &                       & 5.42 & 33.71 & 3.71 & 23.07 & 0.76 & 0.57 & 3.55 \\
Prithvi-300M &                       & 5.38 & 33.49 & \textbf{3.67} & \textbf{22.84} & \textbf{0.76} & \textbf{0.44} & \textbf{2.75} \\
\hline

GEDI L4D \cite{Seo_2025_GEDI_L4D_Imputed_Waveforms_V2}    & \multirow{8}{*}{RH95} & 7.23 & 41.07 & 4.22 & 23.97 & 0.61 & 0.73  & 4.16   \\
Potapov et al. \cite{Potapov_2021}     &                       & 6.89 & 39.14 & 5.04 & 28.64 & 0.64 & 0.29  & 1.66   \\
Tolan et al. \cite{Tolan_2024}       &                       & 9.3  & 52.86 & 6.86 & 38.97 & 0.35 & -5.37 & -30.53 \\
Wagner et al. \cite{Wagner_2025}      &                       & 6.87 & 39.05 & 5.14 & 29.2  & 0.64 & -2.07 & -11.75 \\
AEXB         &                       & 5.67 & 32.19 & 3.88 & 22.05 & 0.76 & 0.33  & 1.9    \\
Prithvi-5M   &                       & \textbf{5.57} & \textbf{31.66} & \textbf{3.84} & \textbf{21.85} & \textbf{0.77} & \textbf{-0.25} & \textbf{-1.43}  \\
Prithvi-100M &                       & 5.61 & 31.88 & 3.93 & 22.35 & 0.76 & -0.09 & -0.53  \\
Prithvi-300M &                       & 5.59 & 31.79 & 3.87 & 21.99 & 0.76 & -0.35 & -2.01  \\
\hline

GEDI L4D \cite{Seo_2025_GEDI_L4D_Imputed_Waveforms_V2}    & \multirow{6}{*}{RH98} & 7.45 & 39.26 & 4.38 & 23.07 & 0.61 & 0.64  & 3.39  \\
Lang et al. \cite{Lang_2023}        &                       & 7.33 & 38.59 & 5.36 & 28.23 & 0.62 & 3.14  & 16.52 \\
AEXB         &                       & \textbf{5.93} & \textbf{31.25} & \textbf{4.09} & \textbf{21.55} & \textbf{0.75} & \textbf{-0.78} & \textbf{-4.09} \\
Prithvi-5M   &                       & 5.96 & 31.41 & 4.22 & 22.22 & 0.75 & -1.2  & -6.33 \\
Prithvi-100M &                       & 6.01 & 31.66 & 4.26 & 22.42 & 0.74 & -1.26 & -6.64 \\
Prithvi-300M &                       & 6.01 & 31.65 & 4.27 & 22.46 & 0.74 & -1.3  & -6.83 \\
\hline

GEDI L4D \cite{Seo_2025_GEDI_L4D_Imputed_Waveforms_V2}    & \multirow{6}{*}{AGBD} & 88.83  & 73.16 & \textbf{42.44} & \textbf{34.96} & 0.45 & \textbf{-6.66} & \textbf{-5.49} \\
ESA Biomass CCI \cite{esa_cci}      &                       & 117.74 & 96.97 & 85.63 & 70.52 & 0.04 & 60.92 & 50.17 \\
AEXB         &                       & 74.35  & 61.23 & 49.93 & 41.12 & 0.62 & 15    & 12.35 \\
Prithvi-5M   &                       & 73.31  & 60.38 & 48.88 & 40.25 & 0.63 & 13.81 & 11.37 \\
Prithvi-100M &                       & 73.32  & 60.38 & 48.52 & 39.96 & 0.63 & 12.82 & 10.56 \\
Prithvi-300M &                       & \textbf{72.9}   & \textbf{60.03} & 48.3  & 39.78 & \textbf{0.63} & 11.61 & 9.56  \\
\hline

\end{tabular*}
\end{table*}

\begin{figure*}[!t]
   \centering
    \includegraphics[width=\textwidth]{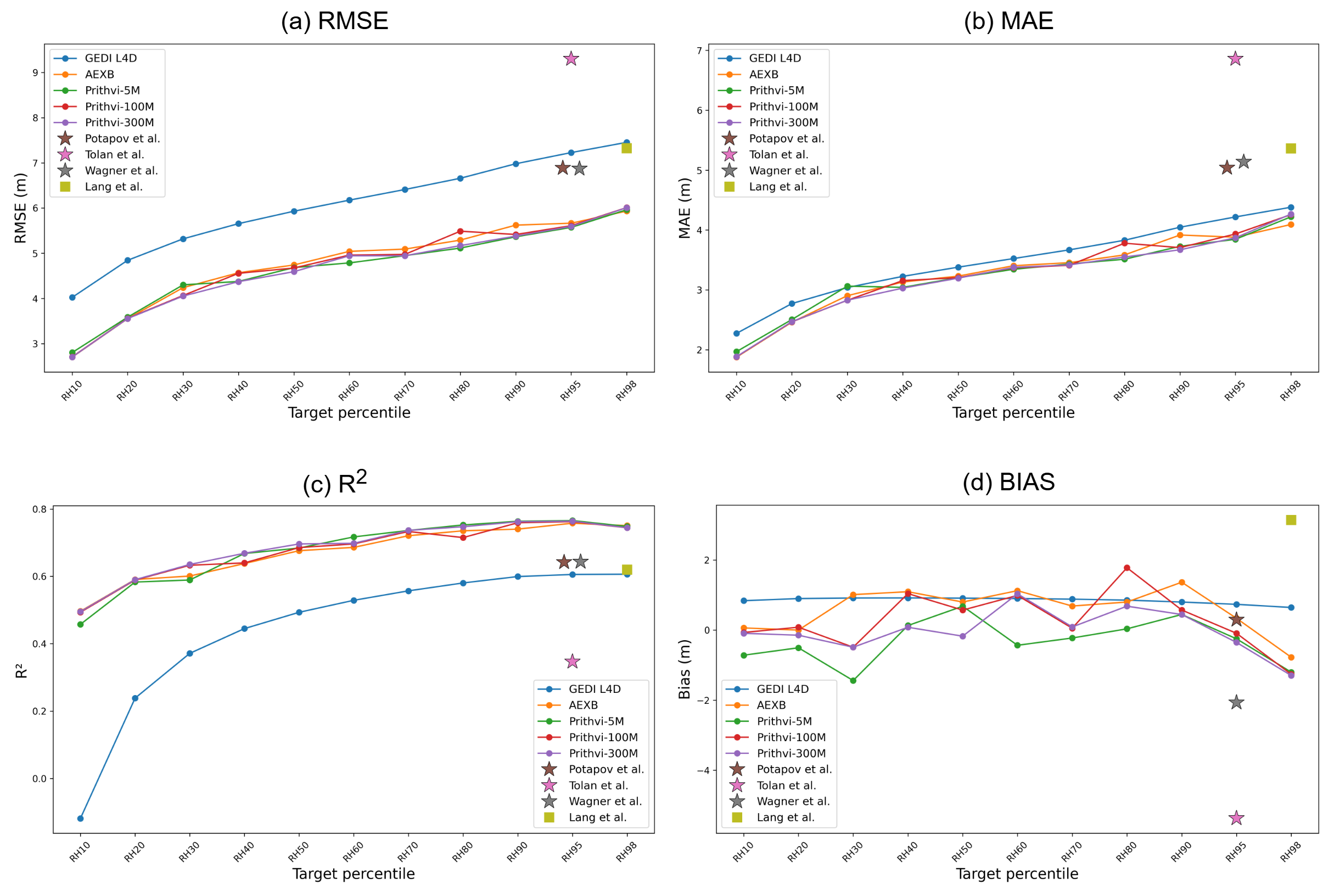}
    \caption{Comparison of our baselines vs available products across different \acs{RH} percentiles.}
    \label{fig:metrics_comparison}
\end{figure*}

\begin{figure*}[!htp]
   \centering
    \includegraphics[width=\textwidth]{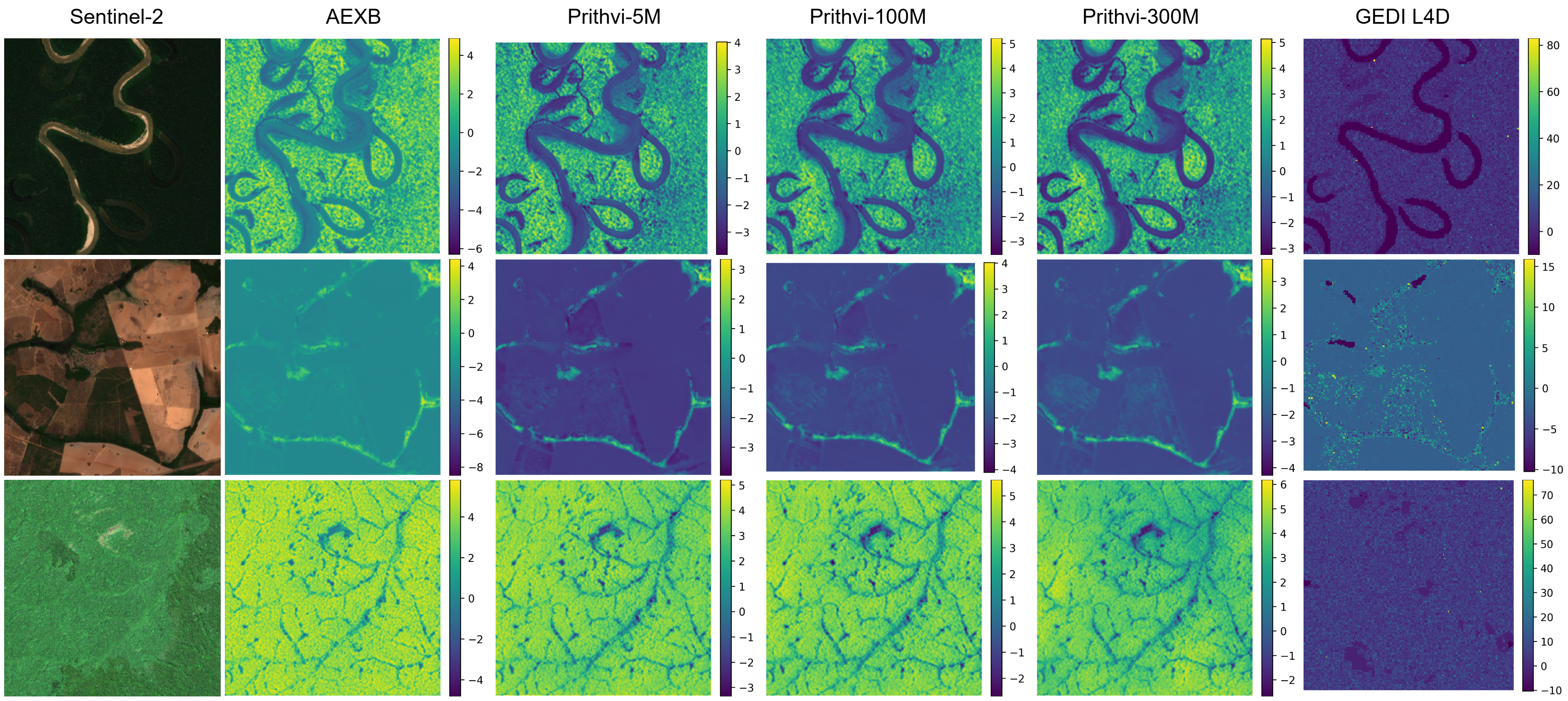}
    \caption{Visual comparison of our baselines vs \acs{GEDI} \acs{L4D} on \acs{RH10}.}
    \label{fig:rh10_comparison}
\end{figure*}

\begin{figure*}[!htp]
   \centering
    \includegraphics[width=\textwidth]{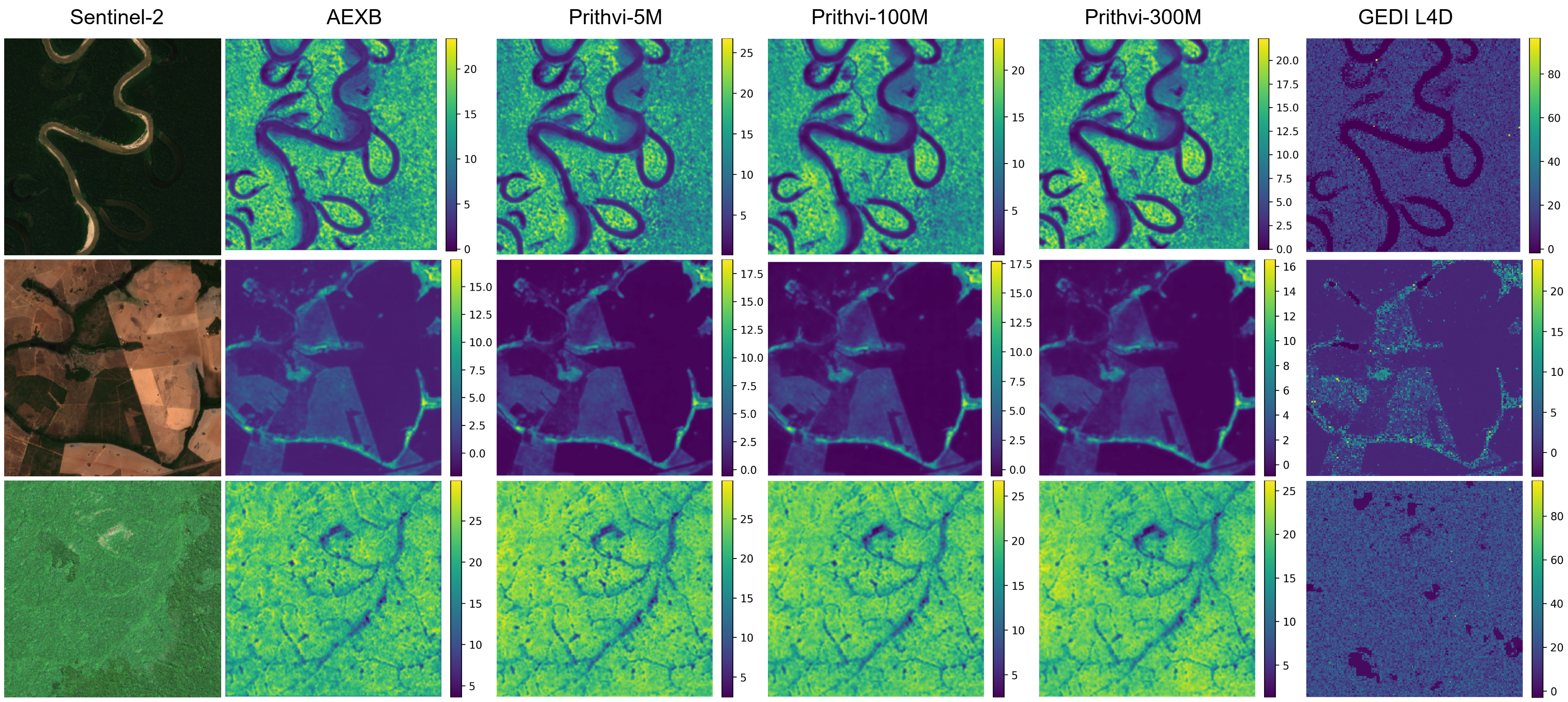}
    \caption{Visual comparison of our baselines vs \acs{GEDI} \acs{L4D} on \acs{RH50}.}
    \label{fig:rh50_comparison}
\end{figure*}

\begin{figure*}[!htp]
   \centering
    \includegraphics[width=\textwidth]{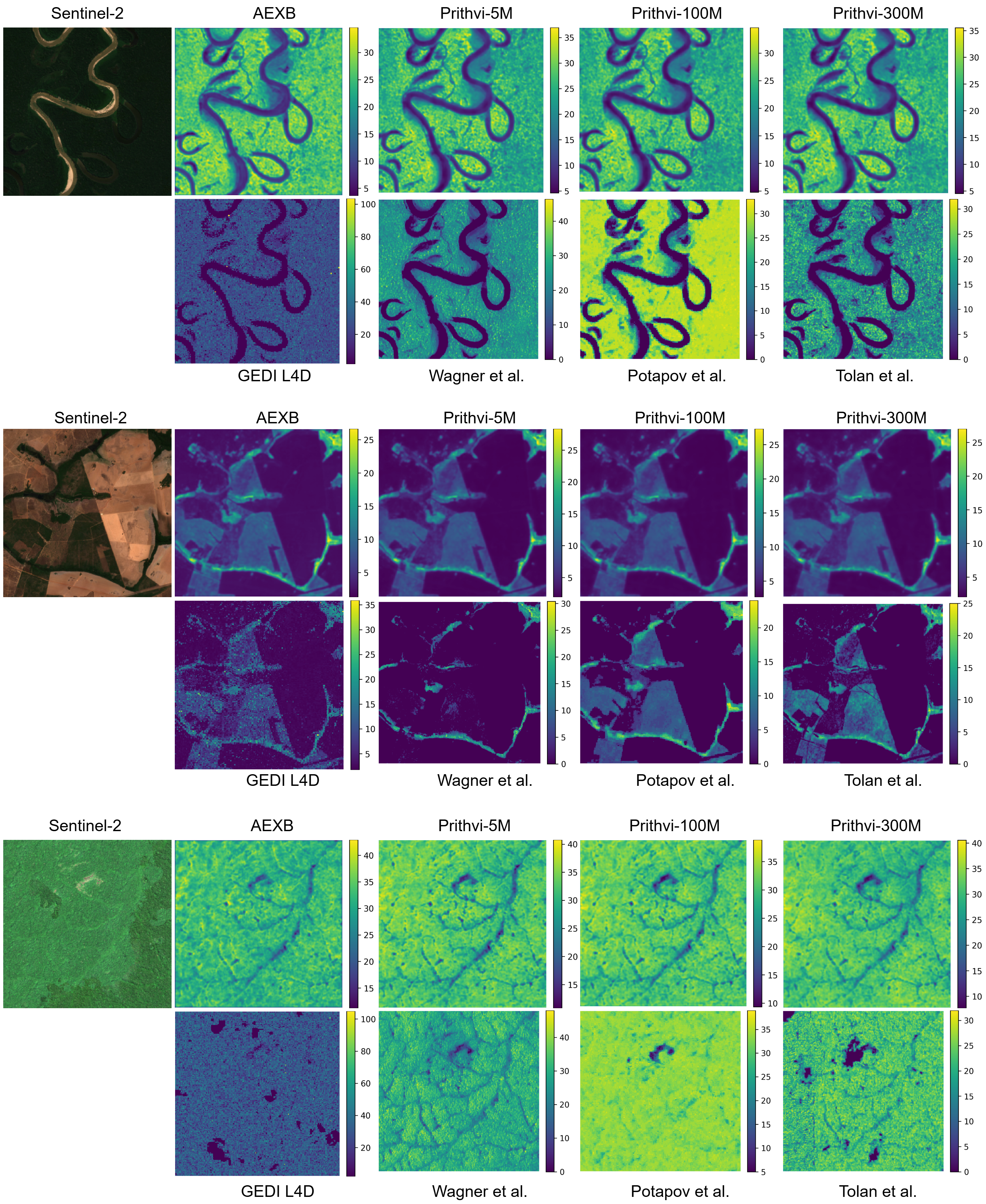}
    \caption{Visual comparison of our baselines vs \acs{GEDI} \acs{L4D}, Wagner et al., Potapov et al. and Tolan et al. on \acs{RH95}.}
    \label{fig:rh95_comparison}
\end{figure*}

\begin{figure*}[!htp]
   \centering
    \includegraphics[width=\textwidth]{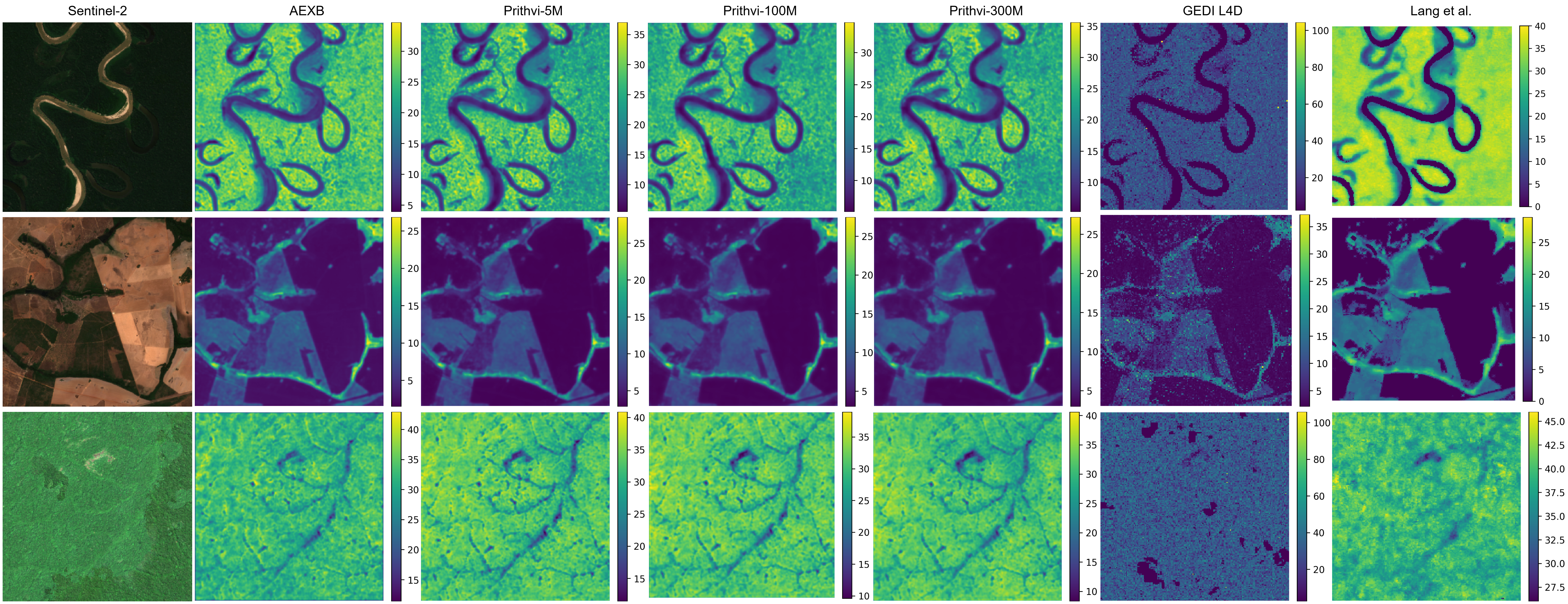}
    \caption{Visual comparison of our baselines vs \acs{GEDI} \acs{L4D} and Lang et al. on \acs{RH98}.}
    \label{fig:rh98_comparison}
\end{figure*}

\begin{figure*}[!htp]
   \centering
    \includegraphics[width=\textwidth]{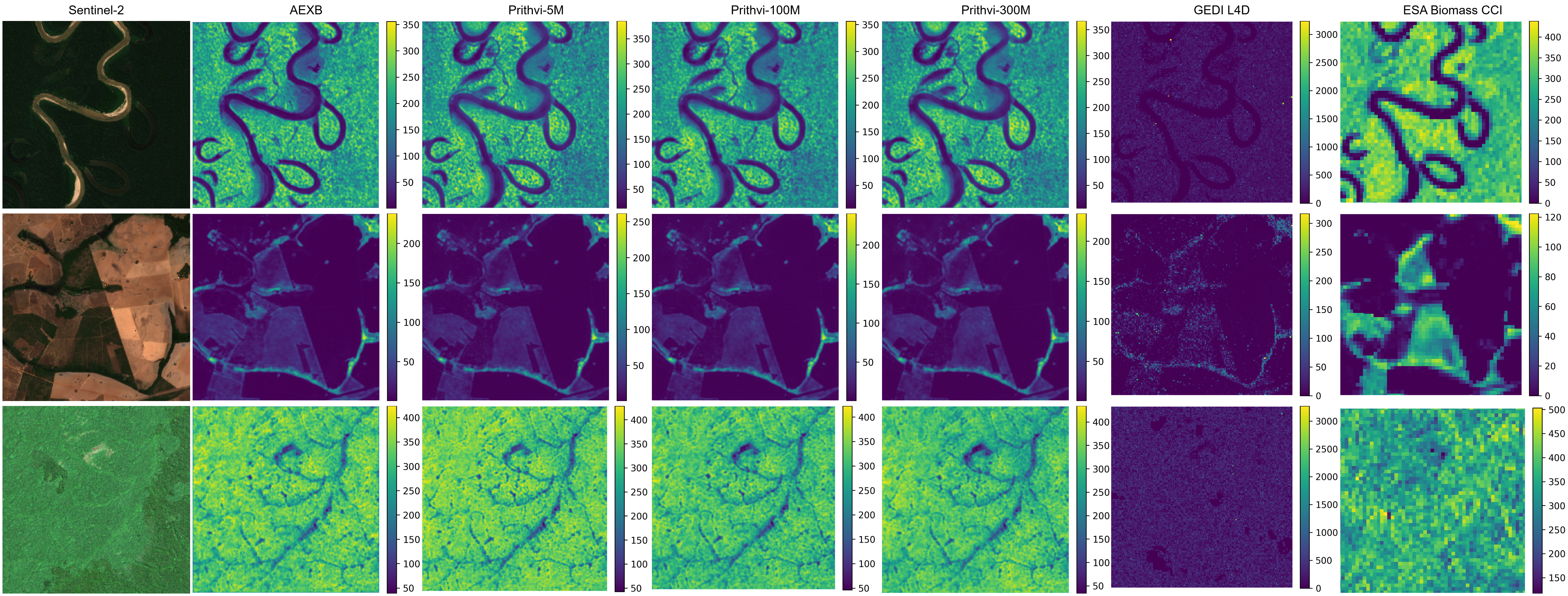}
    \caption{Visual comparison of our baselines vs \acs{GEDI} \acs{L4D} and ESA Biomass CCI on \acs{AGBD}.}
    \label{fig:agbd_comparison}
\end{figure*}

\begin{figure*}[!htp]
   \centering
    \includegraphics[width=\textwidth]{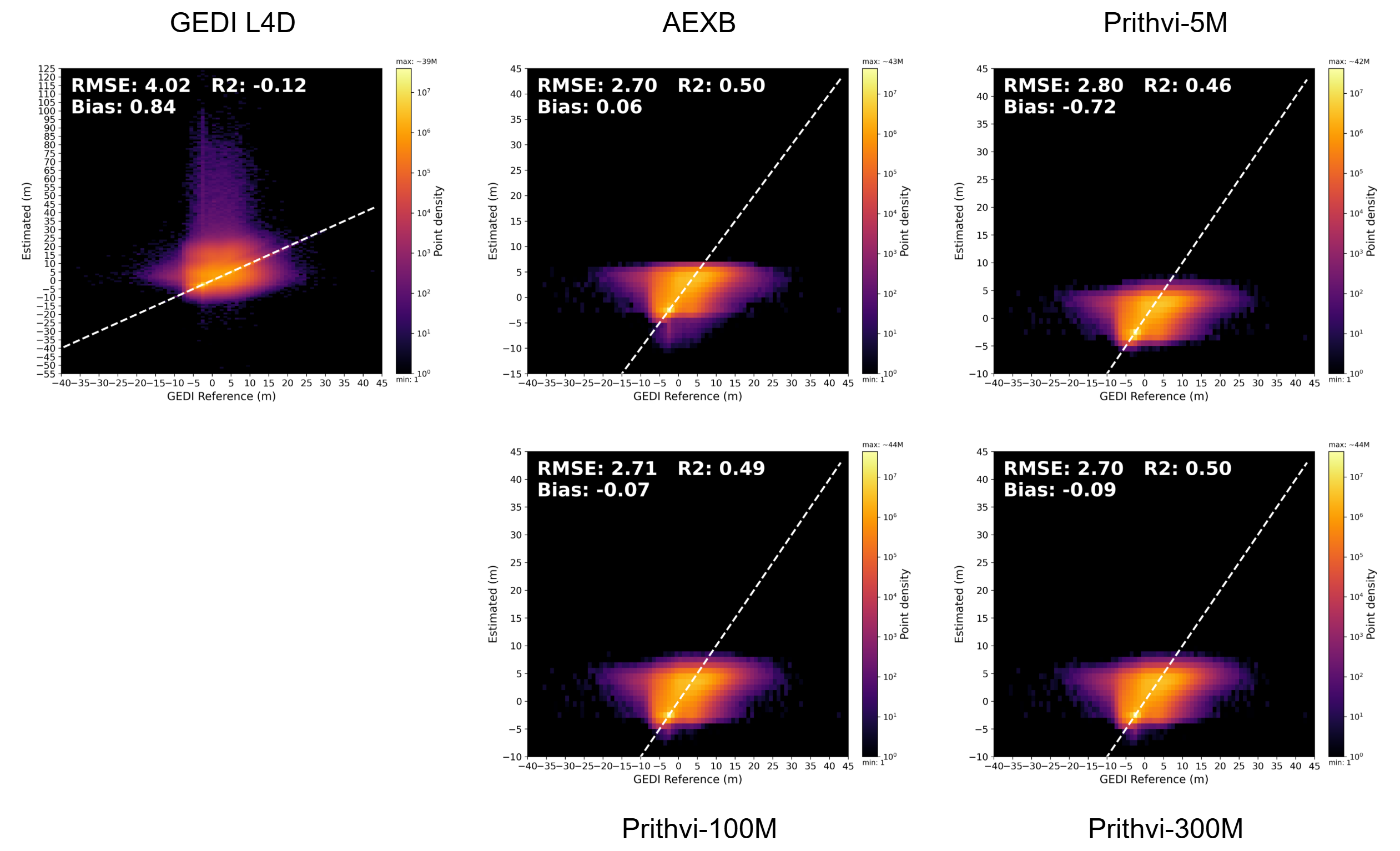}
    \caption{Scatterplot comparison of \acs{RH10} between \acs{GEDI} \acs{L4D} and our baselines.}
    \label{fig:scatterplot_rh10_comparison}
\end{figure*}

\begin{figure*}[!htp]
   \centering
    \includegraphics[width=\textwidth]{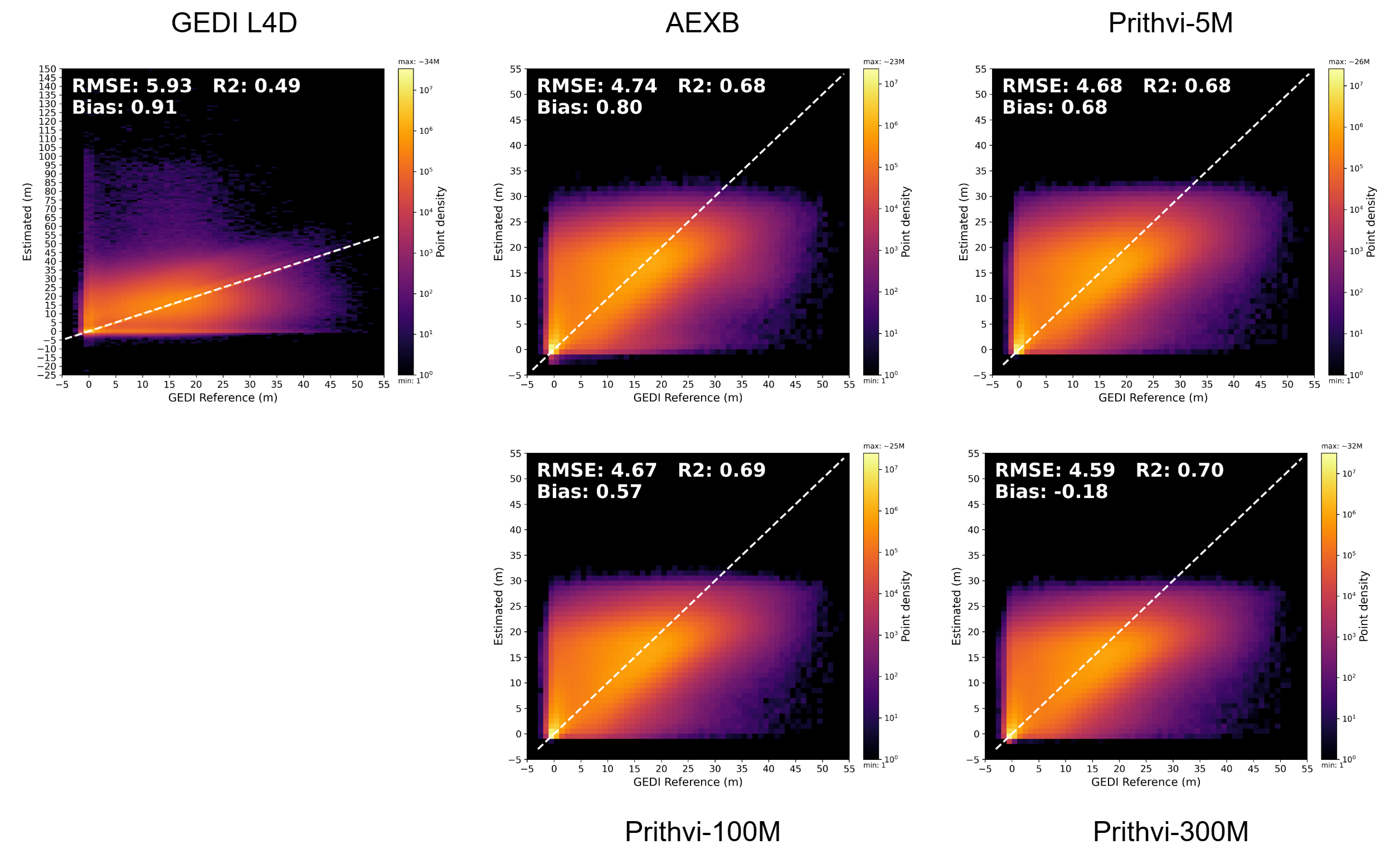}
    \caption{Scatterplot comparison of \acs{RH50} between \acs{GEDI} \acs{L4D} and our baselines.}
    \label{fig:scatterplot_rh50_comparison}
\end{figure*}

\begin{figure*}[!htp]
   \centering
    \includegraphics[width=\textwidth]{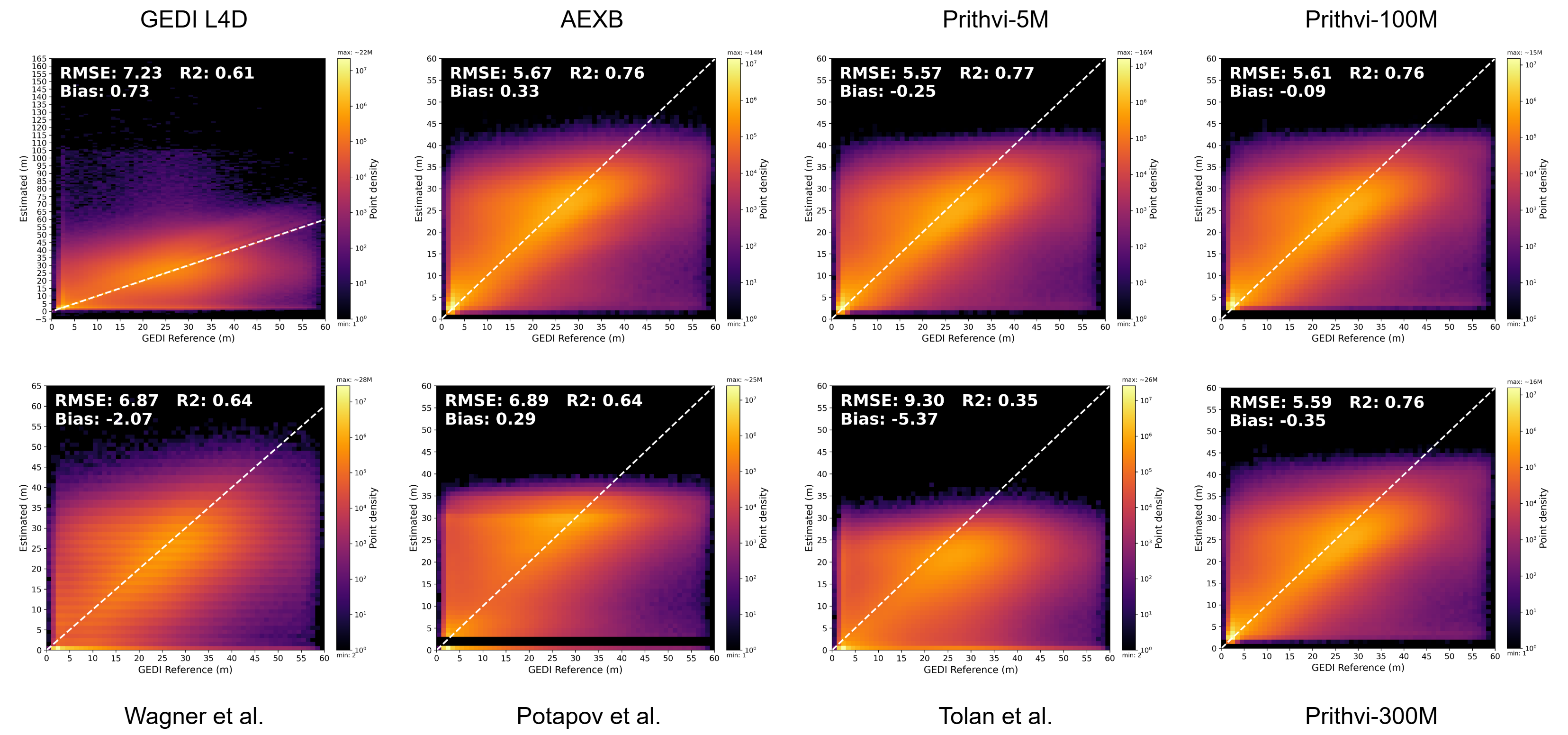}
    \caption{Scatterplot comparison of \acs{RH95} between \acs{GEDI} \acs{L4D}, Wagner et al., Potapov et al., Tolan et al. and our baselines.}
    \label{fig:scatterplot_rh95_comparison}
\end{figure*}

\begin{figure*}[!htp]
   \centering
    \includegraphics[width=\textwidth]{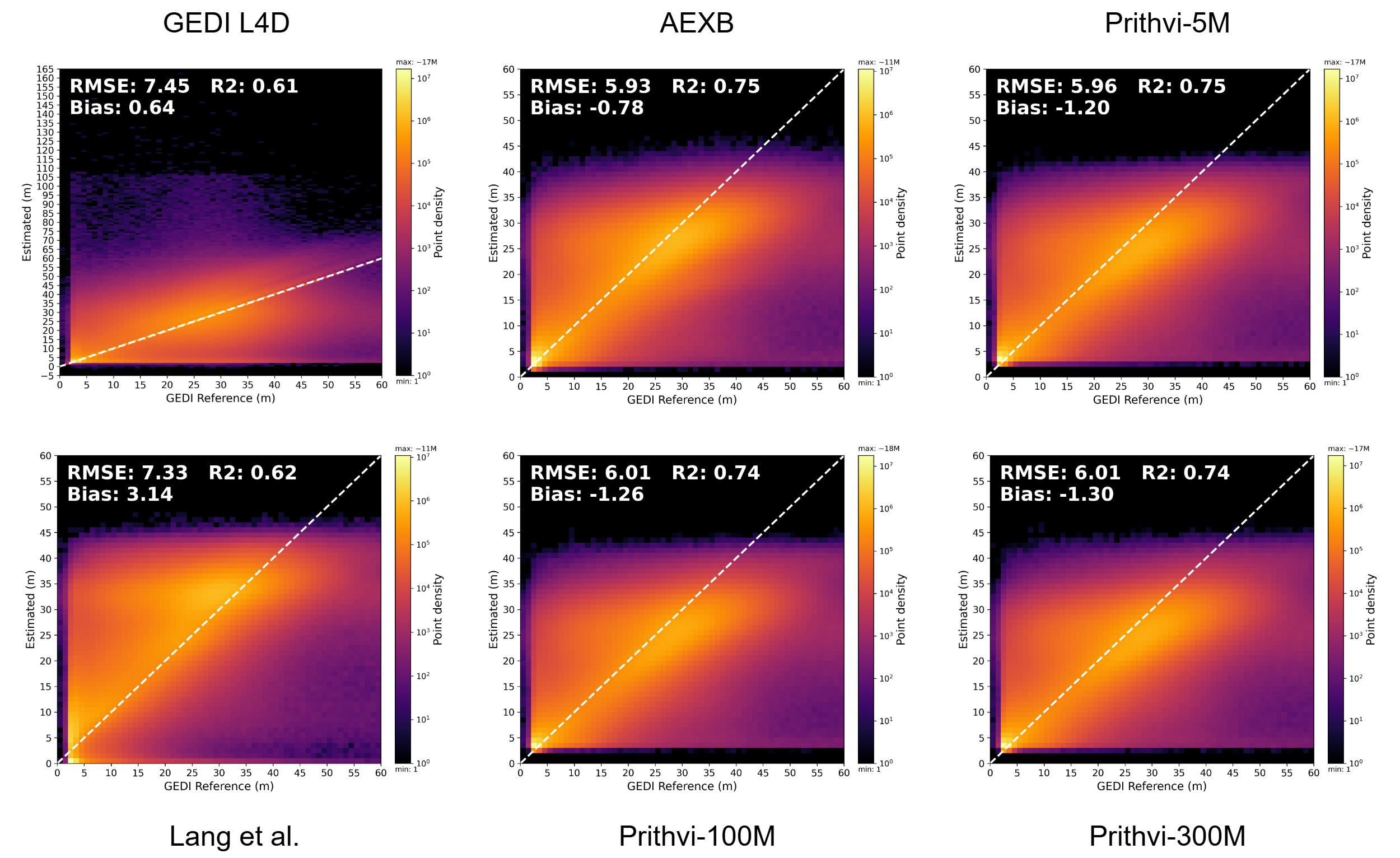}
    \caption{Scatterplot comparison of \acs{RH98} between \acs{GEDI} \acs{L4D}, Lang et al. and our baselines.}
    \label{fig:scatterplot_rh98_comparison}
\end{figure*}

\begin{figure*}[!htp]
   \centering
    \includegraphics[width=\textwidth]{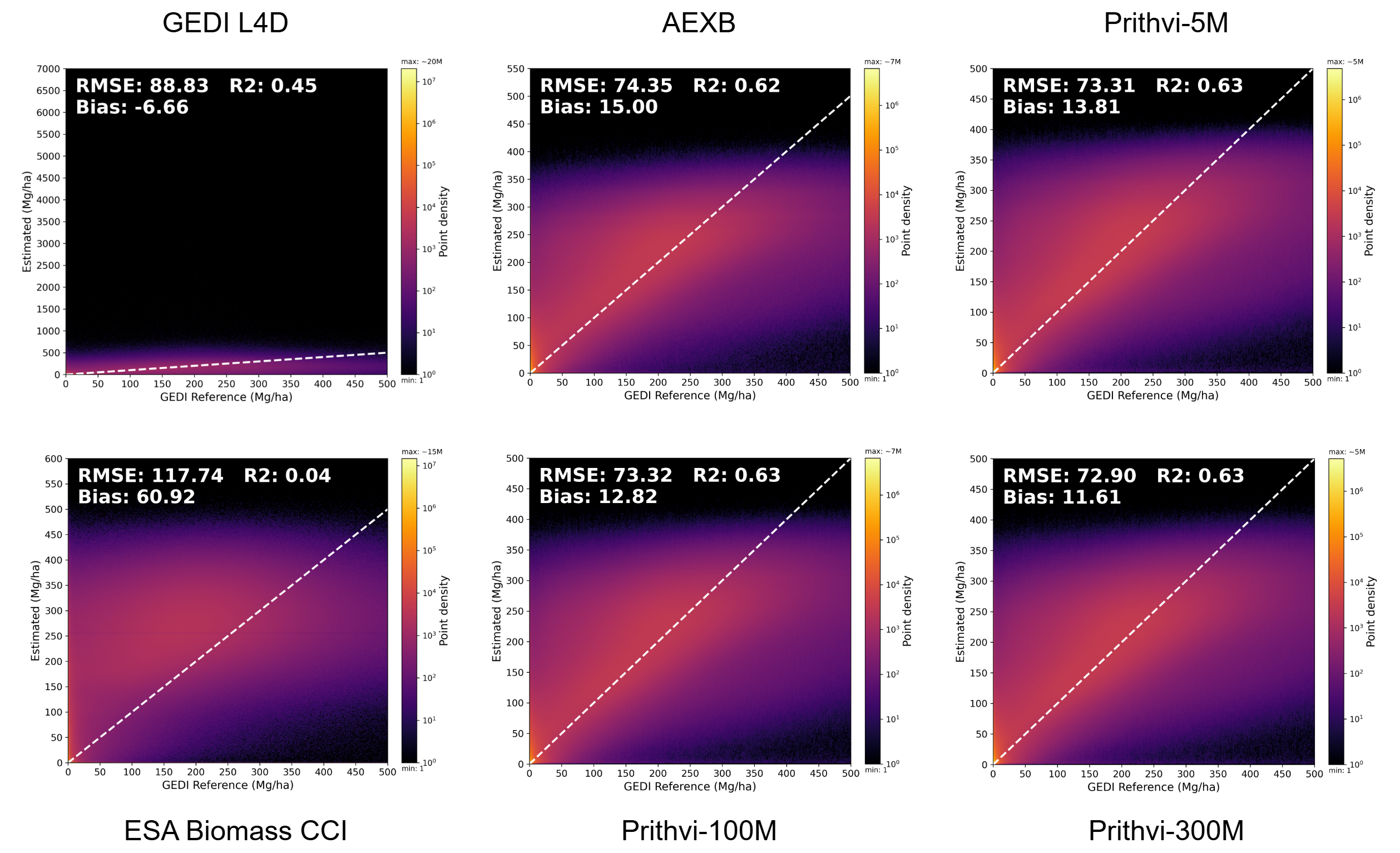}
    \caption{Scatterplot comparison of \acs{AGBD} between \acs{GEDI} \acs{L4D}, ESA Biomass CCI and our baselines.}
    \label{fig:scatterplot_agbd_comparison}
\end{figure*}

To contextualize the \emph{Biomazon} \texttt{\acs{U-RH-AGBD}} baselines against existing wall-to-wall forest-structure products, we evaluate \acs{GEDI} \acs{L4D} \cite{Seo_2025_GEDI_L4D_Imputed_Waveforms_V2}, global canopy-height maps, and \acs{AGBD} products on the same held-out \emph{Biomazon} test grid used for our models. The comparison is product-aligned: \acs{GEDI} \acs{L4D} is evaluated against the matching imputed \ac{RH} percentiles and \acs{AGBD}; Lang et al. \cite{Lang_2023} is compared with \acs{RH98} because their canopy-height product is derived using \acs{GEDI} \acs{RH98}; Potapov et al. \cite{Potapov_2021} is compared with \acs{RH95} because \acs{RH95} is their \acs{GEDI}-based training variable and they also reported \acs{GEDI} \acs{RH95} to have the strongest correlation with the 90th percentile of \ac{ALS}-derived canopy height, because of this, \acs{ALS} based products like Wagner et al. \cite{Wagner_2025} and Tolan et al. \cite{Tolan_2024} are also compared with \acs{RH95}. Santoro et al. \cite{esa_cci} (ESA Biomass CCI) is evaluated only for \acs{AGBD}. All products were reprojected to 20\,m resolution using nearest neighbor interpolation before comparison. Table~\ref{tab:product_aligned_rh_agbd} reports the aggregate metrics, and Fig.~\ref{fig:metrics_comparison} summarizes the \ac{RH} comparison across percentiles. Because these products differ in original spatial resolution, temporal support, input data, and target definitions, the comparison should be read as an external contextual benchmark rather than a strictly controlled model ablation.

Across the \ac{RH} profile, the \emph{Biomazon} baselines consistently improve over \acs{GEDI} \acs{L4D} in \acs{RMSE}, \acs{MAE}, and $R^2$ (Table~\ref{tab:product_aligned_rh_agbd}, Fig.~\ref{fig:metrics_comparison}). The gap is strongest at the lower percentiles, where \acs{GEDI} \acs{L4D} has weak or negative explanatory power, but it remains clear in the upper canopy: at \acs{RH95}, the best baseline reaches 5.57\,m \acs{RMSE} and $R^2=0.77$, compared with 7.23\,m and $R^2=0.61$ for \acs{GEDI} \acs{L4D}, and at \acs{RH98}, \acs{AEXB} gives 5.93\,m \acs{RMSE} and $R^2=0.75$, compared with 7.45\,m and $R^2=0.61$ for \acs{GEDI} \acs{L4D}. The external canopy-height products fall closer to \acs{GEDI} \acs{L4D} than to the \emph{Biomazon} baselines: Potapov et al. and Wagner et al. are competitive with \acs{GEDI} \acs{L4D} in \acs{RH95} \acs{RMSE}, Lang et al. is similar to \acs{GEDI} \acs{L4D} at \acs{RH98}, and Tolan et al. shows the largest negative bias and error over this Amazon test grid. For \acs{AGBD}, the ranking depends on the metric: the Prithvi and \acs{AEXB} baselines reduce \acs{RMSE} and increase $R^2$ relative to both \acs{GEDI} \acs{L4D} and ESA Biomass CCI, but \acs{GEDI} \acs{L4D} retains the lowest \acs{MAE} and smallest absolute bias.

To complement the aggregate metrics, we next inspect whether these differences are also visible in the spatial patterns (Figs.~\ref{fig:rh10_comparison}--\ref{fig:agbd_comparison}) and test-set error distribution (Figs.~\ref{fig:scatterplot_rh10_comparison}--\ref{fig:scatterplot_agbd_comparison}). We focus on representative targets that span the vertical profile: \acs{RH10}, which captures near-ground and lower-canopy returns and is the most challenging case for \acs{GEDI} \acs{L4D} ($R^2=-0.12$, no better than the mean), \acs{RH50} representing mid-canopy structure and \acs{RH95}/\acs{RH98}, which serve as canopy-top proxies and allow comparison with additional height products beyond \acs{GEDI} \acs{L4D}. We also examine \acs{AGBD}, where product differences reflect both structural sensitivity and biomass-specific calibration. The sample maps and test-set scatter plots therefore provide a more diagnostic view of where the numerical gains in Table~\ref{tab:product_aligned_rh_agbd} arise, and whether they correspond to coherent spatial retrievals rather than only improved summary statistics.

A first qualitative observation, evident before any per-method comparison, is that \acs{GEDI} \acs{L4D} spans a physically implausible range of heights. In the \acs{RH10} maps (Fig.~\ref{fig:rh10_comparison}), a percentile that samples near-ground returns and should therefore lie close to zero, isolated \acs{L4D} pixels in the first and third scenes reach 70--80\,m, while the canopy-top \acs{RH98} maps (Fig.~\ref{fig:rh98_comparison}) exceed 100\,m in the same scenes. Both values sit well above any height documented over Amazonia, where the tallest confirmed individual is an 88.5\,m angelim (\emph{Dinizia excelsa}) \cite{tallest_tree}. An \acs{RH98} exceeding 100\,m, and still more a near-ground \acs{RH10} approaching 80\,m, are more consistent with imputation artifacts than with real canopy structure. These extremes appear as sparse, speckled pixels rather than coherent tall-forest patches, consistent with the nearest-shot imputation underlying \acs{L4D} propagating anomalous or noise-corrupted donor waveforms into the gridded field. The same signature appears in \acs{AGBD} (Fig.~\ref{fig:agbd_comparison}): in both samples, scattered \acs{L4D} pixels exceed 3000\,Mg\,ha$^{-1}$, far above any biomass documented for Amazonian forest, and again occur as isolated speckle rather than spatially coherent patches. Across the full test set, the pixel-wise scatterplots (Figs.~\ref{fig:scatterplot_rh10_comparison}--\ref{fig:scatterplot_agbd_comparison}) confirm that this inflation is systematic rather than scene-specific. In every target the \acs{GEDI} \acs{L4D} estimate axis extends far beyond the reference range, reaching roughly 125\,m for \acs{RH10} and 150--165\,m for \acs{RH50}--\acs{RH98} against references below 60\,m, and about 7000\,Mg\,ha$^{-1}$ for \acs{AGBD} against a reference below 500\,Mg\,ha$^{-1}$. The \acs{L4D} predictions consequently form a vertical plume rising above the 1:1 line, whereas the \emph{Biomazon} baselines stay within the reference envelope. The speckle seen in the two map scenes is therefore a property of the \acs{L4D} test-set predictions as a whole.

The three test scenes used in Figs.~\ref{fig:rh10_comparison}--\ref{fig:agbd_comparison} also reveal how the products and our baselines differ in their ability to preserve local spatial structure. The three scenes represent distinct forest contexts: a meandering river floodplain forest (sample~1), a deforestation/agricultural mosaic with remnant forest corridors (sample~2), and a dense terra-firme forest interior (sample~3). Across these settings, the \emph{Biomazon} baselines show coherent local variation that follows visible landscape structure in Sentinel-2, including river margins, forest edges, remnant corridors, and gradual changes within continuous forest. This is most evident in the second sample, where low values align with cleared agricultural fields and higher values trace the remaining forest patches and linear vegetation features. \acs{GEDI} \acs{L4D}, by contrast, often loses such local structure because the dynamic range is dominated by extreme pixels, forcing most of the image into the lower end of the color scale. Its river boundaries are sharply masked and appear as \texttt{NaN} regions (which we assign to minimum value of colormap), unlike the \emph{Biomazon} baselines, which do not impose an explicit water mask but still assign low structural values over river channels. For \acs{RH95}, the \emph{Biomazon} baselines remain spatially consistent across all three samples. Potapov et al. delineates the agricultural mosaic particularly well in the second scene, but appears globally brighter over the floodplain and dense-forest samples, reducing local contrast. Wagner et al. and Tolan et al. retain useful spatial patterns across the samples, although Wagner et al. does not fully follow the Sentinel-2 forest boundaries in the second scene. For \acs{RH98}, Lang et al. spans a wider range than the \emph{Biomazon} baselines, visually consistent with its positive bias in Table~\ref{tab:product_aligned_rh_agbd}. In \acs{AGBD}, both the \emph{Biomazon} baselines and ESA Biomass CCI capture the broad biomass range, but the baselines preserve finer local gradients, whereas \acs{GEDI} \acs{L4D} is again dominated by sparse extreme values reaching the upper end of its 0--3000\,Mg\,ha$^{-1}$ scale.

The scatterplots in Figs.~\ref{fig:scatterplot_rh10_comparison}--\ref{fig:scatterplot_agbd_comparison}
make the calibration differences explicit. The \emph{Biomazon} baselines form compact high-density clouds along the 1:1 line but with a visible upper ceiling. At \acs{RH95} and \acs{RH98} their density stops near 45\,m even where the reference reaches 60\,m, so the tallest canopies are under-predicted. The compression is most pronounced at \acs{RH10}, where the predictions collapse into a narrow cloud near zero. This behaviour largely reflects the \acs{RH10} reference distribution itself, which is narrowly concentrated near zero (Fig.~\ref{fig:distributions}), so the compression is an expected consequence of the target statistics rather than a deficiency of the baselines. The estimate axes themselves are informative. For \acs{GEDI} \acs{L4D}, every target exhibits a long vertical tail in the estimated axis, indicating that the extreme values seen in the sample maps are not isolated cases. This behavior is most striking for \acs{RH10}, where many pixels with near-ground reference values are assigned canopy-height-scale estimates, and for \acs{AGBD}, where estimates extend to several thousand Mg\,ha$^{-1}$. The \emph{Biomazon} baselines avoid these extreme tails and remain within a much narrower, physically plausible range. Among the external height products for \acs{RH95} (Fig.~\ref{fig:scatterplot_rh95_comparison}), Wagner et al., the only one trained over the Amazon Basin, seems to track the reference gradient more closely than Potapov et al. and Tolan et al., although it still scatters broadly and sits slightly below the 1:1 line in line with its mild negative bias (Table~\ref{tab:product_aligned_rh_agbd}). At \acs{RH98} (Fig.~\ref{fig:scatterplot_rh98_comparison}), Lang et al. sits above the 1:1 line at low-to-mid reference, consistent with its positive bias, but shares
the baselines' ceiling near 45\,m. For \acs{AGBD} (Fig.~\ref{fig:scatterplot_agbd_comparison}), the ESA Biomass CCI scatter closely resembles the baselines, both bounded within range and over-predicting low biomass, the main difference being its much lower $R^2$ in Table~\ref{tab:product_aligned_rh_agbd}.

\section{Summary, Limitations and Future Work Directions}
\label{sec:summary_limitations_future}

This section draws together the main findings of the study, the conditions that bound them, and the directions they open for the community. We first summarize what \emph{Biomazon} and its baseline framework establish for joint full-\ac{RH}-profile and \ac{AGBD} modeling over the Amazon Basin. We then set out the principal limitations, which stem largely from the \ac{GEDI}-derived supervision and the dataset construction choices, together with the contextual nature of the product-aligned comparison. Finally, we outline how the benchmark can be used and extended, both as a standardized evaluation setting and as a basis for new datasets
and methods.

\subsection{Summary}
\label{subsec:summary}

\emph{Biomazon} addresses the absence of a benchmark that treats forest vertical structure as a first-class, ordered target by pairing the full \ac{GEDI} \ac{RH} profile with \ac{AGBD} over the Amazon Basin under standardized spatial splits and evaluation protocols. Around this dataset we built a shared encoder--decoder baseline framework and used it to study modality contributions, model scale, training paradigms, and the relationship to existing gridded products. The principal findings are as follows.

\begin{itemize}
  \item \textbf{A structured multimodal benchmark.} \emph{Biomazon} is, to our knowledge, the first 20\,m \acs{ML}-ready dataset to pair the complete 101-band \ac{GEDI} \ac{RH} profile with \ac{AGBD} under leakage-free, tile-level spatial splits, establishing joint full-profile and biomass prediction as a single
  benchmarkable task.

  \item \textbf{Monotone profiles without lower-percentile bias.} The anchored parameterization, which predicts \acs{RH100} and cumulative nonnegative drops, guarantees percentile ordering by construction while still admitting the negative lower-\ac{RH} values present in \ac{GEDI} labels, avoiding the support bias of a
  plain cumulative-sum head.

  \item \textbf{AlphaEarth dominates and capacity saturates.} A shallow \ac{CNN} on the \ac{AEX} field alone matches or exceeds the full multi-sensor Prithvi--\ac{DPT} transformer, while raw-sensor stacking and the 300M backbone add little. This points to \ac{GEDI} supervision density, rather than model capacity, as the
  binding constraint.

  \item \textbf{The useful auxiliary target is task-dependent.} Full-profile (\texttt{\acs{U-RH}}) supervision improves canopy-top \acs{RH98} over single-target \acs{RH98} training, whereas scalar \texttt{\acs{U-RH98-AGBD}} co-training is a more effective companion for \acs{AGBD} than the full profile, in our training setups without task-specific loss weights.

  \item \textbf{Baselines surpass \ac{GEDI} \ac{L4D} and avoid its artifacts.} Across the \ac{RH} profile and \ac{AGBD}, the \emph{Biomazon} baselines consistently improve on \ac{GEDI} \ac{L4D} and remain within a physically plausible range, whereas \ac{L4D} exhibits out-of-range values such as \ac{RH} near 165\,m and \ac{AGBD} of several thousand Mg\,ha$^{-1}$.
\end{itemize}

\subsection{Limitations}
\label{subsec:limitations}

The findings above should be read within several constraints, most of which trace back to the nature of \ac{GEDI} supervision and to deliberate dataset-construction choices rather than to the baseline models alone. We state them here to bound the interpretation of the results and to mark where the benchmark leaves room for
improvement.

\begin{itemize}
  \item \textbf{Performance is supervision-limited, not capacity-limited.} Accuracy plateaus as model size grows, with Prithvi-300M adding no \ac{AGBD} headroom and slightly degrading the \ac{RH} metrics relative to Prithvi-100M. This points to the sparse \ac{GEDI} sampling density, rather than encoder capacity, as the
  binding constraint, so architectural scaling alone is unlikely to move the benchmark.

  \item \textbf{Supervision and evaluation are entirely \ac{GEDI}-derived.} The \ac{AGBD} targets are themselves \ac{L4A} allometric model outputs rather than direct measurements, and the held-out reference is \ac{GEDI} itself. Absolute accuracy is therefore conditioned on \ac{GEDI} quality, with no independent
  airborne-lidar or field validation.

  \item \textbf{The product comparison is contextual, with an embedding caveat.} The external products differ in native resolution, temporal support, and target definition, and are nearest-neighbor reprojected to 20\,m, so the comparison is a contextual benchmark rather than a controlled ablation. Separately, the apparent
  dominance of \ac{AEX} is partly confounded, since it was pretrained against \ac{GEDI} together with the same raw modalities used here.
\end{itemize}

\subsection{Future Work Directions}
\label{subsec:future}

Because \emph{Biomazon} is released as a dataset and protocol rather than a single model, the directions it opens are as much for the wider community as for us. Some follow directly from the limitations above, while others exploit the benchmark as a controlled setting for questions that scalar canopy-height or biomass datasets
cannot pose.

\begin{itemize}
  \item \textbf{A standardized arena for method development and foundation-model benchmarking.} The fixed spatial splits and unified protocol let the community develop and fairly compare new fusion architectures, structured-output methods, and geospatial foundation models for joint \ac{RH}-profile and \ac{AGBD}
  prediction, with our baselines as the reference to improve upon. \emph{Biomazon} can also be incorporated as a structured-regression downstream task in foundation-model evaluation suites such as GEO-Bench \cite{geobench} and Pangea \cite{pangea}, and its large multimodal patch corpus can serve as a source for
  pre-training or continued pre-training, with the resulting transfer assessed under the same protocol.

  \item \textbf{A testbed for physically consistent and uncertainty-aware learning.} Beyond the anchored parameterization benchmarked here, the dataset supports the study of alternative ordering constraints and of probabilistic or distributional heads, and makes the accuracy--consistency tradeoff measurable in a way that
  scalar benchmarks cannot.

  \item \textbf{A controlled setting for evaluating \ac{EO} embeddings.} AlphaEarth, TESSERA, and future embeddings can be compared against raw modalities under identical splits and training recipes, turning the question of when embeddings help \ac{RH}-profile and \ac{AGBD} prediction into a region-specific
  and measurable one.

  \item \textbf{Extension and enrichment of the benchmark.} Natural follow-ons include multi-date versions with disturbance and change targets, extension to other biomes, and the addition of independent airborne-lidar or field validation layers.

  \item \textbf{A reference for downstream science and product assessment.} \emph{Biomazon} can anchor studies of structure--biomass coupling and canopy-strata retrieval, and can serve as an alternative comparison reference, beyond \ac{GEDI} \ac{L4D}, for assessing new gridded \ac{RH}-profile and biomass
  products.
\end{itemize}

\section{Conclusion}
\label{sec:conclusion}

This paper introduced \emph{Biomazon}, a 20\,m multimodal benchmark for Amazon forest structure and biomass modeling. By pairing satellite predictors with the full \acs{GEDI} \acs{RH} profile and \ac{AGBD}, it enables evaluation beyond scalar canopy-height or biomass targets and supports physically consistent vertical-profile prediction under standardized spatial splits.

Our baselines show that anchored monotone \ac{RH} prediction is practical, AlphaEarth embeddings provide strong predictive signal, and larger Prithvi encoders offer limited additional gains. Product-aligned comparisons further show that the \emph{Biomazon} baselines improve over \acs{GEDI} \acs{L4D} across the \ac{RH} profile while avoiding its extreme outliers, though upper-tail compression remains a challenge.

Overall, \emph{Biomazon} provides a common dataset, protocol, and baseline suite for future work on structured forest-profile prediction, structure--biomass modeling, and improved tropical forest monitoring products.

\section{CRediT author statement}

Conceptualization, S.M., R.S.; methodology, S.M.; software, S.M.; validation, S.M.; formal analysis, S.M.; investigation, S.M.; resources, S.M., S.B., M.U., E.Z.; data curation, S.M.; writing---original draft preparation, S.M.; writing---review and editing, S.M., R.S., S.B., M.U., G.C.; visualization, S.M.; supervision, R.S., S.B., M.U., M.R., E.Z., G.C.; project administration, E.Z.; funding acquisition, R.S., G.C.

\section{Funding}

This work was done as part of the 3D-ABC \cite{3dabc} project which is funded by the Helmholtz Foundation Model Initiative supported by the Helmholtz Association. The authors gratefully acknowledge the Gauss Centre for Supercomputing e.V. (\url{www.gauss-centre.eu}) for funding this project by providing computing time through the John von Neumann Institute for Computing (NIC) on the GCS Supercomputer JUWELS \cite{JUWELS} at Jülich Supercomputing Centre (JSC), under GCS/NIC: 3d-abc: 61954.

\section{Data availability}

The dataset, model weights and source code will be made publicly available soon and the respective links will be updated here.

\section{Carbon Impact}

All the experimental runs reported in this paper along with additional supporting experiments, consumed a total energy of $159{,}124.26\,\mathrm{MJ}$ which is equal to $44{,}201.18333\,\mathrm{kWh}$. Using carbon emission factor of Germany from \cite{icha2026strommix}, it results in $44{,}201.18333\,\mathrm{kWh} \times 344\,\mathrm{g\,CO_2/kWh} \approx 15.2\,\mathrm{t\,CO_2}$ emission.

\section{Acknowledgement}

We thank our colleagues at the \emph{SDL AI and ML for Remote Sensing} group of Jülich Supercomputing Centre, \emph{Global Land Monitoring} group of GFZ Helmholtz Centre for Geosciences, and the 3D-ABC project for valuable discussions and feedback throughout this work. We thank NASA's \ac{GEDI} mission for the \acs{L2A} and \acs{L4A} footprint products that underpin the targets of this benchmark; the European Space Agency Copernicus programme for the Sentinel-1, Sentinel-2 and Copernicus GLO-30 Digital Elevation Model data; the Earth Observation Research Center (EORC) of Japan Aerospace Exploration Agency (JAXA) for the ALOS-2 PALSAR-2 ScanSAR products; and Google Research for the Dynamic World V1 land-cover product and the AlphaEarth Foundations embeddings. We also acknowledge Google Earth Engine for the cloud-hosted data access and processing infrastructure that supported dataset construction at Amazon-Basin scale.

\section{Conflicts of Interest}

The authors declare that they have no known competing financial interests or personal relationships that could have appeared to influence the work reported in this paper.

\section{List of Acronyms}
\label{sec:list_acronyms}

\begin{acronym}[CARD4L]
\acro{AEX}{AlphaEarth embeddings}
\acro{AEXB}{AlphaEarth-embeddings base architecture}
\acro{AEX-Base}{AlphaEarth-embeddings base architecture}
\acro{AGB}{Above-Ground Biomass}
\acro{AGBD}{Above-Ground Biomass Density}
\acro{ALS}{Airborne Laser Scanning}
\acro{BOA}{Bottom-Of-Atmosphere}
\acro{CARD4L}{Committee on Earth
Observation Satellites Analysis Ready Data for Land}
\acro{CCDC}{Continuous Change Detection and Classification}
\acro{CCI}{Climate Change Initiative}
\acro{CEOS}{Committee on Earth Observation Satellites}
\acro{CNN}{convolutional neural network}
\acro{DDP}{Distributed Data Parallel}
\acro{DEM}{digital elevation model}
\acro{DN}{digital number}
\acro{DPT}{Dense Prediction Transformer}
\acro{DSM}{digital surface model}
\acro{DTM}{digital terrain model}
\acro{EO}{Earth observation}
\acro{ESA}{European Space Agency}
\acro{FiLM}{Feature-wise Linear Modulation}
\acro{GEDI}{Global Ecosystem Dynamics Investigation}
\acro{GEE}{Google Earth Engine}
\acro{GRD}{Ground Range Detected}
\acro{HLS}{Harmonized Landsat Sentinel-2}
\acro{HPC}{High-Performance Computing}
\acro{IW}{Interferometric Wide Swath}
\acro{L1C}{Level-1C}
\acro{L2A}{Level-2A}
\acro{L4A}{Level-4A}
\acro{L4D}{Level-4D}
\acro{LAI}{leaf area index}
\acro{LDS}{Label Distribution Smoothing}
\acro{LULC}{land use / land cover}
\acro{MAE}{Mean Absolute Error}
\acro{MGRS}{Military Grid Reference System}
\acro{ML}{machine learning}
\acro{MSI}{Multi-Spectral Instrument}
\acro{NIR}{near-infrared}
\acro{QA}{quality assessment}
\acro{RH}{relative height}
\acro{RH10}{GEDI relative height at 10th percentile}
\acro{RH25}{GEDI relative height at 25th percentile}
\acro{RH30}{GEDI relative height at 30th percentile}
\acro{RH50}{GEDI relative height at 50th percentile}
\acro{RH75}{GEDI relative height at 75th percentile}
\acro{RH90}{GEDI relative height at 90th percentile}
\acro{RH95}{GEDI relative height at 95th percentile}
\acro{RH98}{GEDI relative height at 98th percentile}
\acro{RH100}{GEDI relative height at 100th percentile}
\acro{RMSE}{Root Mean Squared Error}
\acro{SAR}{Synthetic Aperture Radar}
\acro{SRTM}{Shuttle Radar Topography Mission}
\acro{SWIR}{shortwave-infrared}
\acro{TL}{temporal--location}
\acro{TOA}{top-of-atmosphere}
\acro{U-RH-AGBD}{Unified joint training of the full RH profile and AGBD}
\acro{U-RH}{Unified training of the full RH profile}
\acro{U-RH98-AGBD}{Unified joint training of RH98 and AGBD}
\acro{VHR}{very-high-resolution}
\acro{ViT}{vision transformer}
\end{acronym}

\bibliographystyle{IEEEtran}
\bibliography{references}

\clearpage
\appendices

\section{Stats of GEDI targets in samples across splits}
\label{sec:app_stats}

\begin{table*}[!b]
\centering
\caption{Descriptive statistics by split of rasterized \acs{GEDI} targets over ALL valid patch pixel samples (256$\times$256, 50\% overlap) present in the dataset.}
\label{tab:gedi_stats}
\renewcommand{\arraystretch}{1.15}
\begin{tabular*}{\textwidth}{@{\extracolsep{\fill}} llrrrrrrrr @{}}
\hline
Product & Split & Count & Mean & Std & Min & 25\% & 50\% & 75\% & Max \\
\hline
\multirow{3}{*}{RH10} & train & 685,962,956 & 0.227 & 3.730 & -65.16 & -2.430 & -1.460 & 2.920 & 51.69 \\
                      & val   &  94,665,114 & 0.568 & 3.705 & -64.54 & -2.350 & -0.630 & 3.310 & 30.53 \\
                      & test  & 183,679,097 & 0.808 & 3.805 & -35.04 & -2.320 & -0.140 & 3.660 & 42.97 \\
\hline
\multirow{3}{*}{RH25} & train & 685,962,956 & 3.887 & 6.049 & -3.50 & -1.230 & 1.010 & 8.690 & 54.13 \\
                      & val   &  94,665,114 & 4.464 & 6.047 & -3.50 & -1.160 & 2.920 & 9.300 & 42.97 \\
                      & test  & 183,679,097 & 4.917 & 6.178 & -3.50 & -1.120 & 3.970 & 9.900 & 47.27 \\
\hline
\multirow{3}{*}{RH50} & train & 685,962,956 & 7.735 & 8.198 & -2.94 & -0.030 & 5.390 & 14.640 & 55.82 \\
                      & val   &  94,665,114 & 8.631 & 8.259 & -2.83 & 0.000 & 7.980 & 15.530 & 51.66 \\
                      & test  & 183,679,097 & 9.250 & 8.332 & -2.91 & 0.110 & 9.220 & 16.170 & 53.71 \\
\hline
\multirow{3}{*}{RH75} & train & 685,962,956 & 11.159 & 9.815 & -2.23 & 1.310 & 9.550 & 19.510 & 57.30 \\
                      & val   &  94,665,114 & 12.391 & 9.975 & -1.90 & 1.340 & 12.550 & 20.780 & 56.26 \\
                      & test  & 183,679,097 & 13.070 & 9.951 & -2.09 & 1.710 & 13.880 & 21.270 & 56.74 \\
\hline
\multirow{3}{*}{RH95} & train & 685,962,956 & 15.325 & 11.371 & -0.41 & 3.780 & 14.280 & 24.850 & 59.32 \\
                      & val   &  94,665,114 & 16.928 & 11.637 & -0.14 & 4.080 & 17.830 & 26.590 & 59.25 \\
                      & test  & 183,679,097 & 17.598 & 11.510 & -0.33 & 5.050 & 19.100 & 26.870 & 59.26 \\
\hline
\multirow{3}{*}{RH98} & train & 685,962,956 & 16.634 & 11.744 & 0.00 & 4.910 & 15.650 & 26.410 & 60.00 \\
                      & val   &  94,665,114 & 18.324 & 12.026 & 0.00 & 5.430 & 19.330 & 28.270 & 60.00 \\
                      & test  & 183,679,097 & 18.987 & 11.882 & 0.00 & 6.500 & 20.590 & 28.500 & 60.00 \\
\hline
\multirow{3}{*}{RH100} & train & 685,962,956 & 18.385 & 12.078 & 0.03 & 6.550 & 17.340 & 28.440 & 102.29 \\
                      & val   &  94,665,114 & 20.167 & 12.356 & 0.03 & 7.300 & 21.220 & 30.380 & 96.49 \\
                      & test  & 183,679,097 & 20.824 & 12.210 & 0.03 & 8.350 & 22.470 & 30.620 & 101.95 \\
\hline
\multirow{3}{*}{AGBD} & train & 683,481,660 & 100.075 & 115.433 & 0.00 & 1.883 & 50.355 & 176.881 & 500.00 \\
                      & val   &  93,712,806 & 115.505 & 120.507 & 0.00 & 1.731 & 83.167 & 201.286 & 500.00 \\
                      & test  & 181,660,684 & 121.423 & 119.943 & 0.00 & 2.604 & 98.454 & 205.688 & 500.00 \\
\hline
\end{tabular*}
\end{table*}

Table~\ref{tab:gedi_stats} reports descriptive statistics of the rasterized \ac{GEDI} targets over \emph{every} valid patch pixel in \emph{Biomazon}, independent of which input modalities are co-present at that pixel. Low-percentile \ac{RH} carries sizeable negative values (e.g., \acs{RH10} minimum of $-65.16$\,m in training), a known characteristic of \ac{GEDI} waveform processing over low-vegetation or bare ground that the supervised heads must reproduce rather than smooth away. Per-percentile means rise monotonically up the profile and reach $\approx 18.4$\,m at \acs{RH100} on the training split. \acs{RH98} maxima sit at exactly $60.00$\,m, and \acs{RH25} minima sit at exactly $-3.50$\,m across all splits, consistent with our capping in the \emph{Biomazon} rasterization pipeline, while \acs{RH100} reaches above $100$\,m; any monotone profile head must accommodate this empirical saturation gap when enforcing the ordering constraint $\hat{r}_p \le \hat{r}_q$ for $p < q$. \ac{AGBD} is strongly right-skewed (Fig.~\ref{fig:distributions}): the training median of $50.4$\,Mg\,ha$^{-1}$ sits at roughly half the training mean of $100.1$\,Mg\,ha$^{-1}$, with a tail extending to the $500$\,Mg\,ha$^{-1}$ cap. Validation and test rows lie above training at every percentile and for \ac{AGBD}, a direct consequence of the val/test eligibility rule (List~\ref{list:tile_split} Step~\ref{step:val_test_split} in Sec.~\ref{subsec:biomazon2}): only tiles where every required band has $\geq 80\%$ valid pixels enter the val/test pool, and those tiles cluster in regions of denser, more continuous forest cover.

Table~\ref{tab:gedi_stats_extras} reports the same statistics restricted to patches where all required modalities (List~\ref{list:tile_split} Step~\ref{step:required_bands}) are jointly present, which is the subset every baseline in Sec.~\ref{sec:results} consumes during training. Val/test rows are unchanged from Table~\ref{tab:gedi_stats}; the training row drops $\sim 19.4\%$ of \ac{RH} pixels and $\sim 19.7\%$ of \ac{AGBD} pixels, mostly from tiles missing an ascending-orbit \ac{SAR} layer. Central tendency shifts upward by well under one training standard deviation (\ac{AGBD} mean $100.08\to 103.35$ and median $50.36\to 56.05$\,Mg\,ha$^{-1}$), while standard deviation and extrema are essentially unchanged. Table~\ref{tab:gedi_stats_extras} is therefore the reference distribution for the baselines reported here, whereas Table~\ref{tab:gedi_stats} is the appropriate reference for methods that consume partially observed patches (e.g., ChannelViT~\cite{bao2024channelvisiontransformersimage}). The percentile coverage of Table~\ref{tab:gedi_stats_extras} is the union of the headline \ac{RH} percentiles reported in the ablation tables of Sec.~\ref{sec:results} (RH25, RH50, RH75, RH95, RH98) and the percentiles available in \ac{GEDI} \ac{L4D} against which we benchmark in Sec.~\ref{subsec:results3} (RH10, RH20, RH30, RH40, RH50, RH60, RH70, RH80, RH90, RH95, RH98); \acs{RH100} is omitted because no reported experiment or external-product comparison uses it. This lets readers look up the split-level statistics of any baseline or product row in Table~\ref{tab:product_aligned_rh_agbd} directly in Table~\ref{tab:gedi_stats_extras} without recomputing those statistics from the raw data.

\begin{table*}[!b]
\centering
\caption{Descriptive statistics by split of rasterized \acs{GEDI} targets over valid patch pixel samples (256$\times$256, 50\% overlap) where all required modalities mentioned in (List~\ref{list:tile_split} Step~\ref{step:required_bands}) are present.}
\label{tab:gedi_stats_extras}
\renewcommand{\arraystretch}{1.15}
\begin{tabular*}{\textwidth}{@{\extracolsep{\fill}} llrrrrrrrr @{}}
\hline
Product & Split & Count & Mean & Std & Min & 25\% & 50\% & 75\% & Max \\
\hline
\multirow{3}{*}{RH10} & train & 553,074,381 & 0.380 & 3.717 & -65.16 & -2.390 & -1.260 & 3.070 & 47.20 \\
                      & val   &  94,665,114 & 0.568 & 3.705 & -64.54 & -2.350 & -0.630 & 3.310 & 30.53 \\
                      & test  & 183,679,097 & 0.808 & 3.805 & -35.04 & -2.320 & -0.140 & 3.660 & 42.97 \\
\hline
\multirow{3}{*}{RH20} & train & 553,074,381 & 3.068 & 5.465 & -42.12 & -1.490 & 0.540 & 7.340 & 53.42 \\
                      & val   &  94,665,114 & 3.383 & 5.421 & -13.73 & -1.460 & 1.710 & 7.640 & 40.5 \\
                      & test  & 183,679,097 & 3.780 & 5.552 & -12.51 & -1.450 & 2.650 & 8.180 & 45.44 \\
\hline
\multirow{3}{*}{RH25} & train & 553,074,381 & 4.091 & 6.097 & -3.50 & -1.190 & 1.490 & 8.980 & 54.13 \\
                      & val   &  94,665,114 & 4.464 & 6.047 & -3.50 & -1.160 & 2.920 & 9.300 & 42.97 \\
                      & test  & 183,679,097 & 4.917 & 6.178 & -3.50 & -1.120 & 3.970 & 9.900 & 47.27 \\
\hline
\multirow{3}{*}{RH30} & train & 553,074,381 & 4.994 & 6.637 & -3.43 & -0.930 & 2.460 & 10.410 & 54.62 \\
                      & val   &  94,665,114 & 5.422 & 6.584 & -3.32 & -0.890 & 4.070 & 10.760 & 44.68 \\
                      & test  & 183,679,097 & 5.922 & 6.710 & -3.39 & -0.890 & 5.170 & 11.400 & 48.47 \\
\hline
\multirow{3}{*}{RH40} & train & 553,074,381 & 6.571 & 7.536 & -3.20 & -0.480 & 4.330 & 12.880 & 55.29 \\
                      & val   &  94,665,114 & 7.111 & 7.489 & -3.06 & -0.440 & 6.130 & 13.290 & 49.12 \\
                      & test  & 183,679,097 & 7.681 & 7.592 & -3.17 & -0.410 & 7.300 & 13.960 & 51.09 \\
\hline
\multirow{3}{*}{RH50} & train & 553,074,381 & 7.979 & 8.293 & -2.94 & -0.030 & 6.060 & 15.010 & 55.82 \\
                      & val   &  94,665,114 & 8.631 & 8.259 & -2.83 & 0.000 & 7.980 & 15.530 & 51.66 \\
                      & test  & 183,679,097 & 9.250 & 8.332 & -2.91 & 0.110 & 9.220 & 16.170 & 53.71 \\
\hline
\multirow{3}{*}{RH60} & train & 553,074,381 & 9.321 & 8.974 & -2.68 & 0.410 & 7.710 & 16.970 & 56.27 \\
                      & val   &  94,665,114 & 10.091 & 8.962 & -2.54 & 0.480 & 9.770 & 17.600 & 53.02 \\
                      & test  & 183,679,097 & 10.744 & 8.998 & -2.65 & 0.670 & 11.050 & 18.200 & 55.09 \\
\hline
\multirow{3}{*}{RH70} & train & 553,074,381 & 10.693 & 9.619 & -2.27 & 0.930 & 9.390 & 18.910 & 56.75 \\
                      & val   &  94,665,114 & 11.590 & 9.635 & -2.16 & 1.040 & 11.590 & 19.690 & 55.44 \\
                      & test  & 183,679,097 & 12.264 & 9.632 & -2.31 & 1.310 & 12.910 & 20.220 & 56.22 \\
\hline
\multirow{3}{*}{RH75} & train & 553,074,381 & 11.424 & 9.943 & -2.01 & 1.230 & 10.270 & 19.920 & 56.98 \\
                      & val   &  94,665,114 & 12.391 & 9.975 & -1.9 & 1.340 & 12.550 & 20.780 & 56.26 \\
                      & test  & 183,679,097 & 13.070 & 9.951 & -2.09 & 1.710 & 13.880 & 21.270 & 56.74 \\
\hline
\multirow{3}{*}{RH80} & train & 553,074,381 & 12.214 & 10.278 & -1.75 & 1.600 & 11.220 & 20.990 & 57.28 \\
                      & val   &  94,665,114 & 13.256 & 10.329 & -1.49 & 1.720 & 13.590 & 21.940 & 56.97 \\
                      & test  & 183,679,097 & 13.939 & 10.281 & -1.86 & 2.170 & 14.910 & 22.370 & 57.23 \\
\hline
\multirow{3}{*}{RH90} & train & 553,074,381 & 14.163 & 11.033 & -1.04 & 2.610 & 13.470 & 23.525 & 58.57 \\
                      & val   &  94,665,114 & 15.384 & 11.129 & -0.70 & 2.880 & 16.090 & 24.680 & 58.39 \\
                      & test  & 183,679,097 & 16.063 & 11.030 & -1.08 & 3.630 & 17.380 & 25.020 & 58.51 \\
\hline
\multirow{3}{*}{RH95} & train & 553,074,381 & 15.585 & 11.516 & -0.41 & 3.590 & 15.020 & 25.310 & 59.32 \\
                      & val   &  94,665,114 & 16.928 & 11.637 & -0.14 & 4.080 & 17.830 & 26.590 & 59.25 \\
                      & test  & 183,679,097 & 17.598 & 11.510 & -0.33 & 5.050 & 19.100 & 26.870 & 59.26 \\
\hline
\multirow{3}{*}{RH98} & train & 553,074,381 & 16.885 & 11.889 & 0.00 & 4.700 & 16.370 & 26.870 & 60.00 \\
                      & val   &  94,665,114 & 18.324 & 12.026 & 0.00 & 5.430 & 19.330 & 28.270 & 60.00 \\
                      & test  & 183,679,097 & 18.987 & 11.882 & 0.00 & 6.500 & 20.590 & 28.500 & 60.00 \\
\hline
\multirow{3}{*}{AGBD} & train & 548,835,261 & 103.345 & 117.404 & 0.00 & 1.630 & 56.049 & 183.690 & 500.00 \\
                      & val   &  93,712,806 & 115.505 & 120.507 & 0.00 & 1.731 & 83.167 & 201.286 & 500.00 \\
                      & test  & 181,660,684 & 121.423 & 119.943 & 0.00 & 2.604 & 98.454 & 205.688 & 500.00 \\
\hline
\end{tabular*}
\end{table*}

\end{document}